\documentclass[amssymb, notitlepage, prx,twocolumn,10pt, aps,prx]{revtex4-1}
\expandafter\let\csname equation*\endcsname\relax
\expandafter\let\csname endequation*\endcsname\relax
\usepackage{empheq}
\usepackage{bm,bbm,dsfont}
\usepackage{amsmath}
\usepackage{amsfonts,upgreek}
\usepackage{amstext,amsthm}
\usepackage{amssymb}   
\usepackage{graphicx}
\usepackage{verbatim}
\usepackage[sort&compress]{natbib}
\usepackage{xcolor}
\usepackage{hyperref}
\usepackage{wrapfig}
\usepackage{mlmodern}
\usepackage{hyperref}
\usepackage[normalem]{ulem} 
\usepackage{soul}
\hypersetup{pdfauthor={UA},pdftitle={Spike},%
            colorlinks, linktocpage=true, pdfstartpage=1, pdfstartview=FitV,%
    breaklinks=true, pdfpagemode=UseNone, pageanchor=true, pdfpagemode=UseOutlines,%
    plainpages=false, bookmarksnumbered, bookmarksopen=true, bookmarksopenlevel=1,%
    hypertexnames=true, pdfhighlight=/O,%
    urlcolor=orange, linkcolor=blue, citecolor=red, 
        }
\newcommand{\comprimi}{%
  \thinmuskip=0mu
  \medmuskip=0mu
  \thickmuskip=0mu
}
\DeclareMathOperator{\de}{d}
\DeclareMathOperator{\e}{e}

\allowdisplaybreaks

\makeatletter
\DeclareFontFamily{OMX}{MnSymbolE}{}
\DeclareSymbolFont{MnLargeSymbols}{OMX}{MnSymbolE}{m}{n}
\SetSymbolFont{MnLargeSymbols}{bold}{OMX}{MnSymbolE}{b}{n}
\DeclareFontShape{OMX}{MnSymbolE}{m}{n}{
    <-6>  MnSymbolE5
   <6-7>  MnSymbolE6
   <7-8>  MnSymbolE7
   <8-9>  MnSymbolE8
   <9-10> MnSymbolE9
  <10-12> MnSymbolE10
  <12->   MnSymbolE12
}{}
\DeclareFontShape{OMX}{MnSymbolE}{b}{n}{
    <-6>  MnSymbolE-Bold5
   <6-7>  MnSymbolE-Bold6
   <7-8>  MnSymbolE-Bold7
   <8-9>  MnSymbolE-Bold8
   <9-10> MnSymbolE-Bold9
  <10-12> MnSymbolE-Bold10
  <12->   MnSymbolE-Bold12
}{}

\let\llangle\@undefined
\let\rrangle\@undefined
\DeclareMathDelimiter{\llangle}{\mathopen}%
                     {MnLargeSymbols}{'164}{MnLargeSymbols}{'164}
\DeclareMathDelimiter{\rrangle}{\mathclose}%
                     {MnLargeSymbols}{'171}{MnLargeSymbols}{'171}
\makeatother

\begin{document}

\hyphenation{ge-ne-ra-li-zed}
\def\hf{\hat{f}}
\def\Ord{\mathcal{O}}
\newcommand{\barr}{\begin{eqnarray}}
\newcommand{\earr}{\end{eqnarray}}
\newcommand{\beq}{\begin{equation}}
\newcommand{\eeq}{\end{equation}}
\newcommand{\be}{\begin{equation}}
\newcommand{\ee}{\end{equation}}
\newcommand{\ra}{\right\rangle}
\newcommand{\la}{\left\langle}
\newcommand{\correc}{ \textcolor{blue} }
\newtheorem{theorem}{Theorem}
\renewcommand{\newblock}{\ }

\newcommand{\bx}{{\boldsymbol{x}}}
\newcommand{\bz}{{\boldsymbol{z}}}
\newcommand{\bk}{{\boldsymbol{k}}}
\newcommand{\bv}{{\boldsymbol{v}}}
\newcommand{\sbv}{{\boldsymbol{\mathsf v}}}
\newcommand{\sbc}{{\boldsymbol{\mathsf c}}}
\newcommand{\sbC}{{\boldsymbol{\mathsf C}}}
\newcommand{\sbQ}{{\boldsymbol{\mathsf q}}}
\newcommand{\bOne}{{\boldsymbol{\mathsf 1}}}
\newcommand{\sbmu}{\boldsymbol{\mu}}
\newcommand{\sblambda}{\boldsymbol{\lambda}}
\newcommand{\hbv}{{\boldsymbol{\hat v}}}
\newcommand{\bA}{{\mathbf{A}}}
\newcommand{\bC}{{\mathbf{C}}}
\newcommand{\bK}{{\mathbf{K}}}
\newcommand{\bW}{{\mathbf{W}}}
\newcommand{\bJ}{{\mathbf{J}}}

\newtheorem{thm}{Theorem}[section]
\newtheorem{lem}[thm]{Lemma}
\newtheorem{prop}[thm]{Proposition}
\newtheorem{corr}[thm]{Corollary}
\newtheorem{result}{Result}[section]
\theoremstyle{remark}
\newtheorem{remark}{Remark}[section]
\newtheorem{assumption}[thm]{Assumption}

\newcommand{\UA}[1]{\textcolor{purple}{[Urte: #1]}}
\newcommand{\GS}[1]{\textcolor{green!50!blue}{#1}}

\newcommand{\gv}[1]{\ensuremath{\mbox{\boldmath$ #1 $}}} 
\newcommand{\uv}[1]{\ensuremath{\mathbf{\hat{#1}}}} 
\newcommand{\abs}[1]{\left| #1 \right|} 
\let\underdot=\d 
\renewcommand{\d}[2]{\frac{d #1}{d #2}} 
\newcommand{\dd}[2]{\frac{d^2 #1}{d #2^2}} 
\newcommand{\pd}[2]{\frac{\partial #1}{\partial #2}} 
\newcommand{\pdd}[2]{\frac{\partial^2 #1}{\partial #2^2}} 
\newcommand{\pdc}[3]{\left( \frac{\partial #1}{\partial #2}
 \right)_{#3}} 
\newcommand{\ket}[1]{\left| #1 \right\rangle} 
\newcommand{\bra}[1]{\left\langle #1 \right|} 
\newcommand{\braket}[2]{\left\langle #1 \vphantom{#2} \right|
 \left. #2 \vphantom{#1} \right\rangle} 
\newcommand{\matrixel}[3]{\left\langle #1 \vphantom{#2#3} \right|
 #2 \left| #3 \vphantom{#1#2} \right\rangle} 
\newcommand{\grad}[1]{\gv{\nabla} #1} 
\let\divsymb=\div 
\renewcommand{\div}[1]{\gv{\nabla} \cdot #1} 
\newcommand{\curl}[1]{\gv{\nabla} \times #1} 
\let\baraccent=\= 
\renewcommand{\=}[1]{\stackrel{#1}{=}} 
\newcommand*\widefbox[1]{\fbox{\hspace{2em}#1\hspace{2em}}}

\newcommand{\numberset}{\mathds}
\newcommand{\N}{\numberset{N}}
\newcommand{\Z}{\numberset{Z}}
\newcommand{\R}{\numberset{R}}
\newcommand{\C}{\numberset{C}}
\newcommand{\Res}{\mathrm{Res}}
\newcommand{\Var}{\mathrm{Var}}
\newcommand{\Cov}{\mathrm{Cov}}
\newcommand{\avg}[1]{\left\langle #1 \right\rangle} 
\newcommand{\T}{\mathcal{T}}
\newcommand{\rs}{\rho^{\star}}
\newcommand{\kk}{\kappa}
\newcommand{\EE}{\mathcal{E}}
\newcommand{\Cc}{\mathcal{C}}

\def\lm{\lambda_-}
\def\lp{\lambda_+}
\def\lpm{\lambda_\pm}

\def\Wt{\tau_\mathrm{W}}
\def\Ht{\tau_\mathrm{H}}
\def\Sm{\mathcal{S}}

\def\Ht{\tau_\mathrm{H}}
\def\Nc{N}
\def\Sm{\mathcal{S}}
\def\rho{\varrho}

\title[Sparse Noise BBP]{Sparse corruption in low-rank matrix inference: the PCA benchmark}

\author{Urte Adomaityte$^*$, Gabriele Sicuro$^\dagger$, Pierpaolo Vivo$^*$}

\address{                 
$^*$King's College London, Department of Mathematics, Strand, London WC2R 2LS, United Kingdom \\
$^\dagger$University of Bologna, Department of Mathematics, Piazza di Porta San Donato 5, 40126, Bologna, Italy
}

\begin{abstract}
Principal Component Analysis (PCA) is a standard tool for extracting a low-rank signal from noisy observations. It is known that when applying PCA to a rank-one signal corrupted by a dense, homogeneous noise, in the large matrix size limit, the celebrated BBP transition occurs, where the emergence of an outlying eigenvalue and the alignment of the corresponding eigenvector occur at the same critical signal strength. Here we study the case of sparse noise corruption. The noise matrix is modelled as the adjacency matrix of a weighted undirected graph with finite average connectivity. Using the replica method, we analytically compute the typical top eigenvalue, the top eigenvector component density, and the overlap with the signal, through recursive distributional equations solved by population dynamics. We identify two signal-strength transitions as functions of graph connectivity: $\theta_{\rm crit}$, marking signal recovery by the top eigenvector and generalising the BBP transition, and $\theta_{\rm b}$, where the signal-related eigenvalue detaches from the bulk. For noise with nonzero mean, these transitions need not coincide because of a structural sparse-graph outlier, leading to a discontinuous transition in the overlap with the top eigenvector. 
For a Random Regular degree distribution, we obtain exact closed-form expressions for both thresholds, the signal eigenvalue and the squared overlap, and show that the same formulas also predict the overlap of the signal with the second eigenvector in the intermediate regime $\theta_{\rm b}<\theta\leq \theta_{\rm crit}$,
where the transition is continuous. For a Poissonian degree distribution, an approximate closure of the recursions yields fast and accurate predictions in the recovery phase that agree with the exact results from population dynamics.
We also recover dense-noise results in the large-connectivity limit, and validate the theory by numerical diagonalisation of large matrices.
\end{abstract}

\maketitle

\section{Introduction}

The estimation of a low-rank signal from a noisy observation in high dimensions is a fundamental problem in machine learning, statistics and probability. This task appears in Principal Component Analysis (PCA) \cite{Jolliffe02PCA,Candes11}, community detection \cite{Holland83,Abbe18} or matrix completion \cite{Candes12}, among others. Spiked random matrix models have been extensively investigated as a prototypical setup for the theoretical modelling of these problems \cite{Johnstone01, BBP05, Forrester23_review}. In the rank-one signal case, one seeks to recover a vector $ \bx \in \R^N$ from a noisy observation 
\begin{equation}
    \bA =\bJ + \frac{\theta}{N} \bx \bx^{\intercal} , \label{eq:definition_intro}
\end{equation} 
where $\theta$ is the signal-to-noise ratio and the noise is modelled via the random matrix $\bJ \in \R^{N \times N}$, typically dense, full-rank and random. 
For various assumptions on the spike and noise matrix, observables of interest that quantify signal recovery are the value of the typical top eigenvalue, the overlap of the top eigenvector with the signal, the spectral density, and the top eigenvalue distribution. Of special interest are BBP-like transitions in the signal strength (named after the authors of Ref.~\cite{BBP05}): a value $\theta_{\mathrm{crit}}$ such that for $\theta>\theta_{\mathrm{crit}}$ the top eigenvalue detaches from the right bulk edge of the spectrum, the top eigenvector aligns non-trivially with the signal, and a naïve spectral algorithm such as classical PCA succeeds in recovering the signal.

\begin{figure*}
    \centering
    \includegraphics[width=0.32\linewidth]{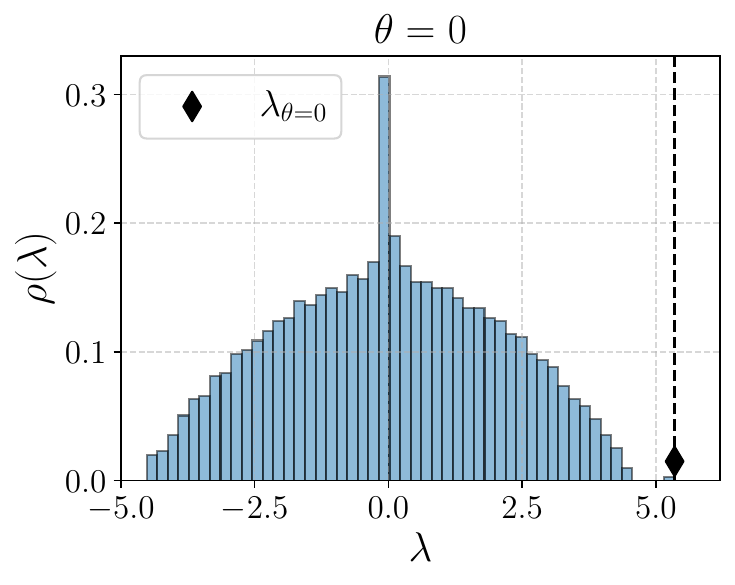}
    \includegraphics[width=0.32\linewidth]{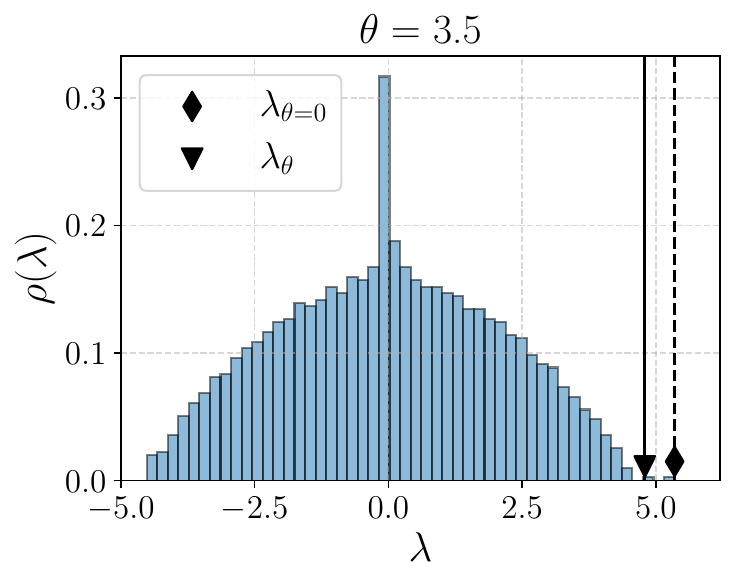}
    \includegraphics[width=0.32\linewidth]{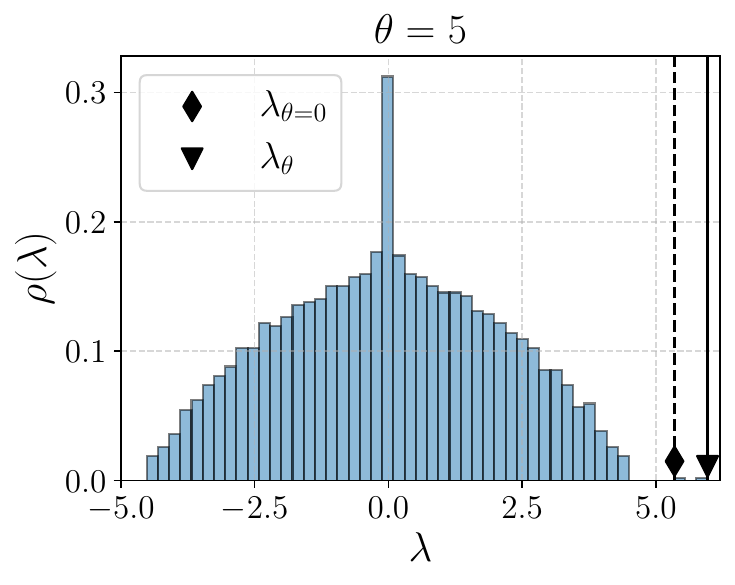} 
  \caption{Eigenvalue density of the matrix $\bA$ of size $N=2000$ as defined in Eq.~\eqref{eq:definition_intro}, with $\bJ$ a pure adjacency matrix of a Poissonian graph with average connectivity $c=4$ and signal $x_i \sim \mathcal{N}(0,1)$. ({\it Left}) Density for $\theta=0$, corresponding to $\bA=\bJ$, where the structural noise eigenvalue $\lambda_{\theta=0}$ is the sole outlier (diamond and dashed line). Matrices are generated using an unbiased configuration model algorithm that can be found in \cite{CoolenAnnibale17}. ({\it Center}) Density for $0<\theta<\theta_\mathrm{crit}$: the eigenvalue associated with the signal $\lambda_\theta$  (triangle and solid line) is either an outlier such that $\lambda_\theta<\lambda_{\theta=0}$, or is buried in the bulk, and the top eigenvector does not correlate with the signal vector $\bx$. ({\it Right}) For $\theta>\theta_\mathrm{crit}$, the eigenvalue associated with the signal $\lambda_\theta$ is the top eigenvalue, and the top eigenvector correlates non-trivially with the signal vector $\bx$.}
  \label{fig:spectra}
\end{figure*}

The classical inference and retrieval setting suffers from a practical drawback, though, as it implicitly assumes that a practitioner observing the full matrix $\bA$ also holds some separate information about the nature of the corrupting noise (e.g., being dense and full rank), which instead cannot be directly observed or probed. What is the cost incurred in the retrieval process in a case where the practitioner applies the na\"ive PCA algorithm to a problem where the noise is actually sparse and structurally inhomogeneous?

To answer this question, we have to tackle the more challenging problem of computing the recovery threshold of naïve PCA---that is, recovery of the signal via the top eigenvector of the matrix $\bA$---when the usual, comforting assumption that the noise $\bJ$ be dense is dropped. 
There is a wealth of important applications in which the \emph{sparse} structure of the noise \footnote{That is, with an extensive number of zero entries.} is relevant, such as video, image, multimedia processing, notably video surveillance and facial recognition (see \cite{Candes11,Vary21} for more detailed accounts), yet the classical random matrix toolkit is ineffective when one cannot rely on rotational invariance and Gaussian assumptions on the matrix entries. Borrowing tools from the physics of disordered systems, we have been able to overcome this technical challenge and present analytical results for important spectral observables in the \emph{sparse spiked problem}, including a precise characterisation of the recovery threshold that generalises the celebrated BBP transition to the sparse noise case. 

For a model like Eq.~\eqref{eq:definition_intro} but with a sparse $\bJ$ (see Section~\ref{sec:model}), we obtain the value of the typical top eigenvalue, the density of the components of the top eigenvector, the typical value of the overlap between the spike vector and the top eigenvector and the density of its components. 
We also compute the threshold value $\theta_{\mathrm{crit}}$ at which the top eigenvector starts to correlate with the spike, i.e., the overlap becomes non-trivial. 
Note that, by allowing for noise with nonzero mean, two events that coincide in the canonical dense BBP transition can separate: detachment of the signal-related eigenvalue from the bulk (occurring at $\theta_{\rm b}$) and the onset of nonzero overlap between the leading eigenvector and the signal (occurring at $\theta_{\mathrm{crit}}$). Thus, there can be an intermediate regime in which the signal is spectrally visible but not recoverable by the top eigenvector. As a consequence, and unlike the classical BBP case, the overlap between the signal and the top eigenvector jumps discontinuously to a nonzero value at  $\theta_{\mathrm{crit}}$. 
We illustrate this mechanism in Fig.~\ref{fig:spectra}.
The spectral observables and corresponding transition values depend on the structural parameters of the spiked matrix model $\bA$ such as the average connectivity and assumptions on the bond weight and spike component distributions. We further show that in the large connectivity limit, we recover known values of the BBP-like transition, the top eigenvalue and the squared overlap for the spiked real Wigner matrix ensemble \cite{Peche06,Capitaine09,Bloemendal_2012}.

We employ the replica formalism \cite{Mezard87} developed by \cite{Kuehn2008, Bianconi08, Kabashima12} and used in \cite{Takahashi14,Susca19,Susca21,Budnick25} to calculate spectral properties of large symmetric matrices with sparse structure, namely when the average connectivity of the noise matrix remains finite as $N\to\infty$. Those works address the spectral properties of sparse matrices without a planted low-rank signal, corresponding in our notation to the zero-signal case, recovered by setting $\theta=0$ in Eq.~\eqref{eq:definition_intro}.
The presence of a non-zero signal $(\theta\neq 0)$ poses significant technical challenges that require a non-trivial reworking of the standard formalism. The extremal observables are determined via the solution of coupled recursive distributional equations (RDEs), i.e., integral equations for auxiliary probability density functions, which can be efficiently solved using a population dynamics algorithm \cite{Mezard00}. RDEs appear in statistical physics and inference \cite{Chertkov10, Semerjian20} in the analysis of infinite-graph-size solutions of message-passing/belief propagation algorithms. They are recovered in the mathematical literature as `stochastic recursive equations' in problems on random graphs; see \cite{Volkovich07_rde,Jelenkovic09_rde,Chen17_rde,Fraiman23_rde} for their analysis of the Google PageRank vector distribution. Other network centrality measures are analysed via the cavity formalism that leads to RDEs solved via population dynamics in \cite{Vivo24}. The cavity method, exact for sparse tree structures \cite{Bordenave10}, is also used as an alternative to the replica calculation and leads to the same results for the spectral density \cite{Slanina11}, top and second eigenpair problems \cite{Susca19, Susca21}.

In the next section, we review the related literature in inference, statistical physics and machine learning.

\subsection{Related works}\label{sec:related_works}

The analysis of detection limits in large spiked random matrix models, where the noise is assumed to be a large random matrix that is added to the low-rank signal matrix, began with works on phase transitions in the spiked Wishart \cite{BBP05} and spiked Wigner \cite{Peche06,Capitaine09,Bloemendal_2012} models. An earlier seminal work \cite{Johnstone01} employs random matrix theory to analyse the distribution of the top eigenvalue of a large covariance matrix model, motivated by the analysis of PCA. It has inspired a wide array of works in multiple domains, with a particular focus on high-dimensional limits, e.g., \cite{barbier16, lesieur17, lelarge17a, montanari24}. Ref.~\cite{BenaychGeorges11} considers the top eigenvalue and squared overlap of the case where either the noise matrix or the signal is rotationally (or unitarily) invariant.
More recently, \cite{dumitriu2026bbpphasetransitiondoubly} have proved the BBP phenomenon for a supercritical sparsity regime for the noise matrix and and for any sparse spike vector with an unbounded number of entries. In Ref.~\cite{ferreira2026bbptransitionleadingeigenvector}, the authors derive spectral properties where the the variance of each entry of the noise matrix is itself a random variable, resulting in a non-monotonic BBP transition line. Similarly, Ref.~\cite{bocchi2026discontinuousbbptransitions} analyses a subset of spiked matrix ensembles, namely deformed Gaussian and reweighted Wishart ensembles, where the BBP transition is discontinuous. 
Large deviations of the top eigenvalues have also been considered for dense random matrices with a rank-one spike \cite{Biroli19}. 

Our paper aims to determine the extent to which naïvely applying PCA affects the recovery of the low-rank signal in high dimensions when the practitioner does not have reliable prior information about the sparsity of the noise. Note that, when instead it is a priori established that the noise structure is sparse, there is an array of specialised PCA algorithms one can apply, namely Robust PCA \cite{Candes11,Wright09} to recover low-rank $L$ and sparse noise $S$ from an observation $L+S$. Notable extensions include \cite{Chandrasekaran09,Zhouchen11,Waters11,Netrapalli14,Vary21,He23}. For a dense noise model, universality of the Fisher information in recovery was established in terms of the minimum mean squared error (MMSE) in \cite{Lesieur15}, where a dense probabilistic analogue of the sparse noise matrix is covered: small noise is added with probability $p$ and large noise with probability $1-p$. 

There is a current line of works that generalises assumptions on the noise matrix $\bJ$ in low-rank matrix inference to analyse algorithmic recovery thresholds, albeit in the case where $\bJ$ remains dense. In particular, known BBP-like transition values and efficient algorithms that incorporate prior information such as Approximate Message Passing (AMP) are derived that perform well where naïve spectral methods such as PCA would fail. 
In particular, Ref.~\cite{Guionnet22} considers heterogeneous but dense noise, prove Gaussian universality of the free energy and the spectrum and determine an information theoretical threshold for the dense degree-corrected Stochastic Block Model. Refs.~\cite{Alberici_2021,Alberici_2022,Behne22a,Mergny24b} consider dense block-structured noise models, while Ref.~\cite{Mergny24a} analyses non-linear observation channels with zero Fisher information and provides a unified view of theoretical thresholds and optimal AMP algorithms for that class of models. Also, Ref.~\cite{Barbier22} considers rotationally invariant dense noise and evaluate how Bayes estimators and AMP perform with mismatched noise priors (incorrectly assuming Gaussian noise in particular), while Refs.~\cite{Barbier23, Barbier25} extend AMP for a spiked model where dense noise is no longer independent but still rotationally invariant, and provide information-theoretical limits along with an efficient algorithm. 
Note, however, that contrary to our setup, they consider dense noise matrices and determine algorithmic thresholds.

Spectral algorithms on a sparse matrix are studied in community detection tasks, where \cite{Krzakala13} first studied the non-backtracking operator (introduced in Ref.~\cite{Hashimoto89} in the context of zeta functions): unlike working with the adjacency matrix, spectral algorithms on this operator achieve information-theoretical performance; its spectral properties are investigated further in \cite{Saade14, Bordenave15}. 
The cavity method is also employed in community detection on sparse adjacency matrices with finite connectivity, see \cite{Decelle11} and references therein.
The non-backtracking operator is also studied in the top eigenpair problem for (unspiked) sparse matrices via the cavity method in \cite{Kabashima12, Susca19}. 
Recently, \cite{Tapias25} analysed the top eigenvalue detection via the real-valued cavity method for the Anderson model \cite{Anderson58} and found that criteria for finding the solution based on the non-backtracking operator are unsatisfactory when the top eigenvector is localised. Localisation transitions of sparse graph adjacency matrices are studied in \cite{Metz10,Metz21}, and a treatment of the case of sparse non-Hermitian matrices is in \cite{Neri12,Metz19}. 
The raw top eigenvector of the adjacency matrix is known to fail in certain sparse community-detection regimes where non-backtracking operators or Bethe Hessians succeed \cite{Krzakala13}. In this work, we elucidate an analytically controlled sparse-BBP mechanism for this kind of failure: the leading eigenvector can remain dominated by a structural sparse-graph mode until the signal strength exceeds $\theta_{\rm crit}$. This explains, in the simplest rank-one sparse-noise setting, how structural modes can obstruct na\"ive spectral recovery. It also provides a baseline against which improvements from more elaborate sparse-aware spectral or message-passing methods can be measured.

More broadly, the structure of `sparse plus low-rank' large size matrices appears in recent works on fine-tuning and compression methods for large language models, motivated by the considerable storage and training costs incurred because of their size.  
In Ref.~\cite{Han24_SLTrain}, the authors improve the memory efficiency of the industry-standard fine-tuning technique low-rank adaptation (LoRA) \cite{hu22lora}, where a dense pre-trained weight matrix is updated with a low-rank correction, by parametrising the dense weight matrix as a sparse plus low rank structure.
Additionally, the `sparse plus low-rank' matrix structure is key in Ref.~\cite{zhang25oats}: an outlier-aware thresholding method (akin to Robust PCA) used to compress LLMs by decomposing the weight matrices into a sum of a sparse matrix and a low-rank matrix; it achieves significant speed-up on inference without retraining and state-of-the-art performance on some benchmarks. The random matrix model we study in this work provides an analytically controlled benchmark for the basic spectral question of when a low-rank component becomes visible, and recoverable by na\"ive PCA, against a sparse background. While empirical neural-network weights have additional structure and substantially greater complexity, this benchmark represents an important first step towards an analytical theory of sparse-plus-low-rank mechanisms in compression and fine-tuning, with extensions to concrete architectures requiring further model-specific analysis.

\subsection{Outline}

The presentation of the model and the formulation of the optimisation problem in terms of replica analysis is provided in Section~\ref{sec:model}.
The main results are outlined in Section~\ref{sec:results} and specialised in Section~\ref{sec:discussion} to particular degree distributions. We conclude and present an outlook for future works in Section~\ref{sec:conclusion}. The Appendices are devoted to the full replica calculations, technical derivations and the description of the population dynamics algorithm.

\section{The model}\label{sec:model}
We will consider a $N\times N$ symmetric random matrix in the form
\begin{equation}
\bA = \bC \odot \bW + \frac{\theta}{N} \bx \bx^\intercal \ ,\label{eq:model_definition}
\end{equation}
where $\odot$ is the Hadamard product multiplying two $N \times N$ matrices element-wise. 

The \textit{signal} $\bx\in\R^N$ is constructed by sampling identically and independently each of its components $x_i$ from a density $\rho_x$ on $\R$. 
We will work with distributions with zero mean and finite second moment. The parameter $\theta \geq 0$ modulates the strength of the rank-one ``spike'' $\bx \bx^\intercal$. 

Both the matrix $\bC=(c_{ij})_{ij}\in\{0,1\}^{N\times N}$ and the matrix $\bW=(W_{ij})_{ij}\in\R^{N\times N}$ are assumed to be symmetric and random, so that $\bJ\coloneqq \bC\odot\bW$ plays the role of a \textit{symmetric noise matrix}. The matrix $\bJ$ is sparse; its average connectivity remains finite as the size $N\to\infty$. To be more precise, we construct $\bC$ by considering the adjacency matrix of an undirected ``auxiliary'' random graph of $N$ vertices sampled from the configuration model ensemble, with degree sequence $\bk\coloneqq(k_1, \dots, k_N)$. The degree $k_i$ of the $i$th node is independently sampled from a distribution $p_k$, identically for all vertices. The distribution $p_k$ is assumed to have finite support with largest element $k_{\rm max}$. 
Given a function $f$ defined on the support of $p_k$, we will use the notation 
$$\langle f(k)\rangle\coloneqq \sum_k f(k)p_k\ .$$ In particular, we will denote
$$c\coloneqq\langle k\rangle$$
the average degree of the vertices in the auxiliary graph.

We will therefore assume the joint distribution of entries of $\bJ$ to be in the form
\begin{equation}\comprimi
    P(\bJ|\bk)=P(\bC|\bk)p(\bW)\prod_{ij}\delta(J_{ij}-c_{ij}W_{ij}) \ . \label{eq:general_J_condk}
\end{equation} 
The distribution of connectivities given the degree sequence has the form
\begin{equation}
P\left(\bC|\boldsymbol k \right)\propto \prod_{i=1}^N\delta_{c_{ii},0}\mathbb I\left(\sum_{j=1}^Nc_{ij}=k_i\right) \label{eq:boundedER_connectivity}
\end{equation}
where $\mathbb I(S)=1$ if $S$ is true and $0$ otherwise.
The density of bond weights is assumed to be in the form
\begin{equation}
    p(\bW)=\prod_{i<j}\delta(W_{ij}-W_{ji})\rho_W(W_{ij})\prod_{i}\delta(W_{ii}) \ ,
\end{equation}
where the density $\rho_W$ has a compact support with upper edge $\zeta$ and $\mathbb E_W[W]\neq0$. The Gershgorin circle theorem \cite{Axler97} implies then that the top eigenvalue $\lambda_{\rm top}$ of $\bA$ is of $\mathcal{O}(1)$ for large $N$.
For illustration, we will consider two special cases specified by the choice of $p_k$:

\begin{description}
    \item[Random regular] In the random regular (RR) setup, we simply pick
    \begin{equation}\label{eq:RR}p_k=\delta_{k,c}
    \end{equation}
    for a certain $c=k_{\rm max}>2$.
    \item[Poisson] In the Poissonian connectivity setup, the distribution $p_k$ is assumed to be a \textit{truncated} Poisson distribution depending on the parameter $\bar c>0$ 
    \begin{equation}\comprimi \label{eq:ER} 
        p_k =\frac{\bar{c}^k}{\Gamma k!},\quad \Gamma\coloneqq \sum_{k=0}^{k_{\mathrm{max}}}\frac{\bar{c}^k}{k!},  \qquad k\in\{0,1,\dots,k_{\rm max}\} ,
    \end{equation}
    where the relation between $c$ and $\bar c$ is given by $c \coloneqq\langle k\rangle= \sum_{k=0}^{k_{\mathrm{max}}}k\frac{\bar c^k}{k!}/ \sum_{k=0}^{k_{\mathrm{max}}}\frac{\bar c^k}{k!}$.
    In this case, the expected average degree $c\to \bar c$ as $k_{\rm max}\to +\infty$. 
\end{description}
\noindent Recall however that the results we obtain are valid for general $p_k$ satisfying the assumptions.

\subsection{The top eigenvalue}\label{sec:formulation}
We will now sketch the main ideas behind the replica calculation to characterise the statistical properties of the typical top eigenvalue $\lambda_{\rm top}$ of $\bA$. We will see, in particular, that the \textit{replica trick} is a powerful tool to study a variety of relevant observables in the large $N$ limit.

Let us start by observing that, by means of the Courant--Fischer--Weyl min-max principle, it is possible to characterise the top eigenvalue $\lambda_{\rm top}$ of $\bA$ as
\[N \lambda_{\rm top}=\max_{\|\bv\|^2=N} \langle\bv,\bA\bv\rangle \ , \]
where $\langle\cdot,\cdot\rangle$ is the dot product between vectors in $\R^N$.
Under a non-degeneracy assumption, the maximum is achieved by a single vector $\bv_{\mathrm{top}}$ (up to a $\Z_2$ symmetry), corresponding to the eigenvector associated to $\lambda_{\rm top}$. This fact suggests that we may recast the problem into a quadratic optimisation problem, namely introduce the Gibbs-Boltzmann distribution in the form
\begin{equation}
P_{\beta}\left(\bv;\bA\right)=\frac{1}{Z_{\beta}(\bA)}\exp\left(\frac{\beta}{2}\langle\bv,\bA\bv\rangle\right)\delta(\|\bv\|^2-N) \ ,\label{eq:hard}
\end{equation}
with the partition function
\begin{equation}
    Z_{\beta}(\bA)\coloneqq \int \de\bv \exp \left(\frac{\beta}{2}\langle\bv,\bA\bv\rangle \right)\delta(\|\bv\|^2-N) \ . \label{eq:Z_lambdatop_soft}
\end{equation}
By means of a saddle-point approach in the $N\to\infty$ limit, the solution is obtained in the low temperature limit $\beta\to\infty$ as the measure concentrates on the eigenvector associated to the top eigenvalue.
As a consequence,
\begin{align}
     \mathbb E_{\bA} \left[ \lambda_{\rm top} \right] = \lim_{N \to \infty}\lim_{\beta \to \infty} \frac{2}{\beta N} \mathbb E_{\bA} [ \ln Z_{\beta}(\bA) ] \ .
\end{align}
Taking the average of the logarithm of the partition function is famously cumbersome, so we employ the celebrated replica trick \cite{Mezard87} to write
\begin{align}
     \mathbb E_{\bA} \left[ \lambda_{\rm top} \right] = \lim_{N \to \infty}\lim_{\beta \to \infty} \frac{2}{\beta N}  \lim_{n\to 0} \frac{1}{n} \ln \mathbb E_{\bA} [ Z_{\beta}^n(\bA)] \ . \label{eq:replica_trick_Zn}
\end{align}
The bulk of the technical part is dedicated to computing $\mathbb E_{\bA} [ Z_{\beta}^n(\bA)]$, the average of the partition function replicated $n$ times, where $n$ is first taken as an integer and then analytically continued to real values around $n=0$. Note that in practice, there is an implicit assumption that the limit $N \to \infty$ and the replica limit $n\to 0$ can be exchanged to give
\begin{align}
     \mathbb E_{\bA} \left[ \lambda_{\rm top} \right] = \lim_{\beta \to \infty} \frac{2}{\beta}   \lim_{n\to 0} \frac{1}{n} \lim_{N \to \infty} \frac{1}{N}\ln \mathbb E_{\bA} [ Z_{\beta}^n(\bA)] \ . \label{eq:replica_trick_exchNn}
\end{align}
The full calculation is presented in Appendix~\ref{app:Afullreplica}. 

\subsection{Eigenvector of the top eigenvalue} 
Let us show now how some properties of the eigenvector associated to $\lambda_{\rm top}$ can be computed via the same trick. Suppose that we are given a function $F\colon \R\times\R^N\times\R^N\to\R$. Let $\bx$ be the spike vector as defined by the model formulation and $\bv_{\mathrm{top}}$ the eigenvector of $\bA$ corresponding to $\lambda_{\rm top}$, hereafter also referred to as \textit{top eigenvector}. Suppose that we are interested in the calculation of $\lim_N\mathbb E_{\bA}[F(u,\bv_{\mathrm{top}},\bx)]$. A strategy is to consider the following auxiliary partition function \cite{Susca19}
\begin{multline}
Z_\beta[\bA;tF]\coloneqq\\
\comprimi=\int\de \bv~\exp\left[\frac{\beta}{2}\langle\bv,\bA\bv\rangle+\beta tN F(u,\bv,\bx)\right] \delta\left(\|\bv\|^{2}-N\right)\label{eq:Z_vtop_multiplier} \ ,
\end{multline}
and compute the quantity of interest as
\begin{multline}
\lim_{N\to+\infty}\mathbb E_{\bA}[F(u,\bv_{\mathrm{top}},\bx)]\\
=\lim_{N\to+\infty}\lim_{\beta\to+\infty}\frac{1}{\beta N}\frac{\partial}{\partial t}\mathbb E_{\bA}[\ln Z_\beta[\bA;tF]]\Big|_{t=0}\\
=\lim_{\beta\to+\infty}\lim_{n\to 0}\lim_{N\to+\infty}\frac{1}{n\beta N}\frac{\partial}{\partial t}\ln \mathbb E_{\bA}\big[\big(Z_\beta[\bA;tF]\big)^n\big]\Big|_{t=0} \ ,\label{remarkable}
\end{multline}
where in the last line a replicated partition function appears and we assume again that one can exchange the limits $n\to0$ and $N \to \infty$. We remark that the partition function $Z_\beta(\bA)$ in Eq.~\eqref{eq:Z_lambdatop_soft} follows from setting $t=0$ in Eq.~\eqref{eq:Z_vtop_multiplier}.

This simple observation allows us to access a series of relevant observables. 
For example, we are interested in the asymptotic \textit{density of the entries of the top eigenvector} $\bv_{\mathrm{top}}$
\begin{equation}
\rho_{\rm top}(u)=\mathbb E_{\bA}\left[\frac{1}{N}\sum_{i=1}^{N}\delta\left(u-v_{\mathrm{top},i}\right)\right] \ .\label{eq:instance_density_top}
\end{equation}
This quantity can be computed via the aforementioned strategy by choosing for $F$ in Eq.~\eqref{remarkable} the function
\[F(u,\bv,\bx)\mapsto F_{\rm top}(u,\bv)\coloneqq\frac{1}{N}\sum_{i=1}^N\delta\left(u-v_{i}\right) \ .\]

Similarly, it is possible to compute the average \textit{overlap} between $\bv_{\mathrm{top}}$ and the spike $\bx$, by estimating first the density
\begin{equation}
    \rho_{\mathrm{ov}}(u) =
    \mathbb E_\bA\left[\frac{1}{N} \sum_{i=1}^N \delta(u - x_i v_{\mathrm{top},i}) \right] \ , \label{eq:instance_density_ov}
\end{equation}
with the choice
$$F(u,\bv,\bx)\mapsto F_{\rm ov}(u,\bv,\bx)\coloneqq \frac{1}{N}\sum_{i=1}^N\delta(u-x_iv_i) \ ,$$
and then computing
\begin{equation}
\lim_{N\to \infty}\mathbb E_{\bA}\left[\frac{\langle\bx,\bv_{\mathrm{top}}\rangle}{N}\right] =\lim_{N\to \infty}\int u\,\rho_{\mathrm{ov}}(u)\de u \ . \label{eq:overlap_calc}
\end{equation}

\section{\label{sec:results}Main results}

The discussion in Section~\ref{sec:formulation} showed that the knowledge of the average replicated partition function provides a significant amount of information on a series of relevant observables. We present the full calculation of a series of crucial quantities in Appendix~\ref{app:Afullreplica}. In this section we will summarise the main results.

Before proceeding, let us introduce a convenient auxiliary quantity and notation. Let $p_k$ be the degree distribution of the configuration model with truncated largest degree $k_{\mathrm{max}}$. We will denote 
\begin{equation}
r_k\coloneqq \frac{k p_k}{c},\qquad k\in\{1,\dots,k_{\rm max}\}
\label{eq:degreecorr}
\end{equation}
the associated \textit{degree-corrected degree distribution}. In the following, given a function $f$ defined on the positive integers, we will denote the average with respect to this distribution as
\[\llangle f(k)\rrangle\coloneqq\sum_{k=1}^{k_{\rm max}}r_kf(k) \ .\]

Moreover, in the following, we will use the shorthand notation $\{\de \mu\}_{s}\coloneqq\prod_{\ell=1}^s \mu(\bz_\ell)\de\bz_\ell$ for a given density $\mu(\bz)$ on a domain of $\R^d$, and the notation $\{u\}_s\coloneqq\sum_{\ell=1}^su_\ell$ for an indexed collection of scalar variables.

\begin{result}[Self-consistent equations and top eigenvalue]\label{res:RDE}
Let $X\sim\varrho_x$ be a random variable. Let  $\uppi(\omega,h)$ be the joint probability density of the pair $(\omega,h)$ with $\omega>0$, solution of the following recursive distributional equation 
\begin{subequations}
    \begin{multline} 
      \uppi(\omega,h) = \mathbb E_X\left\llangle\int \{\de  \uppi\}_{k-1} \{\de \rho_W\}_{k-1} \right. \\\comprimi\left.\times\comprimi \delta \left(\omega - \lambda_\theta+\left\{\frac{W^2}{\omega}\right\}_{k-1} \right)\delta \left(h - \left\{\frac{hW}{\omega}\right\}_{\mathclap{\quad k-1}}  - \ \theta q X \right)\right\rrangle \ , \label{eq:RDE}
    \end{multline}
supplemented with the following conditions that fix the parameters $\lambda_\theta$ and $q$
\begin{align}
    1 &=\comprimi \mathbb E_X\left\langle \int \{ \de  \uppi \}_k \{\de  \rho_W\}_{k} \left(\frac{\left\{\frac{hW}{\omega}\right\}_k + \theta q X}{\lambda_\theta - \left\{\frac{W^2}{\omega}\right\}_k}\right)^{2}\right\rangle \ , \label{eq:lambda_stat_pi_only} \\ 
    q &= \mathbb E_X \left\langle\int \{\de  \uppi\}_k \{\de \rho_W\}_k \frac{X \left\{\frac{hW}{\omega}\right\}_k+\theta q X^2}{\lambda_\theta - \left\{\frac{W^2}{\omega}\right\}_k }\right\rangle \ . \label{eq:C_pi}
\end{align}
\label{eq:RDEs}
\end{subequations}
Then the average top eigenvalue $\lambda_{\rm top}$ of the matrix $\bA$ is equal to the parameter $\lambda_\theta$ satisfying Eqs.~\eqref{eq:RDEs}, namely 
    \begin{equation}
        \mathbb E_{\bA} [\lambda_{\rm top}] = \lambda_\theta \ .
    \end{equation}
\end{result}
The numerical solution $\uppi(\omega,h)$ of Eqs.~\eqref{eq:RDEs} can be efficiently obtained using a population dynamics algorithm, as described in Appendix~\ref{app:popdyn}. The knowledge of $\uppi(\omega,h)$ satisfying Eqs.~\eqref{eq:RDEs} allows us to compute the top eigenvector properties in Eqs.~\eqref{eq:instance_density_top} and \eqref{eq:instance_density_ov}.
\begin{result}[Typical density of the top eigenvector components] The following holds
    \begin{multline}
        \rho_{\mathrm{top}}(u) =\\\comprimi= \mathbb E_X \left\langle\int \{\de  \uppi \}_k \{\de \rho_W\}_k \delta\left(u - \frac{ \left\{\frac{hW}{\omega}\right\}_k +\theta q X}{\lambda_\theta - \left\{\frac{W^2}{\omega}\right\}_k } \right)\right\rangle \ . \label{eq:top_eig_comp_density}
    \end{multline}
\end{result}

\begin{result}[Typical density of the overlap components] The typical overlap density is given by
    \begin{multline}
        \rho_{\mathrm{ov}}(u)= \\\comprimi=\mathbb{E}_X \left\langle\int \{\de  \uppi \}_k \{\de \rho_W\}_k \delta\left(u - \frac{X \left\{\frac{hW}{\omega}\right\}_k + \theta q X^2 }{\lambda_\theta - \left\{\frac{W^2}{\omega}\right\}_k } \right)\right\rangle \ .\label{eq:overlap_comp_density}  
    \end{multline}
\end{result}

\begin{remark}[Degree decomposition]
{Observe that the density of the top eigenvector components and of the overlap between the top eigenvector and the signal in Eqs.~\eqref{eq:top_eig_comp_density} and \eqref{eq:overlap_comp_density} are naturally decomposed as superpositions of contributions coming from nodes of different degrees (see Fig.~\ref{fig:K1_xstdGauss_c4} for explicit examples of such node decomposition). }
\end{remark}

Using {$\rho_{\mathrm{ov}}(u)$ from Eq.~\eqref{eq:overlap_comp_density}}, we easily obtain the average value of the overlap {between the top eigenvector and the signal}.
\begin{result}[Overlap] The overlap between the top eigenvector and the signal is given by
    \begin{align}
    \nonumber &\mathbb E_{\bA} \left[\frac{\langle\bx,\bv_{\mathrm{top}}\rangle}{N} \right] =  \int \de  u \ \rho_{\mathrm{ov}}(u) u = \\&\comprimi=\mathbb{E}_X\left\langle \int \{\de  \uppi \}_k \{\de \rho_W\}_k \ \frac{X \left\{\frac{hW}{\omega}\right\}_k + \theta q X^2}{\lambda_\theta - \left\{\frac{W^2}{\omega}\right\}_k } \right\rangle \label{eq:avg_overlap} \ .
    \end{align}
\end{result}

\begin{remark}[Interpretation of $q$] Note that, in light of Eq.~\eqref{eq:avg_overlap}, Eq.~\eqref{eq:C_pi} implies that $q$ can be interpreted as the average overlap between the top eigenvector and the signal.
\end{remark}

\begin{remark}[Sign of $q$]
Recall that the solution of the problem is a single vector $\bv_{\mathrm{top}}$ up to a $\Z_2$ symmetry, and the signal vector $\bx$ enters the model through $\bx \bx^\intercal$ and is identifiable up to a global sign. In the definition of the overlap, we fix the orientation of $\bv_{\mathrm{top}}$ such that 
\begin{align}
    \nonumber &\mathbb E_{\bA} \left[\frac{\langle\bx,\bv_{\mathrm{top}}\rangle}{N} \right] = q \geq 0 \ .
\end{align}
\end{remark}

\begin{remark}[Reduction to the pure noise matrix]
    Setting $\theta=0$ or $\varrho_x(x)=\delta(x)$ the spike in Eq.~\eqref{eq:model_definition} is removed and the results in Eq.~\eqref{eq:RDE}, \eqref{eq:lambda_stat_pi_only} and \eqref{eq:top_eig_comp_density} recover the properties of the top eigenpair of sparse graphs derived in \cite{Susca19}, as expected. 
\end{remark}

To conclude, we state the results for the critical value $\theta_{\mathrm{crit}}$ for a general degree distribution $p_k$. Let us assume that the spike is sampled from a distribution such that $\mathbb E_X[X]=0$ and $\sigma_x^2\coloneqq \mathbb E_X[X^2]$ finite. The critical value $\theta_{\rm crit}$ is such that for $\theta\leq\theta_{\mathrm{crit}}$ we have $\mathbb E_{\bA}[\lambda_{\rm top}]=\lambda_{\theta=0}$ (average top eigenvalue in absence of a signal), and the average overlap between the signal and the top eigenvector $\mathbb E_{\bA} \left[\frac{1}{N}\langle\bx,\bv_{\mathrm{top}}\rangle \right]=0$, indicating that the signal cannot be recovered by looking at the top eigenvector. For $\theta>\theta_{\mathrm{crit}}$, $\mathbb E_{\bA}[\lambda_{\rm top}]=\lambda_{\theta}$ and the average overlap $\mathbb E_{\bA} \left[\frac{1}{N}\langle\bx,\bv_{\mathrm{top}}\rangle \right]>0$, indicating that signal recovery is possible via PCA. See Fig.~\ref{fig:spectra} for an illustration of the mechanism. These results are derived in Appendix~\ref{app:sec:top_eigenvalue_BBP}.

\begin{result}[Recovery threshold] Under the assumption that $\mathbb E_X[X]=0$ and $\mathbb E_X[X^2]=\sigma_x^2$, the recovery threshold is given by
\begin{equation}
    \theta_{\mathrm{crit}} = \frac{1}{\sigma_x^2 Q(\lambda_{\theta=0})} \ ,\label{eq:theta_crit}
\end{equation}
where $\lambda_{\theta=0}$ is the structural top eigenvalue of the noise matrix $\bJ$, obtained by solving Eqs.~\eqref{eq:RDE} and \eqref{eq:lambda_stat_pi_only} with $\theta=0$ \footnote{Note that this case corresponds to the sparse symmetric matrix case analysed in Ref.~\cite{Susca19}.}, 
and the function
\begin{equation} \label{app:eq:C_xcenteredQ}
    Q(\lambda_\theta)\coloneqq\left\langle \int \{\de  \uppi\}_k \{\de \rho_W\}_k \frac{  1}{\lambda_{\theta} - \left\{\frac{W^2}{\omega}\right\}_k } \right\rangle 
\end{equation} 
appears.
\label{res:theta_crit}
\end{result}

\begin{figure*}
  \centering
    \includegraphics[height=0.32\linewidth]{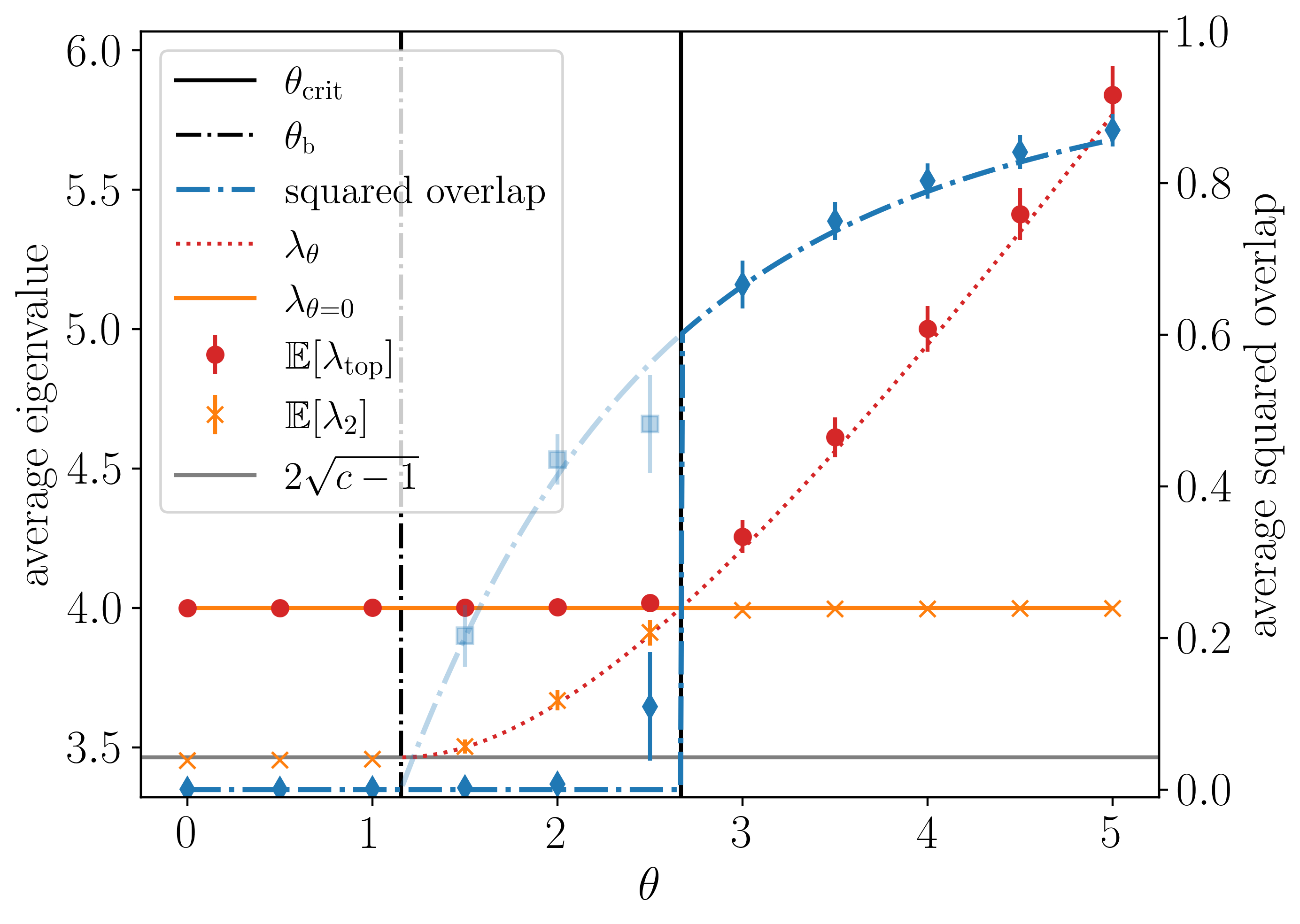}\hfill
    \includegraphics[height=0.32\linewidth]{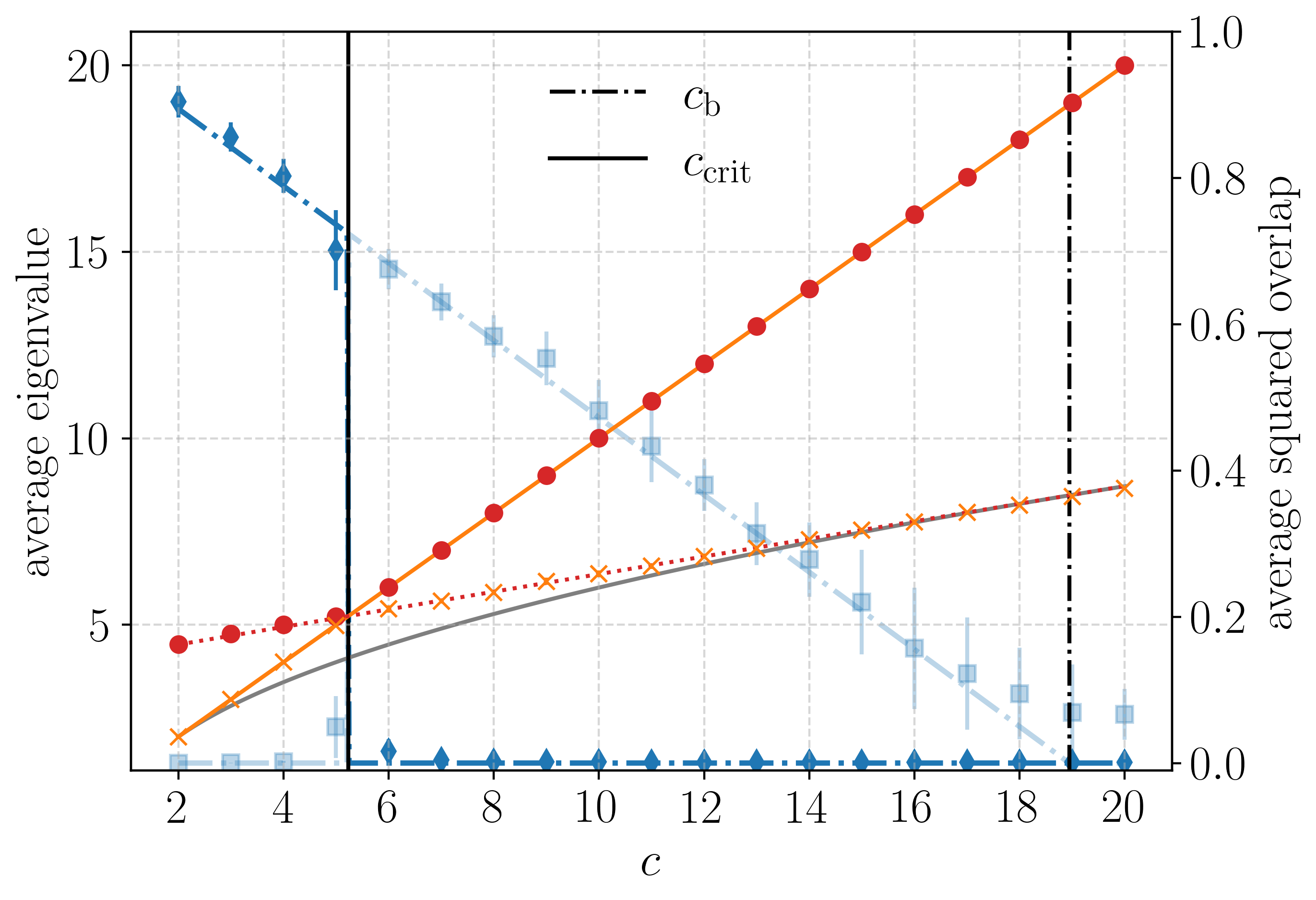}
  \caption{Random regular setup for $\bJ$ with $\rho_W(W)=\delta(W-1)$, $p_k=\delta_{k,c}$ and signal $x_i \sim \mathcal{N}(0,1)$. In the figures the threshold $\theta_{\mathrm{crit}}$  Eq.~\eqref{eq:RRG_theta_c} (black vertical line), the average top eigenvalue (dots) and second top eigenvalue (crosses) from direct diagonalisation for various signal strengths $\theta$ are given. For $\theta\leq\theta_{\mathrm{crit}}$, the structural outlier eigenvalue $\lambda_{\theta=0}=c$ of the symmetric noise matrix $\bJ$ (orange solid line) is the top eigenvalue of $\bA$. For $\theta>\theta_{\mathrm{crit}}$, the eigenvalue associated with the signal $\lambda_{\theta}$ dominates, as predicted by Eq.~\eqref{eq:RRG_lambdasignal} (red dotted line). The squared overlap in Eq.~\eqref{eq:RRG_overlap} (dark blue dashed line) matches the simulation results (diamonds) and jumps from zero to a nonzero value at $\theta_{\mathrm{crit}}$. At $\theta_\mathrm{b}$ Eq.~\eqref{eq:RRG_theta_b}, the eigenvalue associated with the signal $\lambda_\theta$ detaches from the bulk and is the second top eigenvalue for $\theta_\mathrm{b}<\theta<\theta_{\mathrm{crit}}$.
  The average overlap values from those experiments are reported on the left axis (blue diamonds). Notably, the formulas for the top eigenvalue and the squared overlap can be continued into the $\theta_{\rm b}<\theta\leq\theta_{\rm crit}$ phase (light blue curves): the squared overlap of the signal vector and the eigenvector associated with the \textit{second} largest eigenvalue obtained via direct diagonalisation (denoted by squares in the plot) follow this analytical prolongation. ({\it Left}) Results varying $\theta$ at fixed $c=4$. ({\it Right}) Results varying $c$ at fixed $\theta=4$. All simulation results are averaged over direct diagonalisation of $10$ matrices $\bA$ of size $N = 2\times10^3$, with standard deviation error bars.}
  \label{fig:RRG_lambdas_mu0}
\end{figure*}

Consider in fact Eq.~\eqref{eq:C_pi}, which for zero-centered spike components {can be rewritten as
\begin{equation} \label{app:eq:C_xcentered}
q(1- \theta \sigma_x^2 Q (\lambda_\theta))=0 \ .
\end{equation} 
}
{Eq.~\eqref{app:eq:C_xcentered} has two solutions: $q=0$, corresponding to zero overlap between the signal and top eigenvector, and $q\neq 0$ as long as $\theta\sigma_x^2Q(\lambda_\theta)=1$, which is the sparse analogue of the outlier condition for rank-one perturbations \cite{BenaychGeorges11}. The critical threshold $\theta_{\mathrm{crit}}$ has to mark the onset of a nonzero overlap, so it should be the marginal value satisfying $\theta_{\mathrm{crit}}\sigma_x^2Q(\lambda_{\theta=0})=1$ for the top eigenvalue of $\bJ$ in the noiseless case, leading to Eq. \eqref{eq:theta_crit} (see Fig.~\ref{fig:RRG_lambdas_mu0} for an illustration)}. 

\begin{remark}[Dense limit]
    In the large average connectivity limit $c \to \infty$, provided that $\frac{\langle k^2 \rangle - \langle k \rangle^2}{\langle k \rangle^2} \to 0$, the critical value $\theta_{\mathrm{crit}}$ in Eq.~\eqref{eq:theta_crit} converges to $\frac{1}{\sigma_x^2}$, the squared overlap to $\sigma_x^2 - (\theta \sigma_x)^{-2}$ and the top eigenvalue to $\theta \sigma_x^2 + \frac{1}{\theta \sigma_x^2}$, in agreement with the classical BBP result for dense real Wigner matrices \cite{Peche06,Capitaine09,Bloemendal_2012}. This is shown in Appendix~\ref{sec:denselimit}.
\end{remark}

\section{Special cases}\label{sec:discussion}
In this section we will specialise the results in Eq.~\eqref{eq:theta_crit} to the cases of the RR setup specified by Eq.~\eqref{eq:RR}, which will be treated exactly, and the Poissonian setup given via Eq.~\eqref{eq:ER}, where we will perform an approximation whose quality is quantified in Appendix~\ref{app:popdyn}. In particular, we are able to obtain more explicit expressions, respectively exact and approximate, for $\theta_{\mathrm{crit}}, \theta_{\rm b}, \mathbb E_{\bA}[\lambda_{\rm top}]$ and the typical squared overlap, which we can therefore estimate much faster than running a population dynamics algorithm. We will compare our prediction with the results of numerical simulations.

\subsection{The RR setup}
In the RR setup the following holds. The results obtained for this case are exact.
\begin{result}[Random regular setup]\label{res:RRG}
Assume $p_k$ has the form $p_k=\delta_{k,c}$, with $c>2$, and $\rho_W(W)=\delta(W-1)$. 
The critical value of the signal strength from Eq.~\eqref{eq:theta_crit} reads 
\begin{equation}
    \theta_{\mathrm{crit}} = \frac{c(c-2)}{\sigma_x^2 (c-1)} \ . \label{eq:RRG_theta_c}
\end{equation}
Moreover, for $\theta=\theta_{\mathrm{b}} $ with
\begin{equation}
    \theta_{\mathrm{b}} = \frac{c-2}{\sigma_x^2 \sqrt{c-1}} \leq \theta_{\mathrm{crit}}  \label{eq:RRG_theta_b}
\end{equation}
the eigenvalue associated with the signal detaches from the right side of the eigenvalue bulk at $\lambda_{\mathrm{b}} = 2 \sqrt{c-1}$.
For $\theta>\theta_{\mathrm{b}}$, the typical value of the outlier eigenvalue associated to the signal is
\begin{equation}
\lambda_\theta = \frac{c\sqrt{(\theta \sigma_x^2)^2 + 4} - (c-2)\theta \sigma_x^2}{2} \ . \label{eq:RRG_lambdasignal}
\end{equation}  

For $\theta>\theta_{\mathrm{crit}}$, the typical value of the top eigenvalue is
\begin{equation}
\mathbb E_{\bA} [\lambda_{\rm top}] = \lambda_\theta
\end{equation}  
and the typical squared overlap between the signal and the top eigenvector is nonzero and equal to
\begin{equation} \label{eq:RRG_overlap}
    \lim_{N\to \infty}\mathbb E_{\bA}\left[\frac{\langle\bx,\bv_{\mathrm{top}}\rangle^2}{N^2}\right] = \frac{c \theta \sigma_x^4}{2\sqrt{(\theta \sigma_x^2)^2 + 4}} - \frac{(c-2) \sigma_x^2}{2} \ .
\end{equation}

For $\theta\leq \theta_{\mathrm{crit}}$, the top eigenvalue is $\mathbb E_\bA [\lambda_{\rm top}]=c$ (the structural eigenvalue of a random $c$-regular graph) and the squared overlap in Eq.~\eqref{eq:RRG_overlap} is zero. 

For $\theta_{\rm b}<\theta\leq\theta_{\mathrm{crit}}$, the signal-bearing eigenvalue is the second-largest and an outlier, corresponding to the centre panel of Fig.~\ref{fig:spectra}. Then, the typical second largest eigenvalue is equal to $\mathbb E_{\bA} [\lambda_2]=\lambda_\theta<\mathbb E_\bA [\lambda_{\rm top}]=c$ in Eq.~\eqref{eq:RRG_lambdasignal} and the typical squared overlap between the signal and eigenvector associated to the second (signal-bearing) largest eigenvalue, $\lim_{N\to\infty} \mathbb E_{\bA}[\langle \bx, \bv_2 \rangle^2/N^2]$ is equal to Eq.~\eqref{eq:RRG_overlap}.

\end{result}

We discuss Result \ref{res:RRG} for a noise structure $\bJ$ taken as a pure adjacency matrix of a random $c$-regular graph, i.e., by picking $W_{ij}=1$.
Fig.~\ref{fig:RRG_lambdas_mu0} shows that the predictions for the average top eigenvalue, the transition values and the typical squared overlap are in good agreement with direct diagonalisation results from large matrices for standard Gaussian-distributed spike and various values of $\theta$ and $c$. In particular, the left panel of Fig.~\ref{fig:RRG_lambdas_mu0} shows that, for $c=4$, as the signal strength $\theta$ grows, the eigenvalue associated with the signal characterised by the formula in Eq.~\eqref{eq:RRG_lambdasignal} detaches from the bulk at $\theta_\mathrm{b}$ in Eq.~\eqref{eq:RRG_theta_b} and overtakes the structural eigenvalue $\lambda_{\theta=0}=c$ at $\theta_\mathrm{crit}$ in  Eq.~\eqref{eq:RRG_theta_c}, where for $\theta>\theta_\mathrm{crit}$ the squared overlap between the top eigenvector and signal becomes nonzero and is predicted by Eq.~\eqref{eq:RRG_overlap}, reported as a dark blue dash-dot line. We also report, in a light blue dash-dot line, the value of the squared overlap as predicted by Eq.~\eqref{eq:RRG_overlap} as $\theta_{\rm b}<\theta\leq\theta_{\rm crit}$. It corresponds to the scenario described in the central panel of Fig.~\ref{fig:spectra}: the signal strength is such that the eigenvalue associated to the signal is an outlier but is smaller than the leading eigenvalue associated to the noise $\lambda_{\theta=0}$, i.e., it is the second top eigenvalue of $\mathbf A$. We show therefore that Eq.~\eqref{eq:RRG_overlap} accurately predicts the average squared overlap of the signal vector and the eigenvector associated to the \textit{second} largest eigenvalue, obtained via direct diagonalisation. This is an explicit case where the top eigenpair analysis leads to closed-form expressions of spectral properties of the signal-bearing eigenpair as soon as the associated eigenvalue exits the bulk, i.e., for $\theta\geq\theta_{\rm b}$, and therefore becomes spectrally `visible'. This contrasts with the classical BBP intuition, where the signal eigenvalue is typically informative only when it is the top eigenvalue as $\theta>\theta_{\rm crit}$.
\begin{figure}
  \centering
    \includegraphics[width=\linewidth]{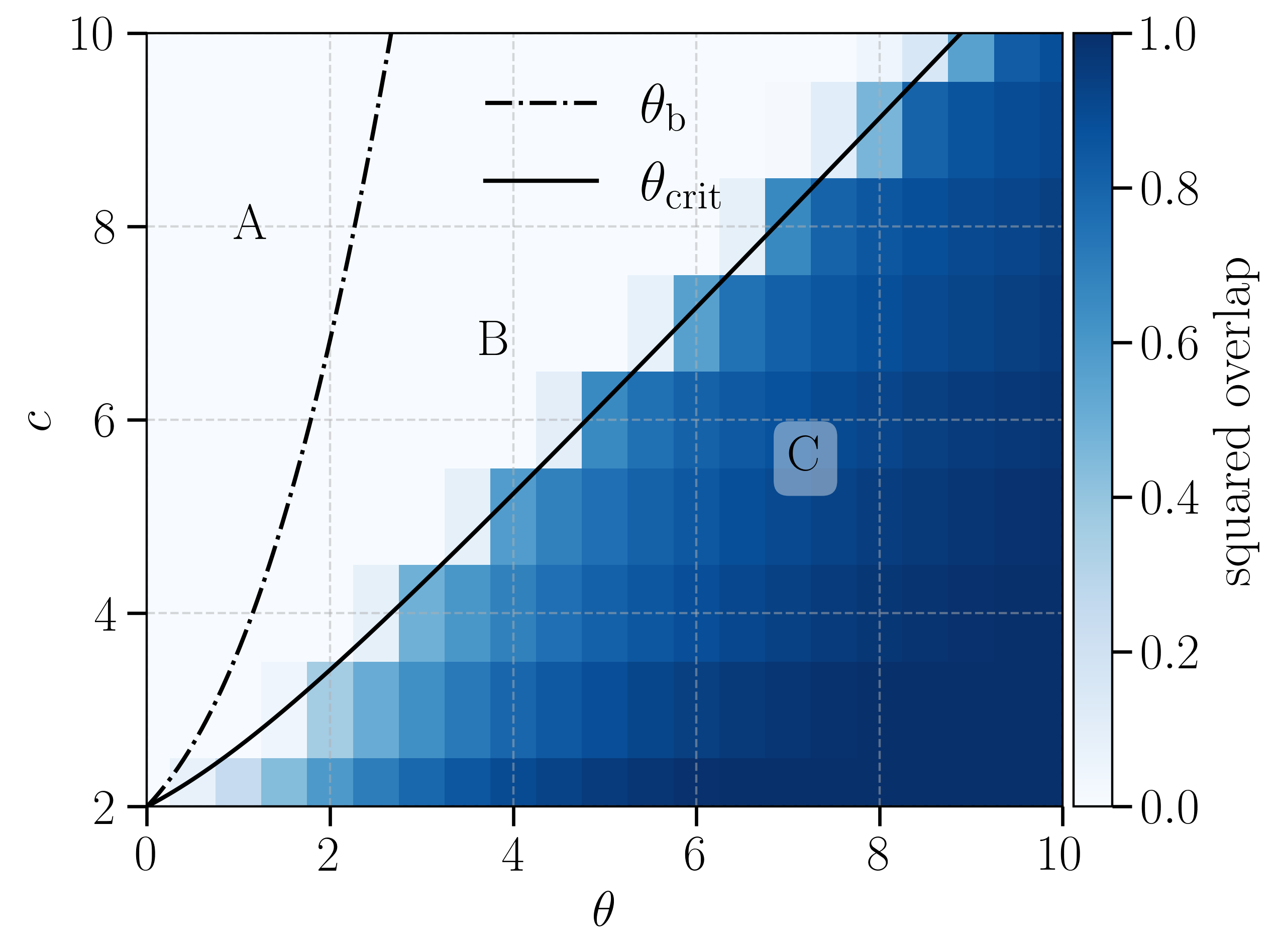} 
  \caption{Random regular setup with $\rho_W(W)=\delta(W-1)$ and signal $x_i \sim \mathcal{N}(0,1)$. The average squared overlap is given in the $(\theta,c)$ parameter space as obtained via numerical diagonalisation of five instances of size $N=10^3$. The dashed line corresponds to $\theta_\mathrm{b}$ in Eq.~\eqref{eq:RRG_theta_b}, where the signal eigenvalue detaches from the bulk. The solid line shows $\theta_\mathrm{crit}$ in Eq.~\eqref{eq:RRG_theta_c}, where the signal eigenvalue becomes the top eigenvalue.}
  \label{fig:RRG_lambdas_mu0_phase}\vspace{-0.5cm}
\end{figure}

\begin{figure*}
    \centering
    \includegraphics[height=0.35\linewidth]{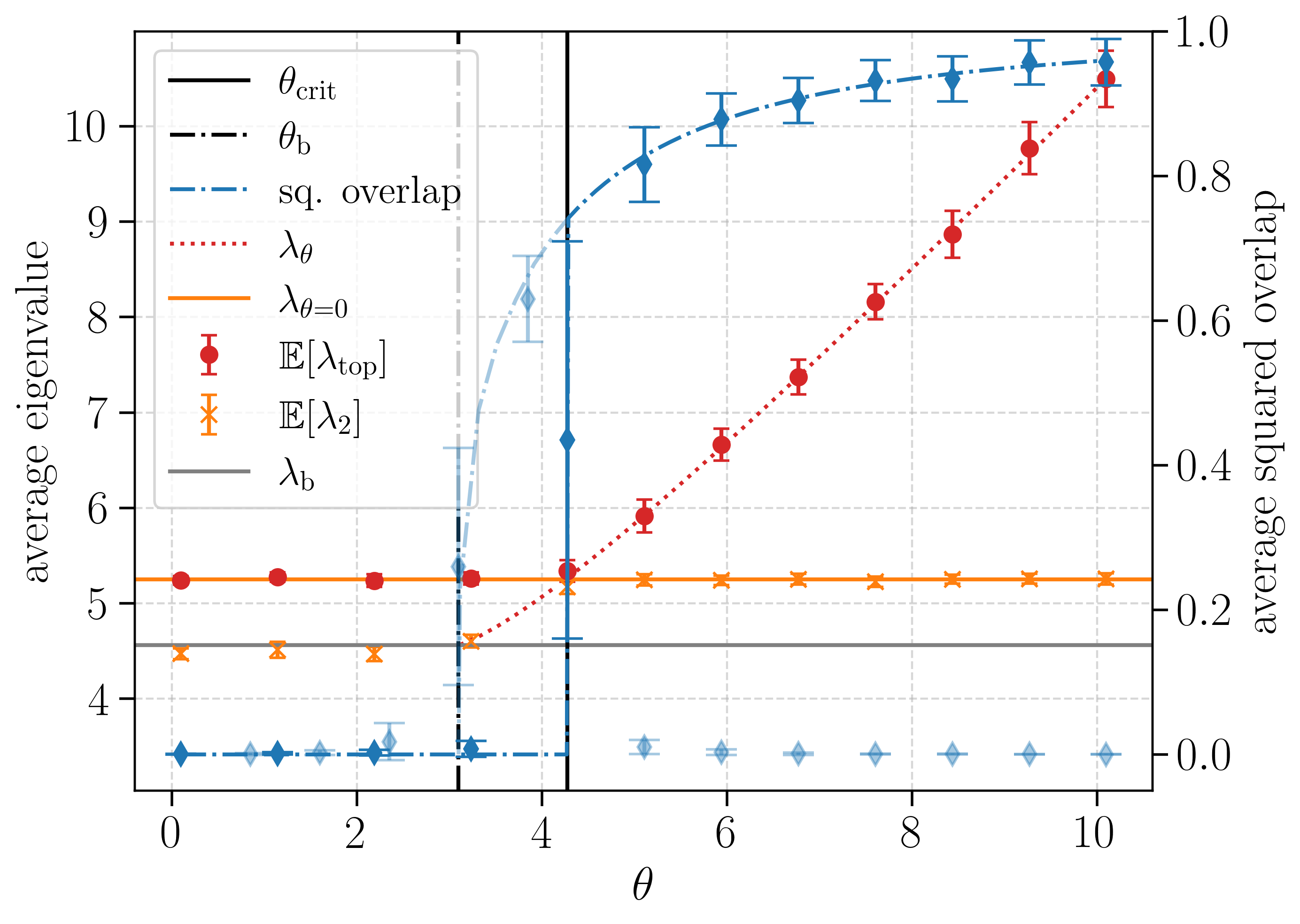}
\hfill
    \includegraphics[height=0.35\linewidth]{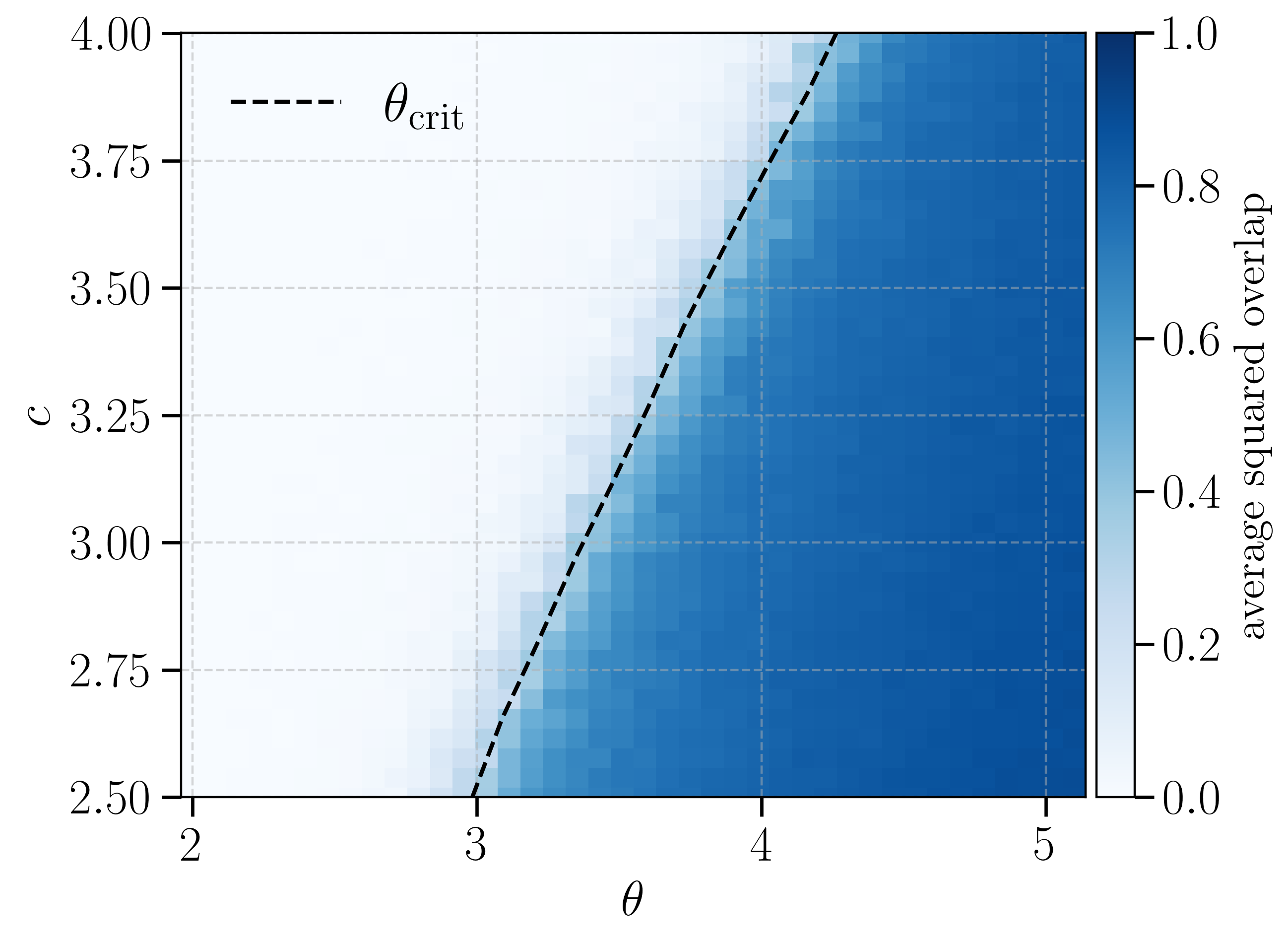} 
  \caption{Poissonian setup with $k_{\mathrm{max}}=20$, $c=4$, $\rho_W(W)=\delta(W-1)$ and signal $x_i \sim \mathcal{N}(0,1)$. All simulation results are averaged over direct diagonalisation of $25$ matrices $\bA$ of size $N = 2\times10^3$ ({\it left}) and $50$ matrices of size $N = 2\times10^3$ ({\it right}), with standard deviation error bars. Theoretical predictions are obtained using the analytical formulas in Result~\ref{res:ER}. ({\it Left})
  Average top eigenvalue (dots) and second top eigenvalue (crosses) from direct diagonalisation for various signal strengths $\theta$. For $\theta\leq\theta_{\mathrm{crit}}$, $\mathbb E[\lambda_{\rm top}]$ (dots) coincides with the structural top eigenvalue $\lambda_{\theta=0}$ of the symmetric noise matrix $\bJ$ (orange solid line). For $\theta>\theta_{\mathrm{crit}}$, the top eigenvalue correlates with the signal and its average takes the value $\lambda_{\theta}$, as predicted by Eq.~\eqref{eq:gen_lambda_signal} (red dotted line). The squared overlap in Eq.~\eqref{eq:gen_overlap} (dark blue dashed line) also matches the simulation results (diamonds) and exhibits a discontinuity at $\theta_{\mathrm{crit}}$. For $\theta_{\rm b}<\theta\le\theta_{\mathrm{crit}}$, corresponding to Fig.~\ref{fig:spectra} (centre), the signal eigenvalue is an outlier but only the second largest eigenvalue of $\bA$. In this window, the light blue diamonds show the measured squared overlap between the signal and the second largest eigenvector of $\bA$. The light blue dashed line is the continuation of Eq.~\eqref{eq:gen_overlap} into this window, shown \emph{only as a guide to the eye}: unlike the RR case (Fig.~\ref{fig:RRG_lambdas_mu0}), it relies on an approximation. The same applies to $\lambda_{\rm b}$ and $\theta_{\rm b}$ (see Remark~\ref{rem:poisson_validity}).} ({\it Right}) Heatmap of the average overlap in the $(\theta,c)$ parameter space obtained via numerical diagonalisation. The transition value $\theta_{\mathrm{crit}}$ Eq.~\eqref{eq:gen_transition} (black dashed line) clearly separates the non-recovery and recovery phases.
  \label{fig:ER_transition}
\end{figure*}

On the right side of Fig.~\ref{fig:RRG_lambdas_mu0} we fix $\theta=4$ and observe that for small noise density $c$ the signal eigenvalue $\lambda_{\theta}$ dominates. As $c$ grows, it is overtaken by $\lambda_{\theta=0} = c$ with the squared overlap between the top eigenvector and signal in Eq.~\eqref{eq:RRG_overlap} becomes zero at $$c_{\mathrm{crit}}=\frac{2+\theta\sigma_x^2+\sqrt{4+\theta^2\sigma_x^4}}{2} \ ,$$ and $\lambda_\theta$ is absorbed by the bulk at $$c_{\rm b} = \frac{1}{4} \left( \theta \sigma_x^2 + \sqrt{\theta^2 \sigma_x^4 +4} \right)^2 + 1 \ .$$ 
Note that our formula for the squared overlap in Eq.~\eqref{eq:RRG_overlap} also predicts the value of the squared overlap of the signal vector and the eigenvector associated to the \textit{second} largest eigenvalue for values $c_{\rm crit}<c<c_{\rm b}$, indicated by a light blue dash-dot line.

Finally, Fig.~\ref{fig:RRG_lambdas_mu0_phase} shows a phase diagram in the $(\theta,c)$ parameter space with signal $\bx\sim\mathcal{N}(\mathbf 0,\mathbf I_N)$. We show the squared overlap computed using direct diagonalisation of large matrices.
In phase A, the top eigenvalue associated with the signal is buried in the bulk, whereas in phase B it has detached from the bulk but remains smaller than the structural eigenvalue at $\mathbb E_{\bA} [\lambda_{\rm top}] = c$. 
Finally, in phase C the eigenvalue associated with the signal is the top eigenvalue predicted by Eq.~\eqref{eq:RRG_lambdasignal} and the squared overlap of the top eigenvector with the signal is of order one.

\subsection{The Poissonian setup}
Relevant quantities can be also specialised in the Poissonian setup via a suitable approximation. Let us first define an auxiliary quantity that is essential to derive the more explicit equations for special cases of degree distributions, namely, let $\tilde m(\lambda)$ be the solution of the self-consistency equation
\begin{equation}
    \tilde m(\lambda) = \left\llangle \frac{1}{\lambda - (k-1) \mathbb E_W[W^2] \tilde m(\lambda)}\right\rrangle \ .  \label{eq:m_lambda}
\end{equation}
In the following we will approximate $m(\lambda) \simeq \tilde m(\lambda)$ to compute the observables in the recovery phase for fluctuating degrees. The derivation and a more detailed discussion of the approximation is in Appendix~\ref{app:sec:top_eigenvalue_BBP}. 
The comparison of this approximation to the population dynamics estimation (which is exact) is in Appendix~\ref{app:popdyn}.

\begin{result}[Poissonian setup]
    Assume that $p_k$ has the form of a truncated Poisson distribution in Eq.~\eqref{eq:ER}. Within the approximation $m(\lambda) \simeq\tilde m(\lambda) $, $\tilde m(\lambda)$ being the solution of the self-consistency equation in Eq.~\eqref{eq:m_lambda}, the function $Q$ in Eq.~\eqref{app:eq:C_xcenteredQ} has the explicit expression 
    \begin{equation}\label{eq:Q_m}
    \comprimi \tilde Q(\lambda_\theta) \simeq\frac{c}{\bar c}m(\lambda_\theta)+\frac{\bar c^{k_{\mathrm{max}}}}{\Gamma k_{\mathrm{max}}!}\frac{1}{\lambda_\theta - k_{\mathrm{max}} \mathbb E_W[W^2] m(\lambda_\theta)}\ ,
    \end{equation} 
    and the critical value of the signal strength from Eq.~\eqref{eq:theta_crit} reads 
    \begin{equation}
        \theta_{\mathrm{crit}} \simeq \frac{1}{\sigma_x^2 \tilde  Q(\lambda_{\theta=0})} \ .\label{eq:gen_transition}
    \end{equation}
    For $\theta>\theta_{\mathrm{crit}}$, the typical value of the eigenvalue associated with the signal is equal to
    \begin{equation}
        \lambda_\theta \simeq \tilde  Q^{-1}\left( \frac{1}{\theta \sigma_x^2} \right) \ , \label{eq:gen_lambda_signal_all}
    \end{equation}
    and the typical value of the top eigenvalue is
    \begin{equation}
        \mathbb E_{\bA} [\lambda_{\rm top}] = \lambda_\theta \ , \label{eq:gen_lambda_signal}
    \end{equation}
    while the typical squared overlap between the signal and the top eigenvector is nonzero and equal to
    \begin{equation} \label{eq:gen_overlap}
        \lim_{N\to \infty}\mathbb E_{\bA}\left[\frac{\langle\bx,\bv_{\mathrm{top}}\rangle^2}{N^2}\right] \simeq -\frac{1}{\sigma_x^2 \theta^2 \tilde  Q'(\lambda_{\theta})} \ .
    \end{equation}
    This scenario corresponds to Fig.~\ref{fig:spectra} (right).

    For $\theta \leq \theta_{\mathrm{crit}}$,
    \begin{equation}
        \mathbb E_{\bA} [\lambda_{\rm top}] = \lambda_{\theta=0} \label{eq:gen_lambda_struct}
    \end{equation}
    and the squared overlap between the signal and the top eigenvector is zero.
    This scenario corresponds to the left and centre panels of Fig.~\ref{fig:spectra}.
    \label{res:ER}
\end{result}

\begin{remark}[The intermediate regime $\theta_{\rm b}<\theta\le\theta_{\rm crit}$]
    Unlike the RR case, the results above rely on the closure in Eq.~\eqref{eq:m_lambda}, which neglects the variance of $1/\omega$ induced by degree fluctuations; this approximation is discussed in more detail in Appendix~\ref{app:sec:general_top_eigenvalue_BBP}, and its remarkable accuracy in the discussed setup is quantified in Appendix~\ref{app:popdyn}. Direct diagonalisation (Fig.~\ref{fig:ER_transition}) shows the same three-phase scenario as in the RR case: the signal eigenvalue detaches from the bulk at some $\theta_{\rm b}<\theta_{\rm crit}$ and, for $\theta_{\rm b}<\theta\le\theta_{\rm crit}$, is the second largest eigenvalue, with the associated eigenvector correlating with the signal. By analogy with the RR case we expect, in the scenario corresponding to Fig.~\ref{fig:spectra} (centre), that at $\theta=\theta_{\rm b}$ with
    \begin{equation}
    \theta_{\mathrm{b}} \simeq \frac{1}{\sigma_x^2 \tilde Q(\lambda_{\rm b})} \ , \label{eq:theta_b_poisson}
    \end{equation}
    the eigenvalue associated with the signal $\lambda_\theta$ detaches from the right bulk edge at $\lambda_{\rm b}$. 
    For $\theta_{\rm b}<\theta\leq\theta_{\mathrm{crit}}$, the signal-bearing eigenvalue is the second-largest and an outlier. Then, the typical second largest eigenvalue is approximated by $\mathbb E_{\bA} [\lambda_2]\simeq\lambda_\theta<\mathbb E_\bA [\lambda_{\rm top}] = \lambda_{\theta=0}$ in Eq.~\eqref{eq:gen_lambda_signal_all} and the typical squared overlap between the signal and eigenvector associated to the second (signal-bearing) largest eigenvalue is approximated via Eq.~\eqref{eq:gen_overlap}, that is,
    \begin{equation} 
        \lim_{N\to \infty}\mathbb E_{\bA}\left[\frac{\langle\bx,\bv_{2}\rangle^2}{N^2}\right] \simeq -\frac{1}{\sigma_x^2 \theta^2 \tilde  Q'(\lambda_{\theta})} \ .
    \end{equation}
    However, unlike in the RR case, these expressions are not under analytical control, and we therefore show the continuation of Eqs.~\eqref{eq:gen_lambda_signal_all} and~\eqref{eq:gen_overlap} into this window, together with $\lambda_{\rm b}$ and $\theta_{\rm b}$, in Fig.~\ref{fig:ER_transition} only as a guide to the eye, motivated by the analogy with the RR case. The good agreement with direct diagonalisation is suggestive, but it is not derived. A controlled analytical continuation into $\theta_{\rm b}<\theta\le\theta_{\rm crit}$ would require a careful treatment of the second largest eigenpair statistics along the lines of Ref.~\cite{Susca21}, which we leave for future work.
    \label{rem:poisson_validity}
\end{remark}

\begin{figure}
    \includegraphics[height=0.55\linewidth]{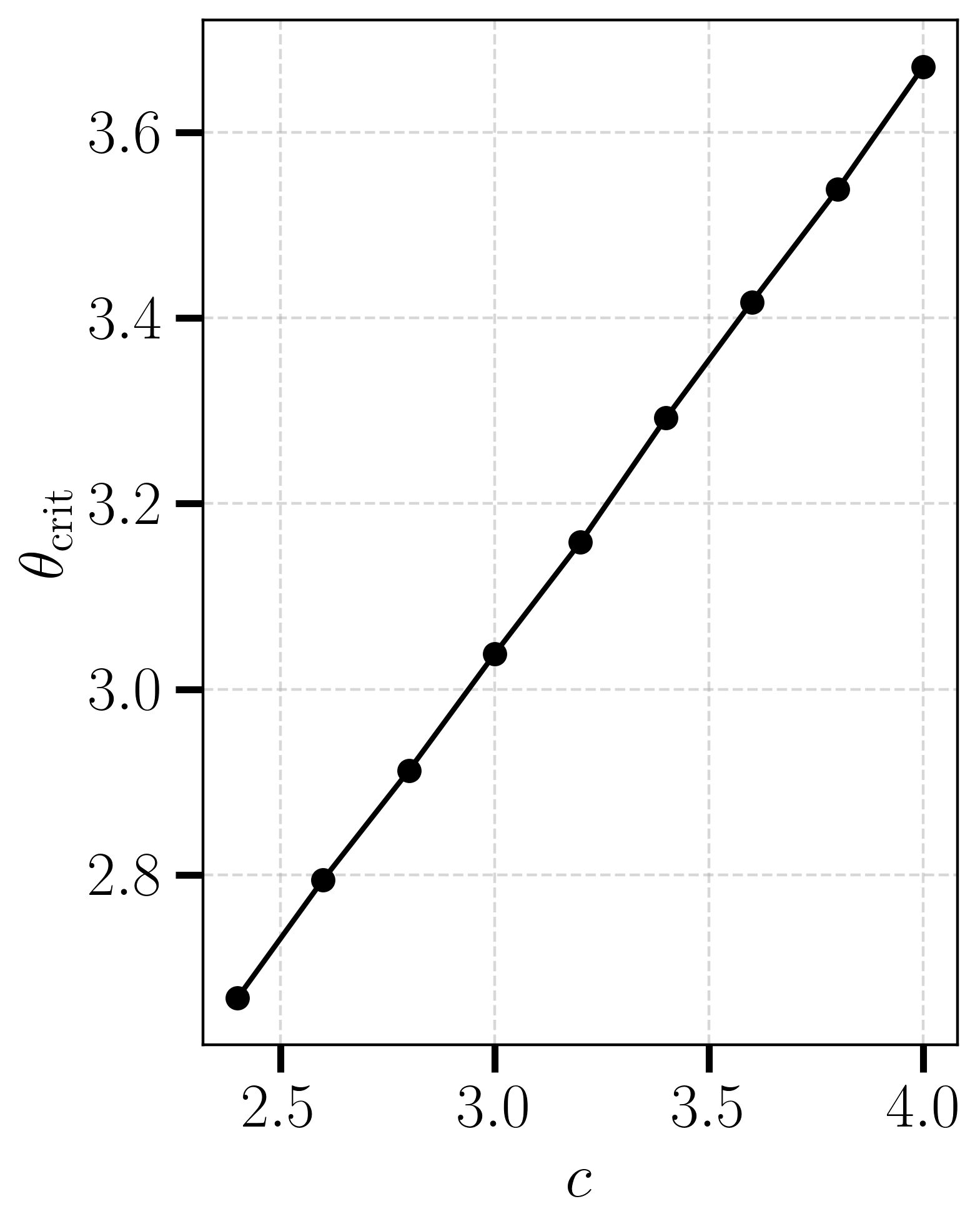} 
    \includegraphics[height=0.55\linewidth]{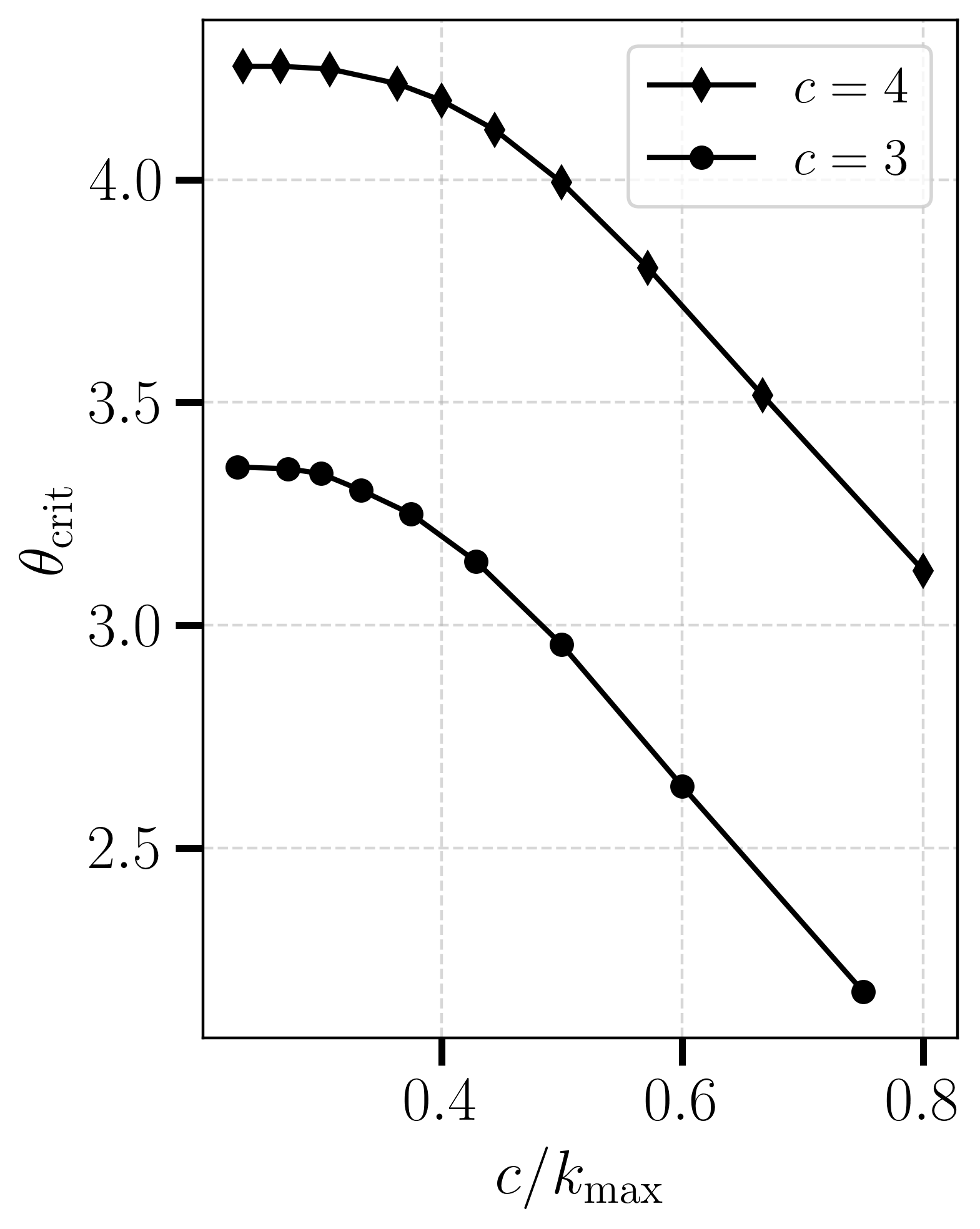} 
  \caption{Poissonian setup with $\rho_W(W)=\delta(W-1)$. The $\lambda_{\mathrm{\theta=0}}$ parameter is obtained via population dynamics for $\theta=0$ with population size $N_p=10^5$. ({\it Left}) The critical threshold $\theta_{\mathrm{crit}}$ as in Eq.~\eqref{eq:gen_transition} for $k_{\max}=6$. ({\it Right}) The critical threshold $\theta_{\mathrm{crit}}$ as in Eq.~\eqref{eq:gen_transition} as a function of the ratio of the average degree $c=3,4$ and $k_{\max}>c$.} \label{fig:K1_xstdGauss_theta_crit}
\end{figure}

\begin{figure}
    \includegraphics[height=0.55\linewidth]{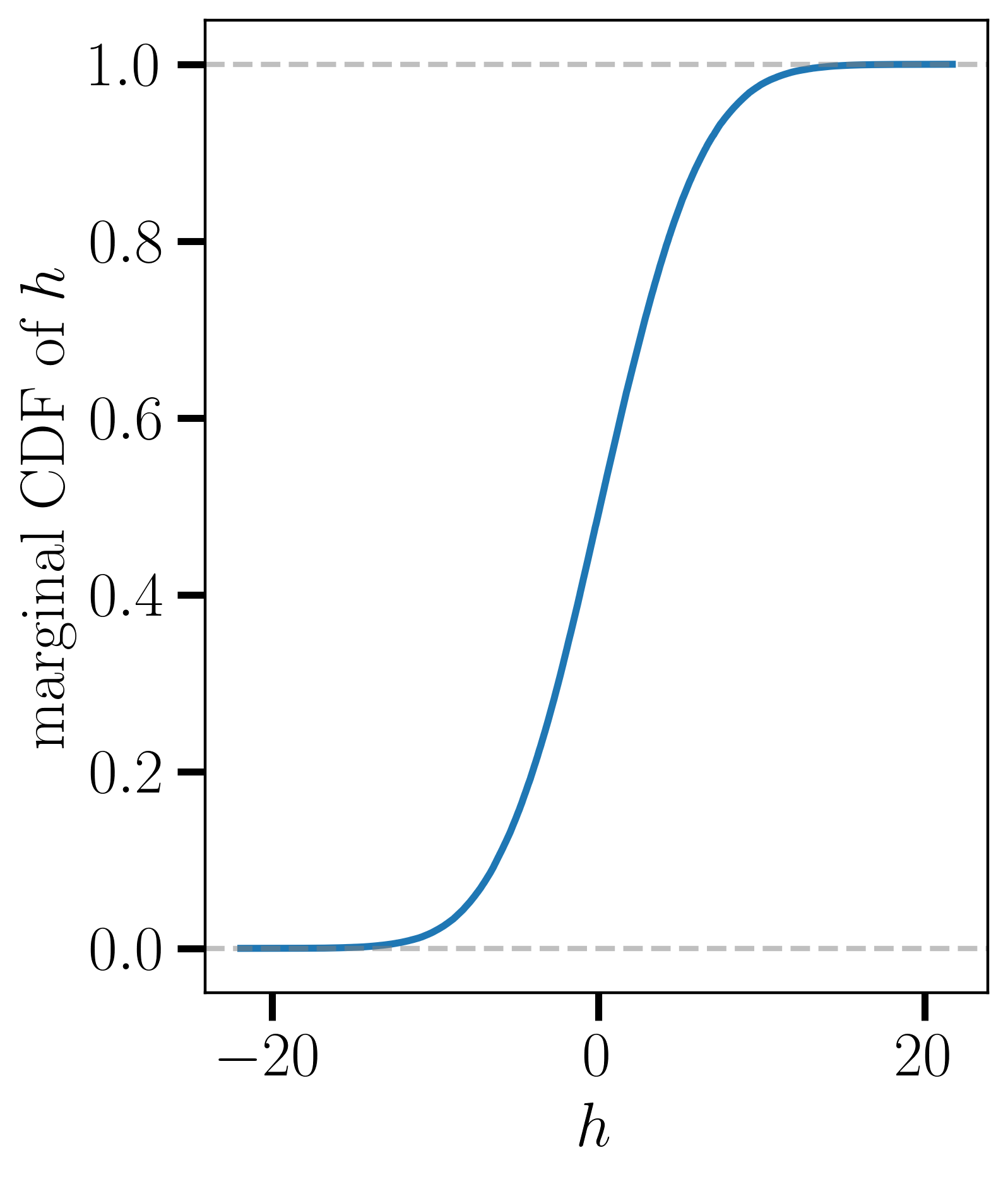}
    \includegraphics[height=0.55\linewidth]{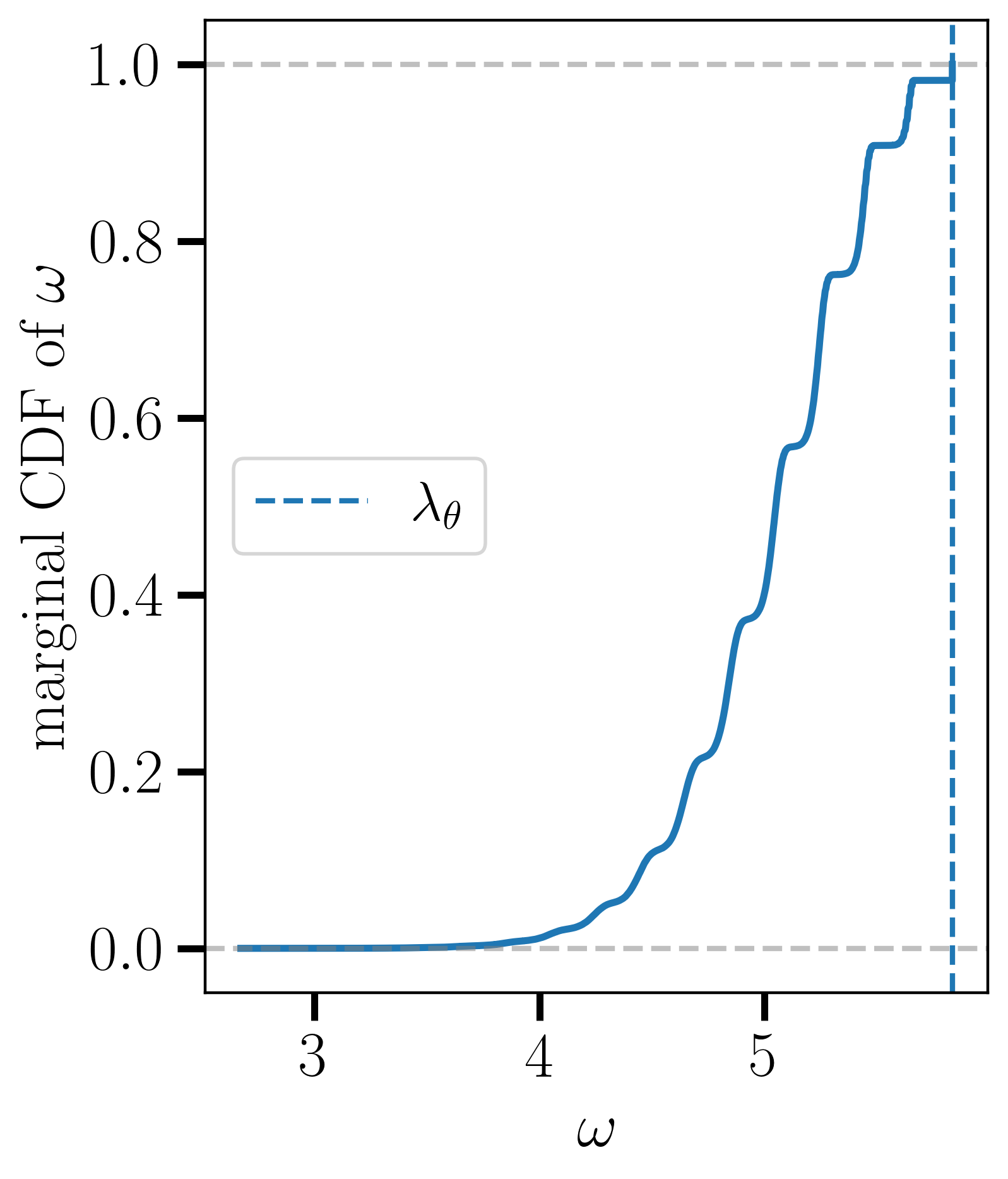} 
  \caption{Poissonian setup with $k_{\mathrm{max}}=20$, $c=4$, $\rho_W(W)=\delta(W-1)$ and signal $x_i \sim \mathcal{N}(0,1)$ of strength $\theta=5$. The marginal cumulative distribution functions of $h$ and $\omega$ are obtained via population dynamics with population size $N_p=2\times 10^5$.  ({\it Left}) The marginal cumulative distribution function of the single-site bias fields $h$. ({\it Right}) The marginal cumulative distribution function  of the single-site inverse variances $\omega$, where the largest value of $\omega$ corresponds to the degree $k=1$ and is equal to $\mathbb E_{\bA} [\lambda_{\rm top}]=\lambda_\theta$. } \label{fig:K1_xstdGauss_c4_marginals}
\end{figure}

\begin{figure*}
    \centering
    \includegraphics[width=0.43\linewidth]{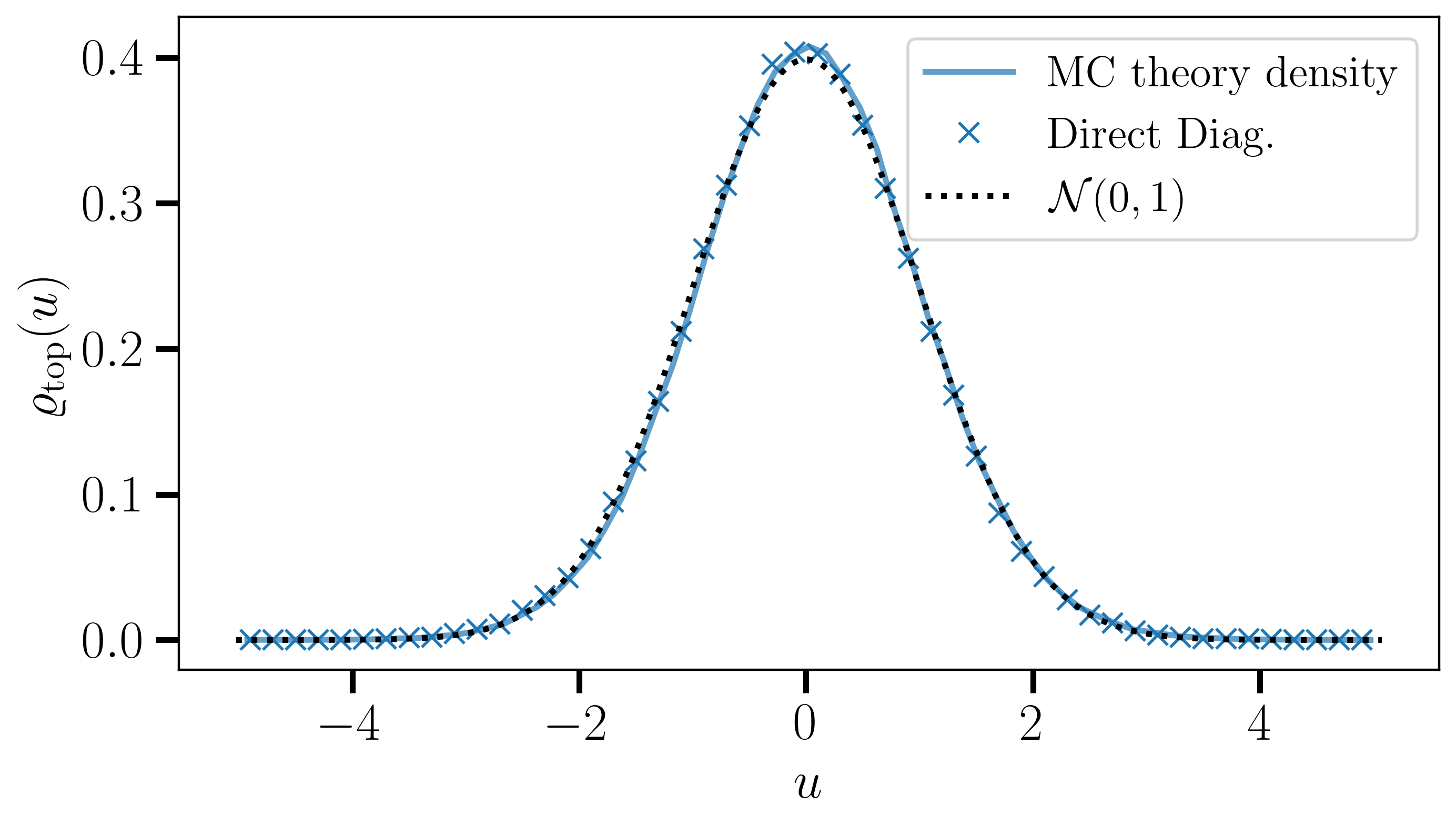}
    \includegraphics[width=0.43\linewidth]{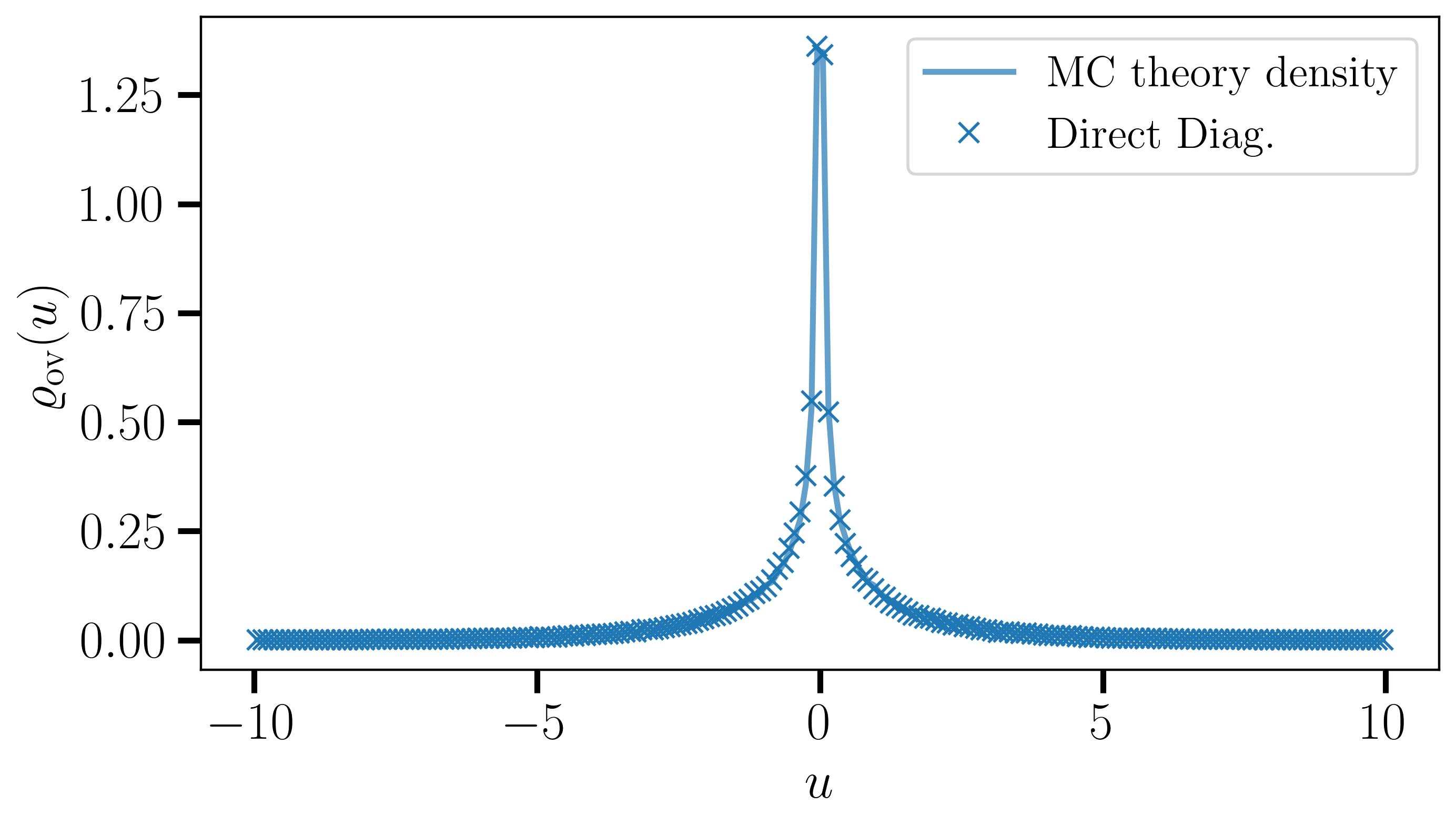} \\
    \includegraphics[width=0.43\linewidth]{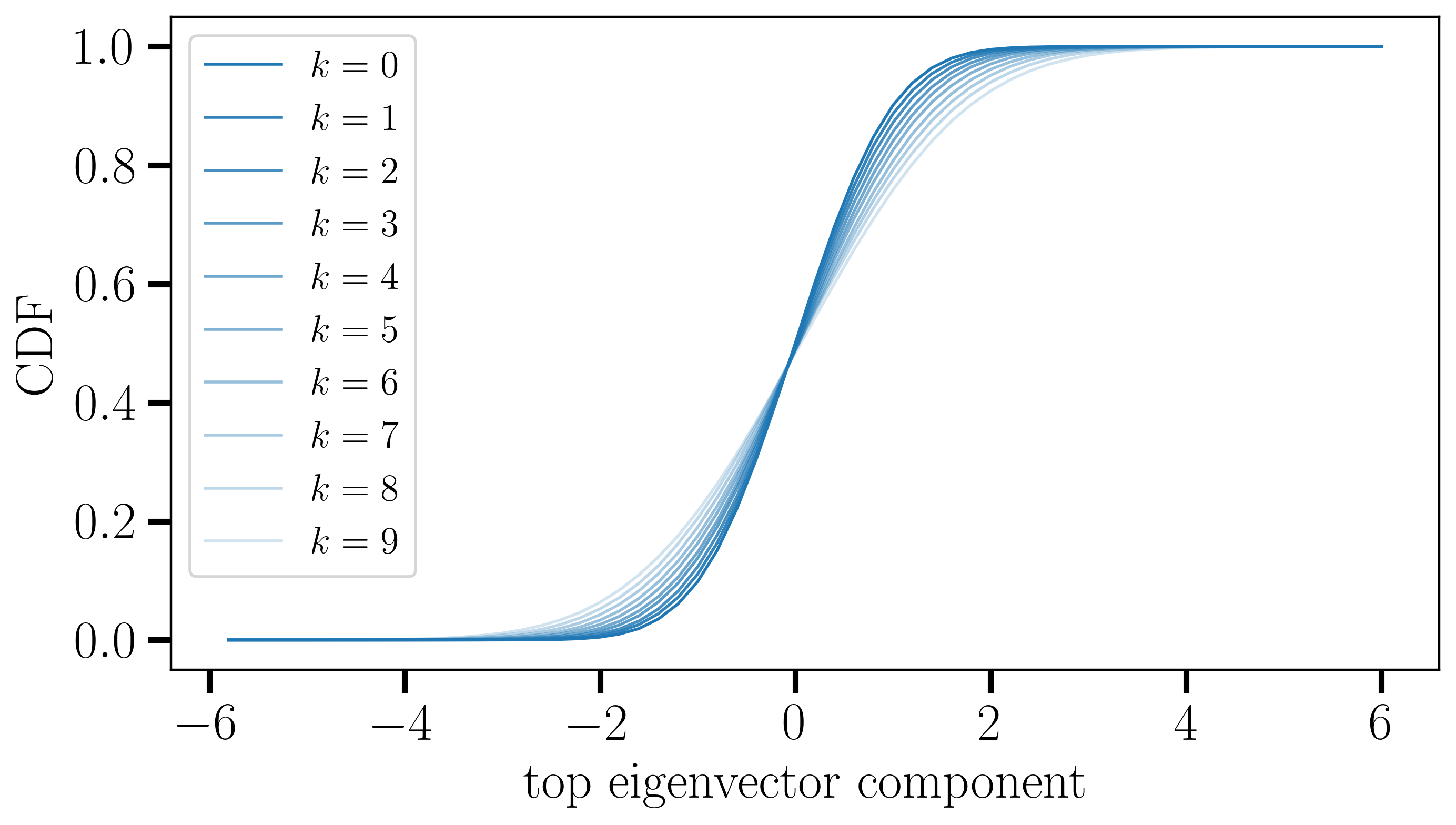} 
    \includegraphics[width=0.43\linewidth]{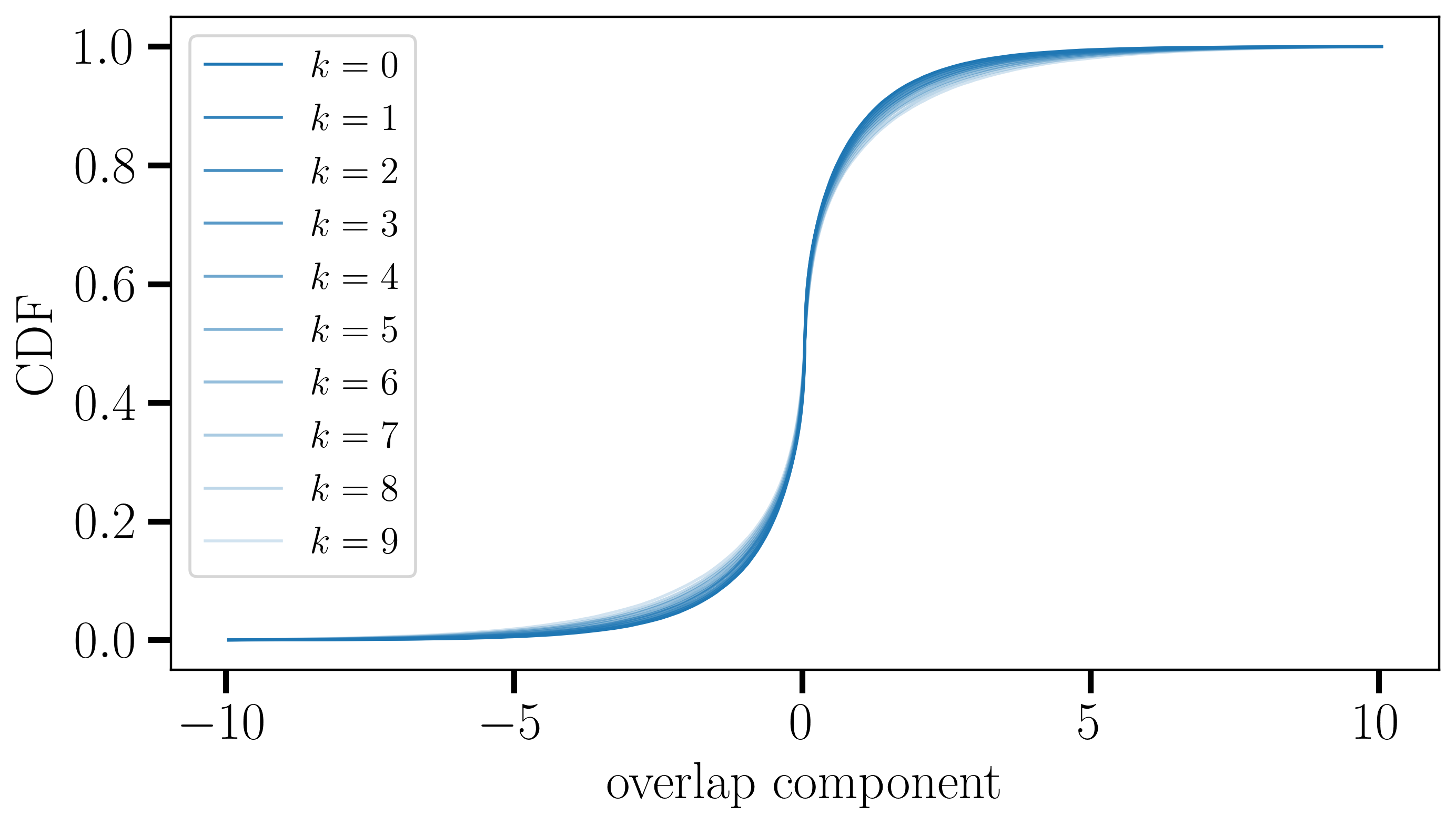}
  \caption{Poissonian setup with $k_{\mathrm{max}}=20$, $c=4$ and $\rho_W(W)=\delta(W-1)$ for $\bJ$, and signal $x_i \sim \mathcal{N}(0,1)$ of strength $\theta=5>\theta_{\mathrm{crit}}$. Plots are obtained via the population dynamics algorithm described in Appendix~\ref{app:popdyn} with population size $N_p=2\times10^5$. ({\it Top left}) The top eigenvector component density obtained via Eq.~\eqref{eq:top_eig_comp_density} (blue line) and direct diagonalisation (crosses). We also plot the standard Gaussian pdf (dashed line), which would be expected in case of full recovery, as a guide for the eye. ({\it Top right}) Overlap component density obtained via Eq.~\eqref{eq:overlap_comp_density} (blue line) and direct diagonalisation (crosses). ({\it Bottom left}) Degree decomposition of the top eigenvector component cumulative distribution function (CDF) in Eq.~\eqref{eq:top_eig_comp_density}. ({\it Bottom right}) Degree decomposition of the overlap component CDF in Eq.~\eqref{eq:overlap_comp_density}. }\label{fig:K1_xstdGauss_c4}
\end{figure*}

In Fig.~\ref{fig:ER_transition}, we present the numerical results obtained for $c=4$, $k_{\mathrm{max}}=20$, unit bond weights $W_{ij}=1$ and a standard Gaussian signal $\bx \sim \mathcal{N}(\mathbf 0,\mathbf I_N)$. For $\theta\leq\theta_{\mathrm{crit}}$, the average top eigenvalue, obtained by direct diagonalisation, stays constant and equal to $\lambda_{\theta=0}$, the structural top eigenvalue of the symmetric noise matrix $\bJ$. The value of $\lambda_{\theta=0}$ is obtained via a population dynamics algorithm using a reduced $\theta=0$ version of the equations in Result~\ref{res:RDE}, analysed in \cite{Susca19}. As soon as the signal strength exceeds $\theta_{\mathrm{crit}}$, the top eigenvalue of $\bA$ becomes correlated with the signal and is well predicted by $\lambda_{\theta}$ as in Eq.~\eqref{eq:gen_lambda_signal}. The second largest eigenvalue of $\bA$, on the other hand, numerically matches the structural eigenvalue $\lambda_{\theta=0}$ of the symmetric noise matrix $\bJ$. At $\theta=\theta_{\rm crit}$, the squared overlap between the top eigenvector and the spike becomes nonzero discontinuously, and simulation results match the prediction in Eq.~\eqref{eq:gen_overlap} (which is an approximation), indicated by dark blue diamonds. 

For $\theta_{\rm b}<\theta\leq\theta_{\rm crit}$, direct diagonalisation shows that the signal eigenvalue is outside of the bulk but remains below the structural outlier $\lambda_{\theta=0}$, so that it is the \textit{second} largest eigenvalue of $\bA$, as in the central panel of Fig.~\ref{fig:spectra}. In this window, the light blue diamonds report the squared overlap between the signal $\bx$ and the eigenvector associated with the \textit{second} largest eigenvalue, which grows continuously from zero, in contrast with the discontinuous jump of the top-eigenvector overlap at $\theta_{\rm crit}$. The light blue dashed line is the continuation of Eqs.~\eqref{eq:gen_lambda_signal_all} and~\eqref{eq:gen_overlap} into this window. It follows the numerical data closely, mirroring the behaviour found exactly in the RR case (Fig.~\ref{fig:RRG_lambdas_mu0}), but, as discussed in Remark~\ref{rem:poisson_validity}, it relies on an approximation for fluctuating degrees and should be regarded as a guide to the eye.

In the right panel, we plot the transition line associated to $\theta_{\mathrm{crit}}$ in Eq.~\eqref{eq:gen_transition} in the parameter space $(\theta,c)$:  the density plot of the average overlap shows how the transition between non-recovery (corresponding to zero overlap between the top eigenvector and signal vector) and recovery (where the overlap is of order one) takes place at $\theta_{\rm crit}$. 

In Fig.~\ref{fig:K1_xstdGauss_theta_crit}, we show the value of the critical transition in the signal strength $\theta_{\mathrm{crit}}$ as in Eq.~\eqref{eq:gen_transition}. In the left panel, we confirm that as the average degree $c$ (and thus the density of the noise matrix) grows, one needs stronger signal to recover it. In the right panel, we plot the same recovery transition against $c/k_{\max}$. For a fixed $c$, as the ratio $c/k_{\max}$ grows, that is, as $k_{\max}>c$ decreases, the recovery transition decreases.

In Fig.~\ref{fig:K1_xstdGauss_c4_marginals} we present the two marginals of the density $\uppi(\omega,h)$ in Eq.~\eqref{eq:RDE} for $c=4$, $k_{\mathrm{max}}=20$ and $\theta=5$ (here in the recovery phase). Namely, we plot $\uppi(h) = \int \de \omega\, \uppi(\omega,h)$ (left) and $\uppi(\omega) = \int \de h\, \uppi(\omega,h)$ (right). Note that, in the latter, the $k=1$ contribution in the RHS of Eq.~\eqref{eq:RDE} produces a $\delta(\omega-\lambda_\theta)$: a vertical dashed line at $\lambda_\theta$ is added to the figure for reference. 

Finally, in Fig.~\ref{fig:K1_xstdGauss_c4}, we show the average top eigenvector component and overlap component densities (top panels), and the individual contributions to the component densities coming from nodes of degrees $k=0,\dots,9$ (bottom panels). In the top panels a good agreement between the theory and direct diagonalisation results is shown.

\section{Conclusions and Outlook}\label{sec:conclusion}
We have studied the problem of recovering a rank-one signal $\frac{\theta}{N}\bm x \bm x^\intercal \in \R^{N\times N}$ from the noisy observation $\bJ + \frac{\theta}{N}\bm x \bm x^\intercal$, where the noise matrix $\bJ \in \R^{N\times N}$ is sparse. The noise is modeled as a sparse symmetric weighted adjacency matrix of a graph constructed from the configuration model with \textit{any} degree distribution $p_k$ with finite mean and bounded maximum degree. 

From a technical standpoint, our main contribution is the extension of the replica formalism used to calculate spectral properties of large matrices with sparse structure in \cite{Kuehn2008,Kabashima10, Kabashima12,Takahashi14,Susca21,Budnick25} to the class of spiked matrix models. The non-trivial extension, i.e., the addition of the perturbation $\frac{\theta}{N}\bm x \bm x^\intercal$ in the model, required the introduction of an auxiliary parameter $q$, identified as the overlap between the signal and top eigenvector, and modifications to recursive distributional equations to account for a spike-induced shift in the recursions that depends on the signal strength $\theta$ and the distribution of the spike components. We also derived the density of the top eigenvector components as in previous work along with a novel observable: the overlap component density; its average also quantifies the average overlap between the top eigenvector and signal vector. 

From a practical standpoint, the main contribution is the identification of the analogue of the BBP transition for rank-one signal recovery with sparse noise: for $\theta>\theta_{\rm crit}$, the signal correlates non-trivially with the top eigenvector. For \textit{any} degree distribution, the recursive distributional equations, solved by a population dynamics algorithm, give $\theta_{\rm crit}$, the typical top eigenvalue $\lambda_\theta$ and the overlap $q$. As expected, in the large connectivity limit our results converge to the known BBP-like transition, the top eigenvalue and squared overlap for dense real Wigner matrices \cite{Peche06,Capitaine09,Bloemendal_2012}.

Another advance of this work is the identification of a second transition in the signal strength: $\theta_{\rm b}\le\theta_{\rm crit}$, at which the signal eigenvalue detaches from the bulk. For noise with nonzero mean, the two thresholds do not need to coincide because of a structural sparse-graph outlier, and as a result the overlap with the top eigenvector jumps discontinuously at $\theta_{\rm crit}$. In the intermediate regime $\theta_{\rm b}<\theta\le\theta_{\rm crit}$ the signal is spectrally visible but not recoverable by na\"ive PCA: the signal-bearing eigenvalue is only the second largest, and it is the associated eigenvector that correlates with the signal, with an overlap that grows continuously from zero.

For random regular noise, we explicitly derive the two thresholds, the bulk edge, the signal eigenvalue and the squared overlap exactly and in closed form, and show that the top-eigenpair formulas remain exact for the sub-leading signal-bearing eigenpair throughout the intermediate regime $\theta_{\rm b}<\theta\le\theta_{\rm crit}$. For Poissonian noise, an approximation via a closure of recursions gives $\theta_{\rm crit}$, the signal eigenvalue and the squared overlap in the recovery phase, $\theta>\theta_{\rm crit}$. Direct diagonalisation shows the same intermediate regime as in the random regular case, and we believe the same mechanism is at work. However, for fluctuating degrees, the bulk edge is only estimated via the same closure, and a controlled treatment of $\theta_{\rm b}<\theta\le\theta_{\rm crit}$ would require the second-eigenpair statistics, which we leave for future work. In all cases we find excellent agreement with direct diagonalisation.

From a practitioner's point of view, our results quantify the cost of ignoring sparse corruption in low-rank matrix inference. Structural spectral features of the sparse noise can mask the signal-bearing eigenpair, separating the point at which the signal becomes spectrally visible from the point at which it is recoverable by na\"ive PCA. We show that the corresponding recovery thresholds depend on the parameters of the sparse noise matrix such as the average connectivity of the graph used to generate it.

A natural extension would be to consider sparse signal models that would contribute to the growing literature of recovery thresholds for Sparse PCA. 
The recursive distributional equations derived here describe the cavity-message distribution induced by the sparse graph structure, and therefore provide a natural analytical starting point for noise-aware sparse PCA methods, which is a clear future direction. For instance, the degree-decomposed overlap statistics identify which degree classes contribute most strongly to the leading eigenvector near the threshold, suggesting natural sparse-aware extensions such as degree-reweighted spectral estimators or operators designed to suppress structural sparse-graph modes.
Moreover, extensions to a configuration model for the noise matrix with correlated degrees, higher-rank signals, structured priors on the signal components or heavy-tailed assumptions on either are also interesting avenues for future work. Finally, localisation phenomenon is observed for zero-centered Gaussian bond weights, where the top eigenvector correlates with the high-degree nodes rather than aligning with the signal: a more thorough analysis of the nature of the localisation is left for future work.

\section*{Acknowledgments}
The authors thank Preben Forer, Sebastian Goldt, Barak Budnick and Giovanni Piccioli for helpful discussions. P.V. acknowledges support from UKRI FLF Scheme (No. MR/X023028/1). This work was partially supported by project SERICS (PE00000014) under the MUR National Recovery and Resilience Plan funded by the European Union - NextGenerationEU.

\newpage
\appendix
\onecolumngrid
\section{Replica derivation}\label{app:Afullreplica}
In this Appendix, we outline the derivations in the replica calculation. The object of main interest is the replicated partition function $\mathbb E_{\bA}\big[\big(Z_\beta[\bA;tF]\big)^n\big]$, with $Z_\beta[\bA;tF]$ given by Eq.~\eqref{eq:Z_vtop_multiplier}, namely
\begin{equation}
Z_\beta[\bA;tF]\coloneqq \int\de \bv\exp\left[\frac{\beta}{2}\langle\bv,\bA\bv\rangle+\beta tN F(u,\bv,\bx)\right] \delta\left(\|\bv\|^{2}-N\right) \ ,
\end{equation}
and $\bA=\bC\odot\bW+\frac{\theta}{N}\bx \bx^\intercal$. We will specifically focus on the cases in which $F$ can be written as
\begin{equation}
F(u,\bv,\bx)=\frac{1}{N}\sum_if(u,v_i,x_i) \ . \label{app:eq:F_defn_f}
\end{equation}
Recall the noise model construction in Eq.~\eqref{eq:general_J_condk} and Eq.~\eqref{eq:boundedER_connectivity}. For the purpose of the calculation, we follow a shortcut procedure whose validity has been proved in Appendix B of \cite{Susca19}.

It consists of the following steps:
\begin{enumerate}
    \item Replace the true connectivity distribution in Eq.~\eqref{eq:boundedER_connectivity} with
    \begin{equation}
P_c(\bC|\boldsymbol k)\propto\prod_{i=1}^N\mathbb I\left(\sum_{j}c_{ij}=k_{i}\right)\delta_{c_{ii},0}\prod_{j>i} \left[ \frac{c}{N}\delta_{c_{ij},1}+\left(1-\frac{c}{N}\right)\delta_{c_{ij},0} \right] \ ,\label{eq:ER_conndist}
\end{equation}
where $\boldsymbol{k}=(k_i)_{i=1}^N$ is a vector of integers $k_i$ i.i.d.~sampled with distribution $p_k$, and $c\coloneqq\langle k\rangle=\sum_k kp_k$. This replacement has not changed the probability of sampling a matrix $\bC$ in the ensemble, as the extra factor is constant for the matrices in the ensemble. 
\item Drop the degree constraint term $\prod_i\mathbb I\left(\sum_{j}c_{ij}=k_{i}\right)$ altogether. The resulting distribution
\begin{equation}
P_{ER}(\bC)=\prod_{i=1}^N\delta_{c_{ii},0}\prod_{j>i} \left[ \frac{c}{N}\delta_{c_{ij},1}+\left(1-\frac{c}{N}\right)\delta_{c_{ij},0} \right] \ ,\label{eq:ER_conndistTrue}
\end{equation}
now corresponds to a sparse Erd\H{o}s-R\'enyi (ER) graph with average connectivity $c$. While the distribution \eqref{eq:ER_conndistTrue} on adjacency matrices $\bC$ is no longer equivalent to either \eqref{eq:ER_conndist} or \eqref{eq:boundedER_connectivity}, it will produce a final result for the average replicated partition function $\mathbb E_{\bA}\big[\big(Z_\beta[\bA;tF]\big)^n\big]$ (see Eq. \eqref{eq:pi_RS} below) involving the Poissonian degree distribution $p_k^{(c)}$ characteristic of ER graphs. 
\item The final result for \emph{generic} degree distributions $p_k$ cited in the main text \eqref{eq:RDE} follows by simply replacing $p_k^{(c)}$ with $p_k$ in the final formulas. The validity of this shortcut was demonstrated in Appendix B of \cite{Susca19}.
\end{enumerate}
By keeping the previous observation in mind, let us start by replicating $Z_\beta[\bA;tF]$ and averaging over $\bA$. We get
\begin{multline}
\mathbb E_{\bA} \left[ \left( Z_\beta[\bA;tF]\right)^n \right]\propto\left(\prod_{a=1}^{n}\int\delta\left(\|\bv_{a}\|^{2}-N\right)\de\bv_{a}\right)\mathbb E_{\bA} \left[ \exp\left( \frac{\beta}{2}\sum_{a=1}^{n}\langle\bv_a,\bA\bv_a\rangle+\beta t N\sum_aF(u,\bv_a,\bx)\right) \right]\\
=\left(\prod_{a=1}^{n}\int\de\bv_{a}\int\de\lambda_a\right)\mathbb E_{\bA} \left[ \exp\left(\mathrm{i}\frac{N\beta}{2}\sum_a\lambda_a- \mathrm{i} \beta\sum_{ai} v_{ai}^2\lambda_a+\frac{\beta}{2}\sum_{a=1}^{n}\langle\bv_a,\bA\bv_a\rangle+\beta tN\sum_aF(u,\bv_a,\bx)\right) \right] \ .\label{eq:rep_Z_first}
\end{multline}
We can perform immediately the expectation over $\bJ = \bC \odot \bW $ \cite{Kuehn2008} by invoking the law of total expectation $\mathbb E_\bA[\bullet]=\mathbb E_\bx[\mathbb E_{\bJ|\bx}[\bullet]]$, {with $\bf C$ sampled from the ER distribution in Eq.~\eqref{eq:ER_conndistTrue}. By using Appendix D of \cite{Susca21_review}
\begin{equation}
  \mathbb E_{\bJ|\bx}\left[\exp\left(\frac{\beta}{2}\sum_{a=1}^n\langle\bv_a,\bJ\bv_b\rangle\right)\right] = 
  \left(\frac{c}{2N}\sum_{ij}\left(\mathbb{E}_W [\mathrm{e}^{\beta W\langle\sbv_i,\sbv_j\rangle}]-1\right)\right)
\end{equation}
we obtain}
\begin{multline}
\mathbb E_{\bx} \left[\mathbb E_{\bJ|\bx}\left[ \exp\left( \frac{\beta}{2}\sum_{a=1}^{n}\langle\bv_a,\bA\bv_a\rangle+\beta t N\sum_aF(u,\bv_a,\bx)\right)\right] \right]\\
=\mathbb E_{\bx} \left[ \exp\left( \frac{c}{2N}\sum_{ij}\left(\mathbb E_W[\e^{\beta W\langle\sbv_i,\sbv_j\rangle}]-1\right)+\frac{\theta\beta}{2N}\sum_a\langle\bv_a,\bx\rangle^2+\beta t N\sum_aF(u,\bv_a,\bx)\right) \right]\\\comprimi
=\exp\left( \frac{c}{2N}\sum_{ij}\left(\mathbb E_W[\e^{\beta W\langle\sbv_i,\sbv_j\rangle}]-1\right)\right)\mathbb E_{\bx} \left[\exp\left(\frac{\theta\beta}{2N}\sum_a\langle\bv_a,\bx\rangle^2+\beta t N\sum_aF(u,\bv_a,\bx)\right) \right] \ ,\label{eq:arg_x}
\end{multline}
where $\sbv_i\coloneqq(v_{ia})_{a\in[n]}$ is the $n$-replicated $i$th-component of $\bv$ \cite{Susca21_review} and we have denoted by $W$ a random variable with distribution $\rho_W$. Let us now introduce the following (normalized) functional order parameters
\begin{align}
\varphi(\sbv) &\coloneqq \frac{1}{N}\sum_{i}\delta^{(n)}(\sbv-\sbv_i) \ ,& 
\psi(\sbv)&\coloneqq\frac{1}{N} \sum_{i=1}^N\delta^{(n)}(\sbv-x_i\sbv_i) \ , \label{eq:Psi}
\end{align}
where $\delta^{(n)}$ is the $n$-dimensional Dirac delta function $\delta^{(n)}(\sbv-\sbv_i)=\prod_{a=1}^n \delta\left(v_{a}-v_{ia}\right)$, by means of a suitable Fourier representation 
\begin{multline}\comprimi
1=\iint\mathcal D\psi\mathcal D\hat\psi~\exp\left(-\mathrm{i}\int\hat\psi(\sbv)\left(N\psi(\sbv)-\sum_i\delta^{(n)}(\sbv-x_i\sbv_i)\right)\de\sbv\right)\iint\mathcal D\varphi\mathcal D\hat\varphi\exp\left(-\mathrm{i}\int\hat\varphi(\sbv)\left(N\varphi(\sbv)-\sum_i\delta^{(n)}(\sbv-\sbv_i)\right)\de\sbv\right)\\\comprimi =
\iint\mathcal D\psi\mathcal D\hat\psi\iint\mathcal D\varphi\mathcal D\hat\varphi~\exp\left(\mathrm{i}\sum_i\hat\psi(x_i\sbv_i)+\mathrm{i}\sum_\mathrm{i}\hat\varphi(\sbv_i)-\mathrm{i}N\int\left(\hat\psi(\sbv)\psi(\sbv)+\hat\varphi(\sbv)\varphi(\sbv)\right)\de\sbv\right) \ ,
\end{multline}
where $\de\sbv=\prod_{a=1}^n\de v_a$. Inserting this representation of the unity within the argument of $\mathbb E_\bx[\bullet]$ in Eq.~\eqref{eq:arg_x}:
\begin{align}
&\exp\left( \frac{c}{2N}\sum_{ij}\left(\mathbb E_W[\e^{\beta W\langle\sbv_i,\sbv_j\rangle}]-1\right)\right)\mathbb E_{\bx} \left[\exp\left(\frac{\theta\beta}{2N}\sum_a\langle\bv_a,\bx\rangle^2+\beta t \sum_a\sum_if(u,v_{ia},x_i)\right) \right]\nonumber\\
&{\comprimi=\exp\left( \frac{c}{2N}\sum_{ij}\left(\mathbb E_W[\e^{\beta W\langle\sbv_i,\sbv_j\rangle}]-1\right)\right)\mathbb E_{\bx} \left[\iint\mathcal D\psi\mathcal D\hat\psi\iint\mathcal D\varphi\mathcal D\hat\varphi\exp\left(i\sum_i\left(\hat\psi(x_i\sbv_i)+\hat\varphi(\sbv_i)\right)\right.\right.}\nonumber\\
&{\comprimi\quad\left.\left.-\mathrm{i}N\int\left(\hat\psi(\sbv)\psi(\sbv)+\hat\varphi(\sbv)\varphi(\sbv)\right)\de\sbv+\frac{\theta\beta}{2N}\sum_a\langle\bv_a,\bx\rangle^2+\beta t \sum_a\sum_if(u,v_{ia},x_i)\right) \right]}  \nonumber\\
&{\comprimi=\iint\mathcal D\varphi\mathcal D\hat\varphi\iint\mathcal D\psi\mathcal D\hat\psi\exp\left(\frac{\theta\beta N}{2}\iint \langle\sbv,\sbv'\rangle\psi(\sbv)\psi(\sbv')\de\sbv\de\sbv'\right)} \nonumber\\
&{\quad\comprimi\times \exp\left( \frac{cN}{2}\iint\de\sbv\de\sbv'\varphi(\sbv)\varphi(\sbv')\left(\mathbb E_W[\e^{\beta W\langle\sbv,\sbv'\rangle}]-1\right)-\mathrm{i} N\int\left(\hat\psi(\sbv)\psi(\sbv)+\hat\varphi(\sbv)\varphi(\sbv)\right)\de\sbv\right)}\nonumber\\
&{\quad\times\comprimi \mathbb E_{\bx} \left[\exp\left(\sum_i \left( \mathrm{i}\hat\psi(x_i\sbv_i)+\mathrm{i}\hat\varphi(\sbv_i)+\beta t \sum_a f(u,v_{ia},x_i)\right) \right) \right] \ .}
\end{align}    
The replicated partition function now reads as
\begin{align}
&\mathbb E_{\bA} \left[ \left( Z_\beta[\bA;tF]\right)^n \right]=\int\de\sblambda\iint\mathcal D\psi\mathcal D\hat\psi\iint\mathcal D\varphi\mathcal D\hat\varphi\nonumber \\
&\comprimi\quad\times \exp\left(-\mathrm{i}N\int\varphi(\sbv)\hat\varphi(\sbv)\de\sbv +\frac{cN}{2}\iint\varphi(\sbv)\varphi(\sbv')\left(\mathbb E_W[\e^{\beta W\langle\sbv,\sbv'\rangle}]-1\right)\de\sbv\de\sbv'+\mathrm{i}\frac{N\beta}{2}\sum_a\lambda_a\right)\nonumber
\\
&{\quad\comprimi\times\exp\left(-\mathrm{i} N\int\psi(\sbv)\hat\psi(\sbv)\de\sbv+\frac{\theta\beta N}{2}\iint\langle\sbv,\sbv'\rangle\psi(\sbv)\psi(\sbv')\de\sbv'\de\sbv\right)}\nonumber\\
&\quad\comprimi\times \left[ \mathbb E_X \int\de\sbv~\exp\left(-\frac{\mathrm{i}\beta}{2}\sum_a\lambda_av_{a}^2+\mathrm{i}\hat\varphi(\sbv)+\mathrm{i}\hat\psi(X \sbv)+\beta t\sum_a f(u,v_{a},X)\right)\right]^N\ ,\label{replicatedFunctional}
\end{align}
where $\sblambda\coloneqq(\lambda_a)_{a \in [n]}$ and $X \sim \varrho_x$ has the same distribution as each i.i.d.~entry of $\bx$. Let us now introduce
\begin{equation}
I_f[\hat\varphi,\hat\psi, \sblambda;t]\coloneqq\frac{1}{\beta n}\ln\mathbb E_X\left[\int\de\sbv~\exp\left(-\frac{\mathrm{i}\beta}{2}\sum_a\lambda_av_{a}^2+\mathrm{i}\hat\varphi(\sbv)+\mathrm{i}\hat\psi(X\sbv)+\beta t\sum_a f(u,v_{a},X)\right)\right] \ . \label{app:eq:I_f_preRS}
\end{equation}
The functional integral \eqref{replicatedFunctional} can now be evaluated for large $N$ by means of a saddle-point approximation 
\begin{align}
    \mathbb E_{\bA} \left[ \left( Z_\beta[\bA;tF]\right)^n \right] &\propto \int\de\sblambda\iint\mathcal D\psi\mathcal D\hat\psi\iint\mathcal D\varphi\mathcal D\hat\varphi 
    \,\e^{N\mathcal S_\beta[\varphi,\hat\varphi,\psi,\hat\psi,\sblambda
    ;t,f]} \approx \e^{N\mathcal S_\beta[\varphi^\star,\hat\varphi^\star,\psi^\star,\hat\psi^\star,\sblambda^\star(t);t,f]} \ , \label{eq:Zn_preRS}
\end{align}
where the action reads as
 \begin{multline}
 \mathcal S_\beta[\varphi,\hat\varphi,\psi,\hat\psi;\sblambda;t,f]\coloneqq -\mathrm{i}\int\left(\varphi(\sbv)\hat\varphi(\sbv)+\psi(\sbv)\hat\psi(\sbv)\right)\de\sbv +\frac{c}{2}\iint\varphi(\sbv)\varphi(\sbv')\left(\mathbb E_W[\e^{\beta W\langle\sbv,\sbv'\rangle}]-1\right)\de\sbv\de\sbv'\\\comprimi+\mathrm{i}\frac{\beta}{2}\sum_a\lambda_a
  +\frac{\theta\beta}{2}\iint\langle\sbv,\sbv'\rangle\psi(\sbv)\psi(\sbv')\de\sbv'\de\sbv+\beta nI_f[\hat\varphi,\hat\psi,\sblambda;t] \ . \label{eq:app:action_nonRS}
 \end{multline}

The saddle-point values of the fields and the multipliers are determined from the following saddle-point equations, obtained differentiating the action w.r.t. $\varphi,\hat\varphi,\psi,\hat\psi$ and $\lambda_a$ and setting the derivatives equal to zero:
\begin{align}
1&=\frac{\mathbb E_X\int\de\sbv\,v_a^2\exp\left(-\frac{\beta}{2}\sum_a\lambda_a^\star(t) v_{a}^2+\mathrm{i}\hat\varphi^\star(\sbv)+\mathrm{i}\hat\psi^\star(X\sbv)+\beta t\sum_a f(u,v_{a},X)\right)}{\mathbb E_X\int\de\sbv'\exp\left(-\frac{\beta}{2}\sum_a\lambda_a^\star(t) v_{a}'^2+\mathrm{i}\hat\varphi^\star(\sbv')+\mathrm{i} \hat\psi^\star(X\sbv')+\beta t\sum_a f(u,v_{a}',X)\right)} \ ,\label{constraintlambdaaoft}\\
\mathrm{i} \hat\varphi^\star(\sbv)&=c\int\varphi^\star(\sbv')\left(\mathbb E_W[\e^{\beta W\langle\sbv,\sbv'\rangle}]-1\right)\de\sbv' \ 
,\\
\varphi^\star(\sbv)&=\frac{\mathbb E_X\left[\exp\left(-\frac{\beta}{2}\sum_a\lambda_a^\star(t)v_{a}^2+\mathrm{i}\hat\varphi^\star(\sbv)+\mathrm{i}\hat\psi^\star(X\sbv)+\beta t\sum_a f(u,v_{a},X)\right)\right]}{\mathbb E_X\left[\int\de\sbv'\exp\left(-\frac{\beta}{2}\sum_a\lambda_a^\star(t){v_{a}'}{}^2+\mathrm{i}\hat\varphi^\star(\sbv')+\mathrm{i}\hat\psi^\star(X\sbv')+\beta t\sum_a f(u,v_{a}',X)\right)\right]} \   ,\\
\mathrm{i} \hat\psi^\star(\sbv)&=\theta\beta\int\langle\sbv,\sbv'\rangle\psi^\star(\sbv')\de\sbv'
,\label{hatpsistardotproduct}\\\comprimi
\psi^\star(\sbv)&=\frac{\mathbb E_X\left[|X|^{-n}\exp\left(-\frac{\beta}{2X^2}\sum_a\lambda_a^\star(t)v_{a}^2+\mathrm{i} \hat\varphi^\star(X^{-1}\sbv)+\mathrm{i} \hat\psi^\star(\sbv)+\beta t\sum_a f(u,X^{-1}v_{a},X)\right)\right]}{\mathbb E_X\left[\int\de\sbv'\exp\left(-\frac{\beta}{2}\sum_a\lambda_a^\star(t){v_{a}'}{}^2+\mathrm{i} \hat\varphi^\star(\sbv')+\mathrm{i} \hat\psi^\star(X\sbv')+\beta t\sum_a f(u,v_{a}',X)\right)\right]} \label{PsiStarDefiningCoft}\ ,
\end{align}
the first one to be intended for all $a=1,\dots,n$.  
We have redefined $\mathrm{i}\lambda_a\to\lambda_a$ 
and explicitly highlighted the $t$-dependence of the multipliers $\lambda_a^\star$ at the saddle-point, which follows from the constraint \eqref{constraintlambdaaoft} and will be important in the following. It follows from \eqref{hatpsistardotproduct} that $\mathrm{i}\hat\psi^\star(\sbv)$ must be a linear function of its arguments
\begin{equation}\mathrm{i}\hat\psi^\star(\sbv)=\theta\beta\langle \sbQ(t),\sbv\rangle \ ,
\qquad \sbQ(t) \coloneqq\int\sbv\, \psi^\star(\sbv)\de\sbv
\end{equation}
in terms of a $t$-dependent $n$-dimensional vector $\sbQ(t)=(q_a(t))$ for $a=1,\dots,n$. Multiplying Eq. \eqref{PsiStarDefiningCoft} by $v_a$ and integrating, we find that $\sbQ(t)$ must satisfy component-wise the self-consistency equation
\begin{equation}
\sbQ(t)=\frac{\mathbb E_X \left[\mathrm{sign}(X^n)X\!\int\!\de \sbv\ \sbv\exp\left(-\frac{\beta}{2}\sum_a\lambda_a^\star(t)v_{a}^2+\mathrm{i}\hat\varphi^\star(\sbv)+\theta\beta X\langle\sbQ(t),\sbv\rangle
+\beta t\! \sum_a\! f(u,v_{a},X)\right)\right]}{\mathbb E_X\left[\int\de\sbv'\exp\left(-\frac{\beta}{2}\sum_a\lambda_a^\star(t) v_{a}'^2+\mathrm{i}\hat\varphi^\star(\sbv')+\theta\beta X\langle\sbQ(t),\sbv'\rangle
+\beta t \sum_a f(u,v_{a}',X)\right)\right]}\ , \label{eq:C_Selfconsist}
\end{equation}
while the saddle-point equations above can be reduced to
\begin{align}
1&=\frac{\mathbb E_X \int\de\sbv\,v_a^2\exp\left(-\frac{\beta}{2}\sum_a\lambda_a^\star(t)v_{a}^2+\mathrm{i}\hat\varphi^\star(\sbv)+X\theta\beta\langle\sbQ(t),\sbv\rangle
+\beta t\sum_a f(u,v_{a},X)\right)}{\mathbb E_X\int\de\sbv'\exp\left(-\frac{\beta}{2}\sum_a\lambda_a^\star(t)v_{a}'^2+\mathrm{i}\hat\varphi^\star(\sbv')+\theta\beta X\langle\sbQ(t),\sbv'\rangle
+\beta t\sum_a f(u,v_{a}',X)\right)} \ ,\label{SPlambdatFields}\\
\mathrm{i}\hat\varphi^\star(\sbv)&=c\int\varphi^\star(\sbv')\left(\mathbb E_W[\e^{\beta W\langle\sbv,\sbv'\rangle}]-1\right)\de\sbv' \ 
, \label{SPlambda_phihat}\\
\comprimi \varphi^\star(\sbv)&=\frac{\mathbb E_X\left[\exp\left(-\frac{\beta}{2}\sum_a\lambda_a^\star(t)v_{a}^2+\mathrm{i}\hat\varphi^\star(\sbv)+\theta\beta X\langle\sbQ(t),\sbv\rangle 
+\beta t\sum_a f(u,v_{a},X)\right)\right]}{\mathbb E_X\left[\int\de\sbv'\exp\left(-\frac{\beta}{2}\sum_a\lambda_a^\star(t){v_{a}'}{}^2+\mathrm{i}\hat\varphi^\star(\sbv')+\theta\beta X\langle\sbQ(t),\sbv'\rangle
+\beta t\sum_a f(u,v_{a}',X)\right)\right]} \ . \label{SPlambda_phi}
\end{align}
Now, recall from Eq. \eqref{remarkable} that we will be interested in extracting the limit $n\to 0$ followed by the limit $\beta\to\infty$ of the action \eqref{eq:app:action_nonRS} evaluated at the saddle-point values of the fields and multipliers. To extract these limits, it is necessary to postulate what kind of invariance the action and the saddle-point fields will enjoy with respect to permutation of replica indices. In the random matrix literature, it has been established \cite{Nagao07,Kuehn2008,Susca19,Susca21,Susca21_review,Budnick25} that a solution that preserves the symmetry among replicas is exact for extreme value problems on sparse random matrices \footnote{A stronger rotationally invariant assumption (where dependence on $\sbv$ would only be through its modulus) was shown to lead to correct solutions for harmonically coupled systems \cite{KuehnZipp07} and spectra of sparse random matrices \cite{Kuehn2008}, but was found to be too restrictive for the top eigenpair problem of the same matrix models \cite{Nagao07,Susca21,Susca21_review}.}. In this paper, we indeed follow the strategy introduced in \cite{Kuehn2008,KuehnZipp07} and represent the fields at the saddle-point as superposition of Gaussians with fluctuating mean and variance, i.e.
\begin{subequations}
\begin{align}
\lambda_a^\star(t)&\equiv \lambda_\theta(t)\qquad\forall a \ , \label{eq:RS_ansatz_lambda} \\ 
\sbQ(t)&=q(t)\bOne_n \ , \label{eq:RS_ansatz_C}\\
 \varphi^\star(\sbv)&=\int\frac{\e^{-\frac{\beta\omega}{2}\|\sbv\|^2+\beta h\langle\bOne_n,\sbv\rangle}}{\mathcal Z_\beta^n(\omega,h)}\uppi(\omega,h)\de\omega\de h \ , \label{phistaransatz} \\
\mathrm{i}\hat\varphi^\star(\sbv)&=\hat\phi\int \e^{\frac{\beta\hat\omega}{2}\|\sbv\|^2+\beta \hat h\langle\bOne_n,\sbv\rangle}\hat\uppi(\hat\omega,\hat h)\de\hat\omega\de\hat h \label{ihatphistaransatz}\ ,
\end{align}\label{app:eq:spansatz}\end{subequations}
for some joint probability densities $\uppi$ and $\hat\uppi$ defined and normalized for $\omega,\hat\omega>0$ and $h,\hat h\in\R$, and for $\mathcal Z_\beta(a,b)\coloneqq\sqrt{\frac{2\pi}{a\beta}}\exp\left(\frac{\beta b^2}{2a}\right)$. The constant $\hat\phi$ is introduced as the conjugate field $\mathrm{i}\hat\varphi^\star(\sbv)$ does not have the interpretation of a density (as opposed to $\varphi^\star$ from \eqref{eq:Psi}) and thus needs not be normalised. 

We now insert the ansatz above into the action \eqref{eq:app:action_nonRS} evaluated at the saddle-point and compute the corresponding $\sbv$ and $\sbv'$ Gaussian integrals as required. For instance
\begin{align}
    \nonumber  -\mathrm{i}\int \de \sbv\,\varphi^\star(\sbv) \hat\varphi^\star(\sbv)&\mapsto - \hat\phi \int \de \uppi \de \hat\uppi \int \de \sbv  \prod_a \frac{\e^{-\frac{\beta}{2}(\omega-\hat \omega) v_a^2 + \beta (h+\hat h) v_a}}{\mathcal Z_\beta(\omega, h)}   \\
    \nonumber &= - \hat\phi \int \de \uppi \de \hat\uppi \left[ \frac{\mathcal Z_\beta(\omega-\hat \omega, h+\hat h)}{\mathcal Z_\beta(\omega, h)} \right]^n = - \hat\phi - \hat\phi n \int \de \uppi \de \hat\uppi \ln \left[ \frac{\mathcal Z_\beta(\omega-\hat\omega, h+\hat h)}{\mathcal Z_\beta(\omega, h)}  \right]+o(n) \ ,
\end{align}
where in the last step we have extracted the $n\to 0$ asymptotic behaviour using $A^n\sim 1+n\ln A$ and the fact that both $\uppi$ and $\hat\uppi$ are normalised to one. Similarly from  \eqref{app:eq:I_f_preRS} and with some notation redundancy,
\begin{align}
\nonumber &\beta I_{f}[\hat\uppi,q(t),\lambda_\theta(t);t]\coloneqq\lim_{n\to 0}\frac{\beta nI_f[\hat\varphi^\star,\hat\psi^\star, \sblambda^\star(t);t]-\hat\phi}{n} \\
\nonumber &=\lim_{n\to 0}\frac{1}{n}\ln \mathbb E_X\left[\int \de\sbv~\e^{-\frac{\beta}{2}\lambda_\theta(t)\sum_a v_{a}^2+\mathrm{i}\hat\psi^\star(X\sbv)+\beta t\sum_a f(u,v_{a},X)+\mathrm{i}\hat\varphi^\star(\sbv)-\hat\phi} \ \right] \\
\nonumber&= \lim_{n\to 0}\frac{1}{n}\ln\mathbb E_X \sum_{s=0}^\infty \frac{\hat\phi^s\e^{-\hat\phi}}{s!} \int \{\de \hat\uppi\}_s \int \de\sbv \prod_{a=1}^s \left(\e^{\frac{\beta}{2}\hat\omega v_a^2 + \beta \hat h v_a}\right) \e^{- \frac{\beta}{2}\lambda_\theta(t)\sum_{a} v_a^{2} +\beta t\sum_{a} f(u,v_{a},X) + \beta \theta X \langle\sbQ(t),\sbv\rangle}  \\
\nonumber&= \lim_{n\to 0}\frac{1}{n} \ln\mathbb E_X \sum_{s=0}^\infty \frac{\hat\phi^s\e^{-\hat\phi}}{s!} \int \{\de \hat\uppi\}_s  \left[  \int \de v \ \e^{-\frac{\beta}{2}(\lambda_\theta(t)-\{\hat\omega\}_s) v^2 + \beta (\theta q(t) X + \{\hat h\}_s)v + \beta t f(u,v,X)} \right]^n\\
\nonumber&=\lim_{n\to 0}\frac{1}{n} \ln\mathbb E_X \left[ 1 + n \sum_{s=0}^\infty \frac{\hat\phi^s \e^{-\hat\phi}}{s!} \int \{\de \hat\uppi\}_s \ln \int \de v \ \e^{-\frac{\beta}{2}(\lambda_\theta(t)-\{\hat\omega\}_s) v^2 + \beta (\theta q(t) X + \{\hat h\}_s)v + \beta t f(u,v,X)} \right]\\
&=\mathbb E_X \sum_{s=0}^{\infty} p^{(\hat\phi)}_s \int \{\de \hat\uppi\}_s \ln \int \de v \ \e^{-\frac{\beta}{2}(\lambda_\theta(t)-\{\hat\omega\}_s) v^2 + \beta (\theta q(t) X + \{\hat h\}_s)v + \beta t f(u,v,X)}\ ,\label{app:eq:Ifn}
\end{align}
where we have first expanded the exponential $\exp(\mathrm{i}\hat\varphi^\star(\sbv))$ of the conjugate field in a Taylor series, then inserted the ansatz \eqref{ihatphistaransatz}, and finally extracted the $n\to 0$ behaviour. We introduced here the shorthands anticipated in the main text, namely $\{\de\hat\uppi\}_s=\prod_{\ell=1}^s \de \hat\omega_\ell \de \hat h_\ell \uppi(\hat\omega_\ell,\hat h_\ell) $ and $\{x\}_s=\sum_{\ell=1}^s x_\ell$. Note the appearance of the Poissonian degree distribution of average $\hat\phi$ $$p^{(\hat\phi)}_k=\frac{\hat\phi^k}{k!}\e^{-\hat\phi}$$ corresponding to the degree distribution of an Erd\H{o}s-R\'enyi graph of average degree $\hat\phi$. In the following, given a function $f(k)$, we will use the shorthand notation
\[\langle f(k)\rangle_{\hat\phi}\coloneqq\sum_{k=0}^\infty f(k)\frac{\hat\phi^k}{k!}\e^{-\hat\phi} \ .\]

The other terms in the action can be treated in a similar way, and we refer to \cite{Susca21_review} for the details of the calculation. 
The introduction of $\uppi$ and $\hat\uppi$ leads to a rewriting
\begin{equation}\comprimi
\mathbb E_{\bA} \left[ \left( Z_\beta[\bA;tF]\right)^n \right] \  \approx \ \e^{N\mathcal S_\beta[\uppi,\hat\uppi,\lambda_\theta(t),q(t);t,f]} \ ,
\end{equation}
where $\mathcal S_\beta[\uppi,\hat\uppi,\lambda_\theta(t),q(t);t,f]$ is the action at the saddle point $\mathcal S_\beta[\varphi^\star,\hat\varphi^\star,\psi^\star,\hat\psi^\star,\sblambda^\star(t);t,f]$ from Eq.~\eqref{eq:Zn_preRS} with the ansatz in Eqs.~\eqref{app:eq:spansatz} inserted. In the $n\to 0$ limit we get the expression
\begin{multline}
s_\beta[\uppi,\hat\uppi,q(t),\lambda_\theta(t);t]\coloneqq\lim_{n\to 0}\frac{1}{n\beta}\mathcal S_\beta[\uppi,\hat\uppi,\lambda_\theta;t,f] = \frac{c}{2} \frac{1}{\beta}\int\mathbb E_W\left[ \ln \frac{\mathcal Z_\beta^{(2)}(h,\omega,h^\prime, \omega^\prime, W)}{\mathcal Z_\beta(\omega, h) \mathcal Z_\beta(\omega^\prime, h^\prime) }\right] \de \uppi(\omega,h) \de \uppi(\omega^\prime, h^\prime)\\ - \frac{\hat\phi}{\beta}\int \ln \frac{\mathcal Z_\beta(\omega-\hat\omega, h+\hat h)}{\mathcal Z_\beta(\omega, h)}  \de\uppi(\omega,h) \de\hat\uppi(\hat \omega,\hat h)
- \frac{\theta}{2} q(t)^2 +\frac{1}{2} \lambda_\theta(t)+ I_{f}[\hat\uppi,q(t),\lambda_\theta(t);t] \ ,\label{app:eq:RS_action}
\end{multline}

where we use the shorthands $\de\uppi(\omega,h) = \uppi(\omega,h)\de\omega\de h$ and $W$ is distributed as $\varrho_W$. We have also introduced
\begin{equation}
    \mathcal Z_\beta^{(2)}(h,\omega,h^\prime,\omega^\prime, W)=\int \de v \mathrm{\de } v^\prime \e^{-\frac{\beta}{2} \omega v^2 + \beta h v - \frac{\beta}{2}\omega^\prime v^{\prime\,2} + \beta h^\prime v^\prime + \beta W v v^\prime}= \mathcal Z_\beta(\omega^\prime,h^\prime) \mathcal Z_\beta \left( \omega- \frac{W^2}{\omega^\prime},h + \frac{h^\prime W}{\omega^\prime}\right) \ .
\end{equation}

The quantity $s_\beta[\uppi,\hat\uppi,q(t),\lambda_\theta(t);t]$ will allow for multiple computations. For example, if we invoke Eq.~\eqref{eq:replica_trick_exchNn} we have
\begin{equation}
\nonumber    \mathbb E_{\bA} \left[ \lambda_{\rm top} \right] = \lim_{\beta \to \infty} \frac{2}{\beta}   \lim_{n\to 0} \frac{1}{n} \lim_{N \to \infty} \frac{1}{N}\ln \mathbb E_{\bA} [ Z_{\beta}^n(\bA)]  
       =2\lim_{\beta\to+\infty} s_\beta(\uppi,\hat\uppi,q(t=0),\lambda_\theta(t=0);t=0)
     \label{eq:replica_trick_exchNnAppendix}
\end{equation}
for the average top eigenvalue, since $Z_{\beta}(\bA)$ is obtained from $Z_\beta[\bA;tF]$ by setting $t=0$. This case is tackled in the next subsection.

\subsection{Saddle point equations for $\uppi$ and $\hat\uppi$ and average top eigenvalue\label{sec:app:avg_top_lambda}}
Setting $t=0$ in the above equations, and inserting the ansatz in Eqs.~\eqref{app:eq:spansatz} into Eq.~\eqref{eq:C_Selfconsist}, upon extraction of the $n\to 0$ limit at finite $\beta$ we obtain the first one of the integral conditions linking $q\equiv q(t=0)$ and $\lambda_\theta\equiv \lambda_\theta(t=0)$, namely
\begin{equation}
    q = \mathbb E_X \left\langle\int \{ \de \hat\uppi \}_k \frac{ \{\hat h\}_k X + \theta q X^2}{\lambda_\theta - \{ \hat \omega \}_k}   \right\rangle_{\hat\phi} \ , \label{eq:Ct0_chat}
\end{equation}
while inserting the ansatz into the saddle point equation \eqref{SPlambdatFields} we obtain 
\begin{align}
    1 = \mathbb E_X\left\langle\int \{ \de \hat\uppi \}_k \left[ \frac{1}{\beta(\lambda_\theta - \{ \hat \omega \}_k)} + \left( \frac{ \{\hat h\}_k + \theta qX }{\lambda_\theta - \{ \hat \omega \}_k} \right)^2 \right] \right\rangle_{\hat\phi} \ . \label{eq:lambda_stat_RS}
\end{align}
Next, inserting the ansatz into the saddle-point equation for $\mathrm{i}\hat\varphi^\star(\sbv)$ in Eq.~\eqref{SPlambda_phihat}, integrating over $\sbv$ and extracting the $n\to0$ limit we get
\begin{equation}
    \hat\phi \int \de \hat \uppi (\hat \omega, \hat h) \ln \mathcal Z_\beta(\hat \omega, \hat h) = c \int \de \uppi(\omega, h) \mathbb E_W\left[\ln \mathcal Z_\beta \left( \frac{W^2}{\omega}, \frac{h W}{\omega}\right) \right] \ , \label{app:eq:variation1} 
\end{equation}
and, crucially, we observe by inspection that {the normalised joint distribution $\hat\uppi$}
\begin{equation}
    \hat \uppi (\hat \omega, \hat h) = \int \de \uppi(\omega, h) \mathbb E_W\left[ \delta\left(\hat \omega - \frac{W^2}{\omega}\right) \delta\left(\hat h - \frac{hW}{\omega}\right) \right] \ \label{app:eq:pihatRDE}
\end{equation}
solves Eq.~\eqref{app:eq:variation1}, provided that $\hat\phi= c$. With this ansatz, $p^{(\hat\phi)}_k$ is precisely the Poisson distribution with average $c$ {corresponding to the starting graph topology in Eq. \eqref{eq:ER_conndistTrue} in the large $N$ limit}.

Similarly, let us insert the ansatz into the saddle point equation for $\varphi^\star(\sbv)$ in Eq.~\eqref{SPlambda_phi} and multiply by a test kernel $K(\sbv)=\prod_{a=1}^n\frac{1}{\mathcal Z_\beta(\omega',h')}\exp\left(-\beta\frac{\omega'}{2} v^2_a + \beta h' v_a \right)$. By integrating both sides over $\sbv$ we obtain, upon extraction of the $n\to0$ limit,
\begin{equation}\comprimi
    \int \de \uppi~ \ln \frac{\mathcal Z_\beta(\omega+\omega', h + h')}{\mathcal Z_\beta(\omega, h)\mathcal Z_\beta(\omega', h')} = \mathbb E_X \left\llangle \int \{ \de \hat\uppi \}_{k-1} \ \ln \frac{\mathcal Z_\beta(\lambda_\theta - \{ \hat \omega \}_{k-1} + \omega', \theta qX + \{\hat h\}_{k-1} + h')}{\mathcal Z_\beta(\lambda_\theta - \{ \hat \omega \}_{k-1}, \theta X q + \{\hat h\}_{k-1})\mathcal Z_\beta(\omega', h')} \right\rrangle_c.\label{app:eq:variation2}
\end{equation}
Note that, given a function $f(k)$, we have denoted $\frac{1}{c}\langle kf(k)\rangle_c\equiv \llangle f(k)\rrangle_c$, namely the average $\llangle\bullet\rrangle_c$ has to be intended with respect to \textit{degree-corrected distribution} associated to a Poissonian distribution of mean $c$, as per Eq.~\eqref{eq:degreecorr} in the main text. As Eq.~\eqref{app:eq:variation2} must hold for all values of the non-integrated variables $\omega', h'$, it can be translated into 
\begin{align}
    \uppi(\omega,h) = \mathbb E_X \left\llangle\int \{\de \hat \uppi \}_{k-1} \delta(\omega - (\lambda_\theta -\{ \hat{\omega}\}_{k-1})) \delta(h - (\{ \hat{h}\}_{k-1} + \theta q X )) \right\rrangle_c \ , \label{eq:pi_RS}
\end{align}
which solves Eq.~\eqref{app:eq:variation2}. 
By inserting Eq.~\eqref{app:eq:pihatRDE} into the right hand side of Eq.~\eqref{eq:pi_RS}, and replacing the Poissonian degree distribution of average $c$ with a \textit{generic} degree distribution $p_k$ with finite connectivity and bounded maximal degree via the anticipated recipe in Appendix B of \cite{Susca19}, we obtain the \emph{recursive distributional equation} \eqref{eq:RDE} in the main text --- namely, an integral equation for the joint probability density function $\uppi(\omega,h)$ that governs how the $\omega$ and $h$ fields need to be distributed on the edges of an associated auxiliary graph whose weighted adjacency matrix is $\bJ$ (see \cite{Susca21_review} for a more detailed explanation in the context of the cavity method). Equations with this structure appear in the mathematical literature on random graphs \cite{Volkovich07_rde,Jelenkovic09_rde,Chen17_rde,Fraiman23_rde} and in statistical physics and inference \cite{Chertkov10, Semerjian20}, and can be efficiently solved using a population dynamics algorithm \cite{Mezard00} as described in Appendix~\ref{app:popdyn}. 

\paragraph{Average top eigenvalue: exact expression from the action}
\label{app:sec:Lagrange}
Once the saddle point equation for $\uppi$ has been identified, we turn our attention to the explicit expression for $\mathbb E[\lambda_{\rm top}]$. In previous results available in the literature on sparse symmetric matrix models solved via the same method, the Lagrange multiplier $\lambda_\theta$ solving the Eqs.~\eqref{eq:RDEs} was numerically observed to provide precisely this quantity \cite{Susca19}: this fact was later proved for the diluted Wishart model \cite{Budnick25}. Below, we outline the proof within our setup. Let us start by observing that the leading $\beta \to \infty$ term of the action $s_\beta$ reads
\begin{equation}
\lim_{\beta\to+\infty}s_\beta[\uppi,\hat\uppi,q,\lambda_\theta;t=0]=\frac{1}{2} \left( -\frac{c}{2}I_1[\uppi] - c I_2[\uppi] + 2\lambda_\theta - \theta q^2 \right) \ ,
\end{equation}
where we have introduced \footnote{Observe that the appearance of the terms $I_1[\uppi]$ and $I_2[\uppi]$ are consistent with the derivation in \cite{Susca19} for a related model. }
\begin{align}
    I_1[\uppi] &\coloneqq \int \de \uppi(\omega,h) \de \uppi(\omega', h') \,\mathbb E_W\left[ \frac{\left(h + \frac{h'W}{\omega'} \right)^2 }{\omega-\frac{W^2}{\omega'}} - \frac{h^2}{\omega} \right], \label{app:eq:I1} \\
    I_2[\uppi] &\coloneqq \int \de \uppi(\omega,h) \de \uppi(\omega', h')\,\mathbb E_W\left[\left( \frac{h + \frac{W h'}{\omega'}}{\omega - \frac{W^2}{\omega'}} \right)^2 \frac{W^2}{\omega'} \right] \ . \label{eq:I2}
\end{align}
From the top eigenvalue formula in Eq.~\eqref{eq:replica_trick_exchNnAppendix} we therefore obtain
\begin{equation}
    \mathbb E_{\bA} [\lambda_{\rm top}] =  - \frac{c}{2} I_1[\uppi] - c I_2[\uppi] + 2\lambda_\theta - \theta q^2 \ .\label{app:eq:avg_top_eigenvalue_start}
\end{equation}
We will obtain a more compact expression by expanding the leading terms of the entropic term $I_{f}[\hat\uppi,q,\lambda_\theta;t=0]$ (that effectively does not depend on $f$) in the $\beta \to \infty$ limit in two different ways, and recall that in the previous section we have obtained the condition $\hat\phi=c$. First, we write down the leading $\beta \to \infty$ term
\begin{equation}
    I_{f}[\hat\uppi,q,\lambda_\theta;t=0] = \frac{1}{\beta}\mathbb E_X \left\langle \int \{ \de \hat\uppi \}_k \ln \mathcal Z_\beta\left(\lambda_\theta - \{ \hat\omega \}_k, \{ \hat h \}_k + \theta q X \right)\right\rangle_c \xrightarrow{\beta \to \infty} \frac{1}{2} \mathbb E_X \left\langle \int \{ \de \hat\uppi \}_k \frac{(\{ \hat h \}_k + \theta q X)^2}{\lambda_\theta - \{ \hat \omega \}_k}\right\rangle_c \ . \label{app:eq:topeig_ER_S3}
\end{equation}
We now proceed in two ways. First, we divide and multiply by the denominator to simplify the first term using the normalisation in Eq.~\eqref{eq:lambda_stat_RS} to obtain
\begin{align}
\lim_{\beta\to+\infty}I_{f}[\hat\uppi,q,\lambda_\theta;t=0] =&\frac{1}{2} \lambda_\theta - \frac{1}{2} \mathbb E_X \left\langle\int \{ \de \hat \uppi \}_k \left( \frac{\{ \hat h \}_k + \theta q X}{\lambda_\theta - \{ \hat \omega \}_k} \right)^2 \{ \hat \omega \}_k\right\rangle_c \\
    \nonumber = &\frac{1}{2} \lambda_\theta - \frac{c}{2} \mathbb E_X \int \de \hat \uppi(\hat\omega,\hat h)\, \hat\omega\left\llangle \int \{ \de \hat \pi \}_{k-1} \left( \frac{\{ \hat h \}_{k-1} + \hat h + \theta q X}{\lambda_\theta - \{ \hat \omega \}_{s-1} - \hat \omega} \right)^2 \right\rrangle_c\\
    \nonumber \underset{\mathrm{Eq.}~\eqref{eq:pi_RS}}{=}& \frac{1}{2} \lambda_\theta - \frac{c}{2} \int \de \hat \uppi \de \uppi \left( \frac{h + \hat h}{\omega - \hat \omega} \right)^2 \hat\omega \\
    \nonumber \underset{\mathrm{Eq.}~\eqref{app:eq:pihatRDE}}{=} &\frac{1}{2} \lambda_\theta - \frac{c}{2} \int \de \uppi(\omega,h) \de \uppi (\omega',h') \mathbb E_W\left[ \left( \frac{h + \frac{Wh'}{\omega'}}{\omega - \frac{W^2}{\omega'}} \right)^2 \frac{W^2}{\omega'}\right] \\
    =&\frac{\lambda_\theta - c I_2[\uppi]}{2}\ , \label{app:eq:S3_lamtop_lhs}
    \end{align}
where in the third step we have inserted the identity $$1=\int \de \omega \de h\ \delta \left(\hat \omega - (\lambda_\theta - \{ \hat \omega \}_{s-1})\right) \delta (\hat h - (\lambda_\theta - (\{ \hat h \}_{s-1} + \theta q X)))$$ and used Eq.~\eqref{eq:pi_RS}, whereas in the last step we have used Eq.~\eqref{app:eq:pihatRDE} and recognised $I_2[\uppi]$ defined in Eq.~\eqref{eq:I2}.

The second way to proceed from Eq.~\eqref{app:eq:topeig_ER_S3} is to expand one of the factors in the numerator, writing
\begin{align}
\lim_{\beta\to+\infty}I_{f}[\hat\uppi,q,\lambda_\theta;t=0] &= \frac{1}{2} \mathbb E_X \left\langle\int \{ \de \hat \pi \}_{k}  \frac{\{ \hat h \}_{k} + \theta q X}{\lambda_\theta - \{ \hat \omega \}_{k}} \{ \hat h \}_k\right\rangle_c +  \frac{\theta q}{2} \mathbb E_X\left\langle\int \{ \de \hat \pi \}_k \frac{\{ \hat h \}_k X + \theta q X^2}{\lambda_\theta - \{ \hat \omega \}_k}\right\rangle\\
    \nonumber&= \frac{c}{2} \int \de \uppi(\omega,h) \de \uppi(\omega',h') \mathbb E_W\left[\frac{h + \frac{Wh'}{\omega'}}{\omega - \frac{W^2}{\omega'}} \frac{W h'}{\omega'}\right] + \frac{\theta q^2}{2}   \\
    &\equiv \frac{cI_1[\uppi]+2\theta q^2}{4}\ , \label{app:eq:S3_lamtop_rhs}
\end{align}
where for the first term we proceeded analogously to the derivation in Eq.~\eqref{app:eq:S3_lamtop_lhs}, and for the second term we used the expression for $q$ in Eq.~\eqref{eq:Ct0_chat}. 
Next, we have noted \cite{Budnick25} that
\begin{equation}
    \frac{h+\frac{h'W}{\omega'}}{\omega - \frac{W^2}{\omega'}} \frac{h'W}{\omega'} = \frac{1}{2}\left[ \left( \frac{\left(h + \frac{h'W}{\omega'}\right)^2}{\omega - \frac{W^2}{\omega'}} - \frac{h^2}{\omega} \right) - \frac{W^2}{\omega \omega'-W^2} \left( \frac{h^2}{\omega} - \frac{h'^2}{\omega'} \right) \right] \ , \label{app:eq:note}
\end{equation}
and integrated both sides with respect to $\de \uppi(\omega,h)$ and $\de \hat \uppi(\hat \omega, \hat h)$. The double integral of the second term on the R.H.S. of the above equation vanishes due to symmetry of the variables $h \leftrightarrow h', \omega \leftrightarrow \omega'$, thus reducing Eq.~\eqref{app:eq:note} to $ \frac{1}{2}I_1[\uppi]$, defined in Eq.~\eqref{app:eq:I1}.
Finally, by equating Eq.~\eqref{app:eq:S3_lamtop_lhs} with Eq.~\eqref{app:eq:S3_lamtop_rhs}, we write $-\frac{c}{2} I_1[\uppi] - c I_2[\uppi] = - \lambda_\theta + \theta q^2 $ which, inserted back into Eq.~\eqref{app:eq:avg_top_eigenvalue_start}, gives the result
\begin{equation}
    \mathbb E_{\bA} [\lambda_{\rm top}] = \lambda_\theta \ . \label{app:eq:avg_top_eigenvalue}
\end{equation}

\subsection{Top eigenvector component density and overlap}

In this section, we show a detailed derivation of quantities of the form in Eq.~\eqref{remarkable}, namely
\begin{align}
    \lim_{N\to+\infty}\mathbb E_{\bA}[F(u,\bv_{\mathrm{top}},\bx)] &=\lim_{\beta\to+\infty}\lim_{n\to 0}\lim_{N\to+\infty}\frac{1}{\beta n N}\frac{\partial}{\partial t}\ln \mathbb E_{\bA}\big[\big(Z_\beta[\bA;tF]\big)^n\big]\Big|_{t=0} \\
    &= \lim_{\beta\to+\infty}\lim_{n\to 0} \frac{1}{\beta n}\frac{\partial}{\partial t} \mathcal S_\beta[\varphi^\star,\hat\varphi^\star,\psi^\star,\hat\psi^\star,\sblambda^\star(t);t,f] \Big|_{t=0} \ ,
\end{align}
where the action $\mathcal S_\beta$ is given in Eq.~\eqref{eq:app:action_nonRS}. By the chain rule, 
\begin{equation}
    \frac{\partial \mathcal S_\beta}{\partial t}  = \frac{\partial \mathcal S_\beta}{\partial \varphi} \Bigg|_{\star} \frac{\partial \varphi^\star}{\partial t} + \frac{\partial \mathcal S_\beta}{\partial \hat \varphi} \Bigg|_{\star} \frac{\partial \hat \varphi^\star}{\partial t} + \frac{\partial \mathcal S_\beta}{\partial \psi} \Bigg|_{\star} \frac{\partial \psi^\star}{\partial t} + \frac{\partial \mathcal S_\beta}{\partial \hat \psi} \Bigg|_{\star} \frac{\partial \hat \psi^\star}{\partial t} + \frac{\partial \mathcal S_\beta}{\partial \sblambda} \Bigg|_{\star} \frac{\partial \sblambda^\star}{\partial t} + \frac{\partial \mathcal S_\beta}{\partial t} \ ,
\end{equation}
where all the starred terms vanish because the action is stationary at the saddle point.
Therefore, only the explicit dependence on $t$ matters in the entropic term which leads us to 
the expression for the observable
\begin{equation}
    \lim_{N\to+\infty}\mathbb E_{\bA}[F(u,\bv_{\mathrm{top}},\bx)] \equiv\lim_{\beta\to+\infty} \frac{\partial}{\partial t}I_f[\hat\uppi,q,\lambda_\theta;t]\Big|_{t=0} \ , \label{app:eq:observable}
\end{equation}
provided the observable is of the form $F(u,\bv,\bx)=\frac{1}{N}\sum_i f(u,v_i,x_i)$ as in Eq.~\eqref{app:eq:F_defn_f} and $I_f$ is defined in Eq.~\eqref{app:eq:Ifn}.

The final result thus reads as
\begin{equation} \label{app:eq:observable_f_reduced}
    \lim_{N\to+\infty}\mathbb E_{\bA}[F(u,\bv_{\mathrm{top}},\bx)]
    = \lim_{\beta\to+\infty} \mathbb E_X \left\langle\int \{\de \hat\uppi \}_k \frac{ \int \de v ~ f(u,v,X)~ \e^{-\frac{\beta}{2}(\lambda_\theta-\{\hat\omega\}_k) v^2 + \beta (\{\hat h\}_k + \theta q X )v} }{\mathcal Z_\beta(\lambda_\theta-\{\hat\omega\}_k, \{\hat h\}_k + \theta q X )}\right\rangle \ ,
\end{equation}
where we have replaced the Poissonian degree distribution with a generic degree distribution $p_k$ as justified at the end of Appendix.~\ref{sec:app:avg_top_lambda}.

\paragraph{Density of the entries of \texorpdfstring{$\bv_{\mathrm{top}}$}{v1}}
In this section, we choose
\[F(u,\bv_{\mathrm{top}},\bx)\mapsto F_{\rm top}(u,\bv_{\mathrm{top}})\coloneqq\frac{1}{N}\sum_{i=1}^N\delta\left(u-v_{1,i}\right),\]
which corresponds to $f(u,v,X)=\delta(u-vg(X))$ with $g(X)=1$. Inserting this into Eq.~\eqref{app:eq:observable_f_reduced}, we get
\begin{align} 
    \nonumber \rho_{\rm top}(u) &= \lim_{N\to+\infty} \mathbb E_{\bA}[F_{\rm top}(u,\bv_{\mathrm{top}})]= \mathbb E_X \left\langle\int \{\de \hat\uppi \}_k \delta\left(u - \frac{ \{\hat h\}_k + \theta q X}{\lambda_\theta - \{ \hat \omega \}_k} \right)\right\rangle \\
    &= \mathbb E_X \left\langle\int \{\de \uppi \}_k \{\de \rho_W\}_k \delta\left(u -\frac{\left\{ \frac{Wh}{\omega}\right\}_k +\theta q X}{\lambda_\theta - \left\{\frac{W^2}{\omega}\right\}_k } \right)\right\rangle \ , \label{app:eq:top_eig_comp_density}
\end{align}
where in the last step we use the definition of $\hat \uppi$ in Eq.~\eqref{app:eq:pihatRDE}. Setting $\theta=0$ or $\varrho_x(x)=\delta(x)$ indeed recovers the top eigenvector density for the unspiked model analysed in \cite{Susca19}.

\paragraph{Overlap between \texorpdfstring{$\bv_{\mathrm{top}}$ and $\bx$}{v1x}}
In this section, we present the calculation to compute the overlap between the top eigenvector and signal vector. We pick
\begin{equation}
    F(u,\bv_{\mathrm{top}},\bx)\mapsto F_{\rm ov}(u,\bv_{\mathrm{top}},\bx)\coloneqq \frac{1}{N}\sum_{i=1}^N\delta(u-x_iv_{1,i})
\end{equation}
which corresponds to the choice $f(u,v,X)=\delta(u-vg(X))$ with $g(X)=X$. 
Inserting this into Eq.~\eqref{app:eq:observable_f_reduced}, we obtain
\begin{align}
    \nonumber \rho_{\mathrm{ov}}(u) &= \lim_{N\to+\infty} \mathbb E_{\bA}[F_{\rm ov}(u,\bv_{\mathrm{top}},\bx)] = \lim_{\beta\to+\infty} \mathbb E_X \left\langle\int \{\de \hat\uppi \}_k \frac{ \int \de v ~ \delta\left(u-x v\right)~ \e^{-\frac{\beta}{2}(\lambda_\theta-\{\hat\omega\}_k) v^2 + \beta (\{\hat h\}_k + \theta q X )v} }{\mathcal Z_\beta(\lambda_\theta-\{\hat\omega\}_k, \{\hat h\}_k + \theta q X )}\right\rangle  \\
    &= \mathbb E_X \left\langle\int \{\de \uppi \}_k \{\de \rho_W\}_k\delta\left(u - \frac{ X \left\{ \frac{W h}{\omega}\right\}_k +\theta q X^2}{\lambda_\theta - \left\{ \frac{W^2}{\omega}\right\}_k } \right) \right\rangle \ .
\end{align}
Then, we express the average overlap as 
\begin{equation}\label{app:eq:overlap_calc}
    \lim_{N\to \infty}\mathbb E_{\bA}\left[\frac{\langle\bx,\bv_{\mathrm{top}}\rangle}{N}\right] =\lim_{N\to \infty}\int u\,\rho_{\mathrm{ov}}(u)\de u = \mathbb E_X \left\langle\int \{\de \uppi \}_k \{\de \rho_W\}_k  \frac{ X \left\{\frac{W h}{\omega}\right\}_k +\theta q X^2}{\lambda_\theta - \left\{ \frac{W^2}{\omega}\right\}_k } \right\rangle \ .
\end{equation}
Note that this is the same expression as Eq.~\eqref{eq:Ct0_chat} and we thus identify the auxiliary variable $q$ as the typical value of the overlap.

\twocolumngrid
\section{Explicit results for the threshold, top eigenvalue and squared overlap in the RR and truncated Poisson models}\label{app:sec:top_eigenvalue_BBP}

\label{app:sec:general_top_eigenvalue_BBP}
In this Appendix, we outline the derivation of the critical threshold $\theta_\mathrm{crit}$ in the special cases discussed in the main text. We start by defining 
\begin{equation}
\label{app:eq:ml}
m(\lambda_\theta) \coloneqq \int \de  h \de  \omega\, \frac{\uppi(\omega,h)}{\omega} \ .
\end{equation}
Let $W\sim\rho_W$ be a random variable. By Eq.~\eqref{eq:RDE} we obtain
\begin{align}
    \nonumber m(\lambda_\theta) &= \left\llangle\int \{\de  \uppi\}_{k-1} \{\de \rho_W\}_{k-1} \frac{1}{\lambda_\theta - \left\{\frac{W^2}{\omega}\right\}_{k-1}}\right\rrangle \\
    \nonumber&\simeq \frac{1}{\lambda_\theta}\left\llangle\sum_{p=0}^\infty \left(\frac{(k-1)\mathbb E_W[W^2]m(\lambda_\theta)}{\lambda_\theta}\right)^p\right\rrangle\\
    &= \left\llangle\frac{1}{\lambda_\theta - (k-1) \mathbb E_W[W^2] m(\lambda_\theta)} \right\rrangle \ , \label{app:eq:mlambda}
\end{align}
where we have replaced the average of the $p$-th power of the random sum $\{W^2/\omega\}_{k-1}$ by the $p$-th power of its average in the second step, and then resummed the geometric series. The replacement is exact whenever $\uppi$ is a point mass in $\omega$ and the bond weights are deterministic. This is the case in the random regular section in Appendix~\ref{app:sec:RRG_top_eigenvalue_BBP} and in the large connectivity limit of Appendix~\ref{sec:denselimit}, where the sums concentrate by the law of large numbers. For fluctuating degrees and bond weights, Eq.~\eqref{app:eq:mlambda} is an approximation, which we find to be accurate over the range of $\lambda_\theta$ relevant for the observables computed below; a direct comparison with the numerically exact solution obtained via population dynamics is presented in Appendix~\ref{app:popdyn}.
We also recall that $\llangle\bullet\rrangle$ is the average of a function of $k$ over $r_k\propto kp_k$, degree-corrected distribution of $p_k$ in Eq.~\eqref{eq:degreecorr}. 

\subsection{Random regular noise}\label{app:sec:RRG_top_eigenvalue_BBP}

Exact, closed-form analytic expressions for the critical threshold, top eigenvalue and squared overlap can be obtained in the Random Regular setup in Eq.~\eqref{eq:RR}. In this case, each row of the noise matrix $\bJ$ has exactly $c$ nonzero elements. We set for simplicity $\rho_W(W)=\delta(W-1)$ so that the RDE for $\uppi(\omega,h)$ in Eq.~\eqref{eq:RDE} reads as
\begin{multline}
    \uppi(\omega,h) =\\
    \comprimi=\mathbb E_X \int \{\de  \uppi \}_{c-1} \delta \left(\omega -\lambda_\theta + \left\{\frac{1}{ \omega}\right\}_{c-1} \right)\delta \left(h - \left\{\frac{h}{\omega}\right\}_{\mathclap{\quad c-1}}- \theta qX \right)
\end{multline}
and the adopted ansatz therein becomes $\uppi(\omega, h) = \delta(\omega - \bar{\omega}) \mathbb E_X \delta(h - \bar{h}(X)) $,
where $X\sim \rho_x$ has zero mean and variance $\sigma_x^2$ .
First, notice that the ansatz gives us the recursive equation for $\bar{\omega} = \lambda_\theta - \sum_{\ell=1}^{c-1} \frac{1}{\bar{\omega}} = \lambda_\theta - (c-1)\frac{1}{\bar{\omega}}$. Solving for $\bar\omega$,
\begin{equation} \label{eq:ghat}
    m(\lambda_\theta) = \frac{1}{\bar\omega} = \frac{\lambda_\theta - \sqrt{\lambda_\theta^2 - 4(c-1)}}{2(c-1)} \ ,
\end{equation}
The reduced expressions of the normalisation and $q$ are 
\begin{equation}\comprimi
    1 = \mathbb E_X \left( \frac{\left\{\frac{h}{\omega}\right\}_c + \theta q X}{\lambda_\theta -  \left\{\frac{1}{\omega}\right\}_c}\right)^{2} \ \mathrm{and} \ q = \mathbb E_X\left[ \frac{X\left\{\frac{h}{\omega}\right\}_c+\theta qX^2}{\lambda_\theta - \left\{\frac{1}{\omega}\right\}_c}\right] \ , \label{eq:C_RRG}
\end{equation}
respectively. Recall Eq.~\eqref{app:eq:C_xcenteredQ} 
\begin{equation}
    Q(\lambda_\theta)\coloneqq\left\langle \int \{\de  \uppi\}_k \{\de \rho_W\}_k \frac{  1}{\lambda_{\theta} - \left\{\frac{W^2}{\omega}\right\}_k } \right\rangle \ ,
\end{equation} 
which, upon the insertion of the RRG ansatz $\uppi(\omega,h)$, gives 
\begin{equation}
Q(\lambda_\theta)=\frac{1}{\lambda_\theta - c/\bar{\omega}} = \frac{(c-2)\,\lambda_\theta - c\sqrt{\lambda_\theta^2 - 4(c-1)}}{2\left(c^2 - \lambda_\theta^2\right)} \ ,
\end{equation}
which is the Kesten--McKay Stieltjes transform \cite{Kesten59, McKay81} for a random $c$-regular graph. 
Also, recall that for $q\neq0$ Eq.~\eqref{app:eq:C_xcentered} requires $\theta \sigma_x^2 Q (\lambda_\theta)=1$, which is the sparse analogue of the outlier condition for rank-one perturbations \cite{BenaychGeorges11}. Then, the top eigenvalue associated with the signal that detaches from the bulk can be expressed as $\lambda_\theta=c\bar\omega^{-1} + \theta \sigma_x^2$.  
Finally, we use Eq.~\eqref{eq:ghat} to obtain the explicit expression
\begin{equation}
    \lambda_\theta = \frac{c}{2}\sqrt{\theta^2 \sigma_x^4 + 4} - \frac{c-2}{2} \theta \sigma_x^2 \ , \label{app:eq:lamvar}
\end{equation}
which is the value of the eigenvalue associated with the spike once it detaches from the bulk.
Recall that, in the thermodynamic limit, the spectrum of the adjacency matrix of a random $c$-regular graph with connectivity $c$ has bulk edges at $\pm 2 \sqrt{c-1}$ and the structural top eigenvalue is $\lambda_{\theta=0}=c$. For a treatment of elementary spectral properties of graph adjacency matrices, see \cite{Brouwer11}. 
The known result for the right bulk edge can also be verified independently by using the recursion for $m(\lambda_\theta)$ which is exact because the density of $\omega$ is a point mass and we are working with unit bond weights. 
Namely, setting $p_k=\delta_{k,c}$, for which the degree-corrected distribution is $r_k=\frac{k p_k}{c}=\delta_{k,c}$ and setting $\mathbb E_W[W^2]=1$ as  $\rho_W(W)=\delta(W-1)$, we use Eq.~\eqref{app:eq:mlambda} to define
\begin{equation}\label{eq:bulklambdaF}
F(\lambda,m):=m-\frac{1}{\lambda-(c-1) m} \ .
\end{equation}
The real resolvent branch outside the bulk is determined by $F(\lambda,m)=0$. To find the edge, we regard this equation as implicitly defining \(\lambda\) as a function of \(m\). The bulk edge is a branch point of this real solution. 
Such conditions are closely related to the linear-stability criteria used in cavity approaches to determine spectral boundaries and the disappearance of outliers in sparse locally tree-like random matrices \cite{Neri_Metz_outliers16}.
At such a branch point, we have that
\begin{equation}
\frac{\partial F}{\partial m}= 1-\frac{(c-1)}{\left[\lambda-(c-1)m\right]^2} =0 \ .
\end{equation}
At the right edge, we define \(m_{\rm b} :=  m(\lambda_{\rm b})\) to write \((\lambda_{\rm b},m_{\rm b})\), and obtain these two values by solving $F(\lambda_{\rm b},m_{\rm b})=0 $ simultaneously with $\partial_m F(\lambda_{\rm b},m_{\rm b})=0$.
Respectively, this gives
\begin{equation}\label{eq:bulklambda1} 
m_{\rm b}= \frac{1}{\lambda_{\rm b}-(c-1) m_{\rm b}} \ 
\end{equation}
and
\begin{equation}\label{eq:bulklambda2}
1= \frac{(c-1)}{\left[\lambda_{\rm b}-(c-1) m_{\rm b}\right]^2} \ .
\end{equation}
These two equations determine the right bulk edge \(\lambda_{\rm b}\) of the eigenvalue spectrum and the corresponding value \(m_{\rm b}\). For \(\lambda>\lambda_{\rm b}\), the resolvent equation Eq.~\eqref{app:eq:mlambda} admits a stable real solution, and has a critical point at \(\lambda=\lambda_{\rm b}\). At this critical point, we solve for \((\lambda_{\rm b},m_{\rm b})\) by solving Eq.~\eqref{eq:bulklambda1} and Eq.~\eqref{eq:bulklambda2} simultaneously, which is possible due to the simple form of the random regular degree distribution.
Rearranging, we obtain
\begin{equation}\label{app:eq:hello}
\lambda_{\rm b}=\frac{1}{m_{\rm b}} + (c-1) m_{\rm b} \ ,
\end{equation}
and substituting it into the second equation gives $m_{\rm b}=1/\sqrt{(c-1)}$, and picking the positive branch we obtain
\begin{equation}
\lambda_{\rm b}=\frac{1}{m_{\rm b}}+(c-1) m_{\rm b}=2\sqrt{c-1} \ ,
\end{equation}
which agrees with known results \cite{Brouwer11}.

Using the previous facts, two thresholds in $\theta$ are derived for the eigenvalue associated with the spike.
The first threshold $\theta_{\mathrm{b}}$ corresponds to the value of the signal-to-noise ratio at which $\lambda_\theta$ detaches from the bulk, i.e., where $\lambda_\theta=\lambda_{\mathrm{b}}=2 \sqrt{c-1}$, which gives the condition
\begin{equation}
    \theta_{\mathrm{b}} = \frac{c-2}{\sigma_x^2 \sqrt{c-1}} \ . \label{eq:theta_b}
\end{equation}
 The second threshold $\theta_{\rm crit}$ corresponds to the condition $\lambda_{\theta=0}= \lambda_{\theta=\theta_{\mathrm{crit}}}$ expressing that $\lambda_\theta$ overtakes the top structural eigenvalue $\lambda_{\theta=0}$, giving
\begin{equation}
    \theta_\mathrm{crit} = \frac{c(c-2)}{\sigma_x^2 (c-1)} \ . \label{eq:theta_c}
\end{equation}

Simply inverting Eq.~\eqref{eq:theta_b}, we obtain the value of $c$ at which, for a given spike strength $\theta$ and variance $\sigma_x^2$, the eigenvalue associated with the spike gets buried in the bulk
\begin{equation}
    c_{\rm b} = \frac{1}{4} \left( \theta \sigma_x^2 + \sqrt{\left(\theta \sigma_x^2\right)^2 +4} \right)^2 + 1 \ , \label{app:eq:c_b}
\end{equation}
whereas inverting Eq.~\eqref{eq:theta_c}, we can determine the value of $c$ at which the structural eigenvalue takes over the one associated with the spike, namely
\begin{equation}
    c_\mathrm{crit} = \frac{1}{2} \left( \theta \sigma_x^2 + \sqrt{\left(\theta \sigma_x^2\right)^2 + 4} \right) + 1 \ . \label{app:eq:c_c}
\end{equation}
To compute the squared overlap, consider the partition function defined in Eq.~\eqref{eq:Z_lambdatop_soft}
\begin{equation}\comprimi
    Z_\beta(\bA) = \int \de \bv~\exp\left[\frac{\beta}{2}\langle\bv,\bJ\bv\rangle + \frac{\beta \theta}{N} \langle \bx, \bv \rangle^2 \right] \delta\left(\|\bv\|^{2}-N\right) \ .
\end{equation}
For $\theta>\theta_{\rm crit}$, we take the derivative of its logarithm with respect to the signal strength parameter, which gives us in the large $\beta$ limit 
\begin{equation} \label{eq:overlap_derivation}
    \frac{\partial \mathbb{E}_{\bA}[\lambda_{\mathrm{top}}]}{\partial \theta} = \lim_{\beta \to \infty} \frac{2}{N} \frac{\partial}{\partial \theta} \mathbb E_\bA \ln Z_\beta(\bA) = \mathbb E_\bA \left[ \frac{\langle\bx,\bv_{\mathrm{top}}\rangle^2}{N^2} \right] \ ,
\end{equation}
where we recall the Courant--Fischer--Weyl formulation in Section~\ref{sec:formulation}, and note this is known as the Hellmann–Feynman method which holds for any non-degenerate eigenpair. We thus obtain the value of the squared overlap for $\theta>\theta_{\mathrm{crit}}$ by differentiating the expression in Eq.~\eqref{app:eq:lamvar}
\begin{equation}
    \mathbb E_\bA \left[ \frac{\langle\bx,\bv_{\mathrm{top}}\rangle^2}{N^2} \right] = \frac{c \theta \sigma_x^4}{2\sqrt{(\theta \sigma_x^2)^2 + 4}} - \frac{(c-2) \sigma_x^2}{2} \ .
\end{equation}
The above expression also gives the typical squared overlap between the signal and the eigenvector associated with the \textit{second}-largest, signal-bearing eigenvalue in the metastable regime. More precisely, as $\theta_{\rm b}<\theta\leq\theta_{\rm crit}$, we have $\lambda_{{\rm b}}<\lambda_\theta\leq\lambda_{\theta=0}$ and
\begin{equation}
    \mathbb E_\bA \left[ \frac{\langle\bx,\bv_{2}\rangle^2}{N^2} \right] = \frac{c \theta \sigma_x^4}{2\sqrt{(\theta \sigma_x^2)^2 + 4}} - \frac{(c-2) \sigma_x^2}{2} \ ,
\end{equation}
where $\bv_{2}$ is the eigenvector associated with the second largest eigenvalue.

\subsection{{Poissonian noise}}\label{app:sec:ER_top_eigenvalue_BBP}

Let us now consider the {truncated Poissonian noise case, namely a noise matrix generated by using $\rho_W(W)=\delta(W-1)$ and $$p_k =\frac{\bar{c}^k}{\Gamma k!} \ ,\qquad \Gamma\coloneqq \sum_{k=0}^{k_{\mathrm{max}}}\frac{\bar{c}^k}{k!} \ .$$ We will present here the derivation of Result~\ref{res:ER}.
Consider Eq.~\eqref{app:eq:C_xcenteredQ} for $Q(\lambda_\theta)$ that determines the critical value $\theta_{\mathrm{crit}}$,
\begin{equation} \label{app:eq:C_xcenteredAppendix}
    Q(\lambda_\theta) = \left\langle \int \{\de  \uppi\}_k \{\de \rho_W\}_k \frac{  1}{\lambda_{\theta} - \left\{\frac{W^2}{\omega}\right\}_k } \right\rangle \ ,
\end{equation}
where the average $\langle\bullet\rangle$ is taken as usual w.r.t. $p_k$. We now wish to re-express the right hand side of \eqref{app:eq:C_xcenteredAppendix} in terms of $m(\lambda_\theta)$, as defined in Eq.~\eqref{app:eq:ml}. To do so, we express $m(\lambda_\theta)$ in a different way. Multiplying both sides of Eq. ~\eqref{eq:RDE} by $1/\omega$ and integrating, we obtain
\begin{multline*}
m(\lambda_\theta)=\\
\begin{aligned}
&\comprimi=\sum_{k=1}^{k_{\mathrm{max}}}\frac{\bar c^k}{(k-1)! c\Gamma}\int \{\de  \uppi\}_{k-1} \{\de \rho_W\}_{k-1}\frac{  1}{\lambda_{\theta} - \left\{\frac{W^2}{\omega}\right\}_{k-1}}\\
&\comprimi=\sum_{k=0}^{k_{\mathrm{max}}-1}\frac{\bar c}{c}\frac{\bar c^k}{\Gamma k!}\int \{\de  \uppi\}_{k} \{\de \rho_W\}_{k}\frac{  1}{\lambda_{\theta} - \left\{\frac{W^2}{\omega}\right\}_{k}}\\
&\simeq \frac{\bar c}{c}\left[\tilde Q(\lambda_\theta)-R(\theta,k_{\mathrm{max}})\right]\ ,
\end{aligned}\end{multline*}
where the correction term $R(\theta,k_{\mathrm{max}})$ takes care of the mismatch between the upper limits of the sums. It reads
\begin{multline*}
R(\theta,k_{\mathrm{max}})\coloneqq\\
\begin{aligned}
&\comprimi=\frac{\bar c^{k_{\mathrm{max}}}}{\Gamma k_{\mathrm{max}}!}\int \{\de  \uppi\}_{k_{\mathrm{max}}} \{\de \rho_W\}_{k_{\mathrm{max}}}\frac{  1}{\lambda_{\theta} - \left\{\frac{W^2}{\omega}\right\}_{k_{\mathrm{max}}} }\\
&\comprimi\simeq\frac{\bar c^{k_{\mathrm{max}}}}{\Gamma k_{\mathrm{max}}!}\frac{1}{\lambda_\theta - k_{\mathrm{max}} \mathbb E_W[W^2] m(\lambda_\theta)}\ ,
\end{aligned}\end{multline*}
where we have again expanded and resummed the geometric series. Summarising, we get
\begin{equation}\label{eq:Qlambda}
\comprimi   \tilde  Q(\lambda_\theta) \simeq \frac{c}{\bar c}m(\lambda_\theta)+\frac{\bar c^{k_{\mathrm{max}}}}{\Gamma k_{\mathrm{max}}!}\frac{1}{\lambda_\theta - k_{\mathrm{max}} \mathbb E_W[W^2] m(\lambda_\theta)}\ ,
\end{equation}
where $m(\lambda_\theta)$ satisfies Eq. \eqref{app:eq:mlambda}. The result for $\theta_{\mathrm{crit}}$ in Eq.~\eqref{eq:gen_transition} readily follows. }

Then, for a set value of $\theta>\theta_{\mathrm{crit}}$, the top eigenvalue associated with the signal is an outlier and we obtain its value via
\begin{equation}
    \lambda_\theta \simeq \tilde  Q^{-1}\left(\frac{1}{\theta \sigma_x^2 } \right), \label{app:eq:lam_signal_m}
\end{equation}
which can be easily determined by numerically solving Eq.~\eqref{eq:Qlambda} using Eq.~\eqref{app:eq:mlambda}.
Using Eq.~\eqref{app:eq:lam_signal_m} we obtain the value of the squared overlap for $\theta>\theta_{\mathrm{crit}}$ as
\begin{equation}
    \mathbb E_\bA \left[ \frac{\langle\bx,\bv_{\mathrm{top}}\rangle^2}{N^2} \right] \simeq -\frac{1}{\sigma_x^2 \theta^2 \tilde Q'(\lambda_{\theta})} \ .
\end{equation}

\paragraph*{Range of validity and the bulk edge.
The results of this subsection rely on the closure in Eq.~\eqref{app:eq:mlambda}, which neglects the variance of $1/\omega$ induced by degree fluctuations. They should therefore be regarded as approximations, whose accuracy we quantify in Appendix~\ref{app:popdyn} for $\theta>\theta_{\mathrm{crit}}$. The bulk edge $\lambda_{\rm b}$ shown in Fig.~\ref{fig:ER_transition} is obtained by the same branch-point argument used in the random regular case (starting from Eq.~\eqref{eq:bulklambdaF}), now applied to the closure Eq.~\eqref{app:eq:mlambda} with a general degree-corrected distribution $r_k$, i.e.\ by solving
\begin{align}
m_{\rm b}&\simeq\Big\llangle\frac{1}{\lambda_{\rm b}-(k-1)\mathbb E_W[W^2]\,m_{\rm b}}\Big\rrangle \ , \\
1&\simeq\Big\llangle\frac{(k-1)\mathbb E_W[W^2]}{\big[\lambda_{\rm b}-(k-1)\mathbb E_W[W^2]\,m_{\rm b}\big]^2}\Big\rrangle
\end{align}
simultaneously. Unlike in the random regular case, this is only an estimate, since the closure is not exact for fluctuating degrees and the spectral density does not have a sharp edge. The continuation of $\lambda_\theta$ and of the squared overlap into the intermediate regime $\theta_{\rm b}<\theta\le\theta_{\rm crit}$, together with $\lambda_{\rm b}$ and $\theta_{\rm b}$, is therefore shown only as a guide to the eye; see Remark~\ref{rem:poisson_validity} for a discussion.}

\subsection{Large connectivity limit}\label{sec:denselimit}
In this section, we follow \cite{Susca19, Susca21_review} to determine the large connectivity limit $c=\langle k \rangle \to \infty$ of our results for sparse matrices. We impose the condition $\frac{\langle k^2 \rangle - \langle k \rangle^2}{\langle k \rangle^2} \to 0$ (i.e., the distribution concentrates around $c=\langle k\rangle$ in this limit) and rescale the weights in $\bW$ to be of the form $W_{ij} = \frac{1}{\sqrt{c}} Z_{ij}$. Let us also assume the signal distribution to be zero-centered with variance $\sigma_x^2$. In this limit and with the introduced assumptions, we expect to recover the well known BBP-like transition value $\theta_{\mathrm{crit}}=\frac{1}{\sigma_x^2}$ for the spiked (real) Wigner matrix ensemble \cite{Peche06,Capitaine09,Bloemendal_2012}. In this model, for $\theta>\theta_{\mathrm{crit}}=\frac{1}{\sigma_x^2}$, the top eigenvalue is $\lambda_\theta=\sigma_x^2\theta + \frac{1}{\sigma_x^2\theta}$ and the squared overlap between the signal and the top eigenvector is $\sigma_x^2-(\theta \sigma_x)^{-2}$. 

In our setup, the transition value is given in Eq.~\eqref{eq:theta_crit}
\begin{equation}
    \frac{1}{\theta_{\mathrm{crit}}} = \sigma_x^2 \left\langle \int \{\de  \uppi\}_k \{\de \rho_W\}_k \frac{ 1}{\lambda_{\theta=0} - \left\{\frac{W^2}{\omega}\right\}_k} \right\rangle \ . \label{app:eq:thetacrit_fordense}
\end{equation}
In what follows, we shall be looking for an ansatz for the density $\uppi(\omega,h)$ as $\langle k \rangle \to \infty$. First, consider the argument of the first $\delta$-function on the R.H.S of Eq.~\eqref{eq:RDE}, which expresses the following identity in distribution,
\begin{equation}
    \omega \stackrel{\rm d}{=} \lambda_{\theta=0} - \frac{1}{c} \sum_{\ell=1}^{k -1} \frac{Z_\ell^2}{\omega_\ell} \ ,
\end{equation}
where $k$ is randomly distributed as $p_k$ and $Z_\ell$ is a set of $k-1$ rescaled bond weights.
Given our assumptions and the concentration of $k$ around $c$, by the Law of Large Numbers (LLN) as $\langle k \rangle \to \infty$, we have a concentration phenomenon leading to a deterministic value for $\omega$, which we denote $\bar\omega$ and satisfies 
\begin{equation}
    \bar{\omega} = \lambda_{\theta=0} - \frac{\mathbb E_Z[Z^2]}{\bar{\omega}} \ .\label{eq:recursion_omegabar}
\end{equation}

Specialising for simplicity to $\mathbb E_Z[Z]=0$ and $\mathbb E_Z[Z^2]=1$, the spectral density of $\bJ$ is known to converge, as $c \to \infty$, to the Wigner semicircle law $\rho(\lambda)=\frac{1}{2\pi}\sqrt{4-\lambda^2}$ with the right bulk edge at $\lambda_{\theta=0}=2$. 
Inserting $\lambda_{\theta=0}=2$ and $\mathbb E_Z[Z^2]=1$ into the recursion for $\bar \omega$ in Eq.~\eqref{eq:recursion_omegabar} we obtain $\bar \omega = 1$. 
We introduce therefore the ansatz $\uppi(\omega, h) = \delta(\omega - 1) \uppi_h(h)$, where $\uppi_h$ is the density of the $\{ h_\ell\}$ variables. Note, however, that $\uppi_h$ is irrelevant for the calculation of $\theta_{\rm crit}$ as  $\{ h_\ell \}$ do not appear in Eq.~\eqref{app:eq:thetacrit_fordense}.
Inserting $\lambda_{\theta=0}=2$ and the ansatz $\uppi(\omega, h) = \delta(\omega - 1) \uppi_h(h)$ into Eq.~\eqref{app:eq:thetacrit_fordense} we finally obtain the anticipated value
\begin{equation}
    \theta_{\mathrm{crit}} = \frac{1}{\sigma_x^2} 
\end{equation}
in the large connectivity limit.

For $\theta>\theta_{\mathrm{crit}}$ where the overlap is nonzero, the condition in Eq.~\eqref{eq:C_pi} (as $q\neq 0$) gives in the large connectivity limit
\begin{equation} \label{app:eq:lambda}
  \theta \sigma_x^2=  \lambda_\theta - \frac{1}{\bar{\omega}}=\bar\omega
\end{equation}
where the large-$c$ limit of the first $\delta$-function on the R.H.S of Eq.~\eqref{eq:RDE} (of the form as Eq.~\eqref{eq:recursion_omegabar} with $\lambda_\theta$) has been used. This leads to
\begin{equation}
    \lambda_{\theta} = \theta \sigma_x^2 + \frac{1}{\theta \sigma_x^2} \ .
\end{equation}

Finally, the squared overlap is given
\begin{equation}
    \mathbb E_\bA \left[ \frac{\langle\bx,\bv_{\mathrm{top}}\rangle^2}{N^2} \right] = \frac{\partial \lambda_{\theta}}{\partial \theta} = \sigma_x^2 - \frac{1}{\theta^2 \sigma_x^2}\ ,
\end{equation}
once again recovering all relevant values in the spiked real Wigner case \cite{Peche06,Capitaine09,BenaychGeorges11, Bloemendal_2012}.

\section{Spiked population dynamics}
\label{app:popdyn}
In this Appendix, we describe the implementation of the population dynamics algorithm (PDA) \cite{Mezard00} adopted to efficiently determine the joint probability density $\uppi(\omega,h)$ in Eq.~\eqref{eq:RDEs}.
This algorithm is outlined for $\theta>\theta_{\mathrm{crit}}$ (otherwise we are in the $\theta=0$ case studied in \cite{Susca19}).

The joint probability density function $\uppi(\omega,h)$ that solves Eq.~\eqref{eq:RDE} is represented by a set of $N_p$ pairs $\mathcal P= \{(\omega_i, h_i) \}_{i=1}^{N_p}$, initialised using a random uniform distribution on $[5,20]$ for $\omega$ and $[0,10]$ for $h$. For a given set of parameters $(q_\mathrm{init},\lambda_{\theta,\mathrm{init}}, \theta)$ (we usually initialise with generic values $q_\mathrm{init}=0.5$, $\lambda_{\theta,\mathrm{init}}=10$), we perform the following steps:
\begin{enumerate}
    \item Sample an integer $k$ from $r_k\propto kp_k$.
    \item Select at random $k-1$ pairs $\{(\omega_\ell,h_\ell)\}_{\ell=1}^{k-1}$ from $\mathcal P$.
    \item Sample $k-1$ values $W_{\ell}$ with density $\rho_W$ and a value $X\sim \varrho_x$.
    \item Compute a new pair $(\omega_{\mathrm{new}},h_{\mathrm{new}})$ as
    \begin{equation}
    \nonumber \left(\lambda_{\theta,\mathrm{init}} - \left\{\frac{W^2}{\omega}\right\}_{k-1}, \ \left\{\frac{hW}{\omega}\right\}_{k-1} + \theta q_\mathrm{init} X\right)
    \end{equation} as prescribed by Eq.~\eqref{eq:RDE}.
    \item Replace a randomly selected pair in $\mathcal P$ with $(\omega_{\mathrm{new}},h_{\mathrm{new}})$.
    \item Monitor four moments of the population for a set number of steps of the algorithm. Once they plateau, $\mathcal P$ has reached equilibrium and the process ends, otherwise return to (1) and repeat with the updated population and same initial parameters $(q_\mathrm{init},\lambda_{\theta,\mathrm{init}}, \theta)$.
\end{enumerate}
Once the population $\mathcal P$ has reached equilibrium, we check the conditions that the population must satisfy. Compute the constants
\begin{equation}\comprimi
\nonumber \alpha_1 = \mathbb E_X \left\langle\int \{ \de  \pi \}_k \{\de \rho_W\}_{k} \left(\frac{ \left\{\frac{hW}{\omega}\right\}_k + \theta qX}{\lambda_\theta - \left\{\frac{W^2}{\omega}\right\}_k }\right)^{2}\right\rangle \  \label{eq:resc_lam_2}
\end{equation}
that checks Eq.~\eqref{eq:lambda_stat_pi_only} and 
\begin{equation}\comprimi
\nonumber \alpha_2 = \theta \mathbb E_X \left\langle\int \{ \de  \pi \}_k \{\de \rho_W\}_{k} \frac{X^2}{\lambda_{\theta} - \left\{\frac{W^2}{\omega}\right\}_k} \right\rangle \  \label{eq:resc_lam_22}
\end{equation}
that checks Eq.~\eqref{eq:C_pi} (we know $q\neq0$ so we divide that equation by $q$ on both sides).
If, within some pre-fixed tolerance, $\alpha_1=1$ and $\alpha_2=1$ for an equilibrated population $\mathcal P$, we have obtained the population of interest that satisfies Eqs.~\eqref{eq:RDE}, \eqref{eq:lambda_stat_pi_only} and \eqref{eq:C_pi}.

Otherwise, if $\alpha_1\neq1$ and $\alpha_2\neq1$, instead of searching over a grid $(q,\lambda_\theta)$, we perform rescaling steps informed by $\alpha_1,\alpha_2$ that considerably speed up convergence to the desirable values. Rescale $q_{\mathrm{new}}=\frac{1}{\sqrt{\alpha_1}} q_{\mathrm{init}}\mapsto q_{\mathrm{init}}$ and $\lambda_{\theta,\mathrm{new}} = \alpha_2 \lambda_{\theta,\mathrm{init}}\mapsto \lambda_{\theta,\mathrm{init}} $ and repeat the above PDA steps (1--6) with parameters $(q_{\mathrm{init}}=q_{\mathrm{new}}, \lambda_{\theta,\mathrm{init}} = \lambda_{\theta,\mathrm{new}},\theta)$. Upon convergence, if $\alpha_1\neq1$ and $\alpha_2\neq1$, rescale $q_\mathrm{new}$ and $\lambda_{\theta,\mathrm{new}}$ and repeat the PDA steps (1--6) until we have obtained $\alpha_1=1$ and $\alpha_2=1$ (up to a suitable tolerance threshold).

\begin{figure}[b]
\centering
\includegraphics[width=\columnwidth]{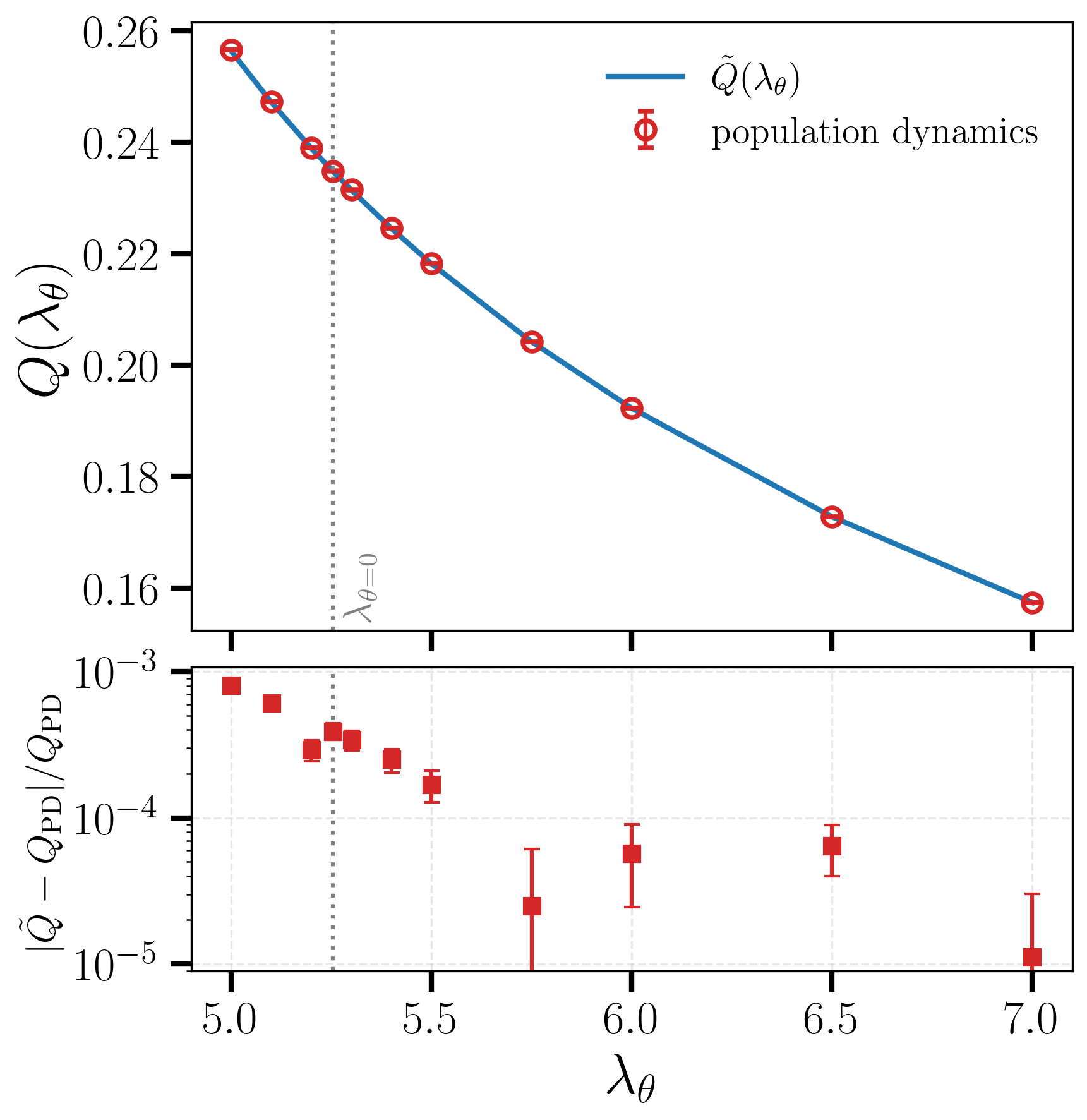}
\caption{Accuracy of the closure in Eq.~\eqref{app:eq:mlambda}, for the truncated Poissonian setup with $c=4$, $k_{\max}=20$ and $\varrho_W(W)=\delta(W-1)$. (\emph{Top}) The closed-form $\tilde Q(\lambda_\theta)$ of Eq.~\eqref{eq:Qlambda} (solid line) against $Q(\lambda_\theta)$ evaluated by Monte Carlo sampling over the equilibrated population (circles). (\emph{Bottom}) Relative discrepancy between the two. It stays below $10^{-3}$ over the whole range, including at $\lambda_{\theta=0}$ (dotted line), the value that fixes $\theta_{\rm crit}$ in Eq.~\eqref{eq:gen_transition}. Population size $N_p=2\times 10^5$; $Q$ is averaged over $150$ sweeps after equilibration, with error bars of standard error.}
\label{fig:closure}
\end{figure}

The above population-dynamics procedure is formulated for generic initial values of $q_\mathrm{init},\lambda_{\theta,\mathrm{init}}$, which need not coincide with their true fixed-point values; the iterative rescaling based on the monitored moments $\alpha_1$ and $\alpha_2$ is precisely what updates these quantities until the self-consistency conditions are satisfied. We have observed that this iterative rescaling procedure converges reliably in the cases tested, including degree distributions for which no closed-form threshold is available.
Note, however, that in the settings of interest we have obtained $\lambda_{\theta,\mathrm{init}}$ and $q_\mathrm{init}$ explicitly. We use them as input parameters $(q,\lambda_{\theta}, \theta)$: for the truncated Poissonian degree distribution in Result~\ref{res:ER}, we use Eq.~\eqref{eq:gen_lambda_signal_all} to get $\lambda_{\theta}$ and Eq.~\eqref{eq:gen_overlap} to get $q$, and for random regular in Result~\ref{res:RRG} we use Eq.~\eqref{eq:RRG_lambdasignal} for $\lambda_\theta$ and Eq.~\eqref{eq:RRG_overlap} for $q$. Once these values are obtained, we perform one sweep of the PD algorithm (steps 1--6) to obtain the equilibrium population $\mathcal P$ and use it to compute relevant observables.

\begin{figure}
  \centering
    \includegraphics[width=\linewidth]{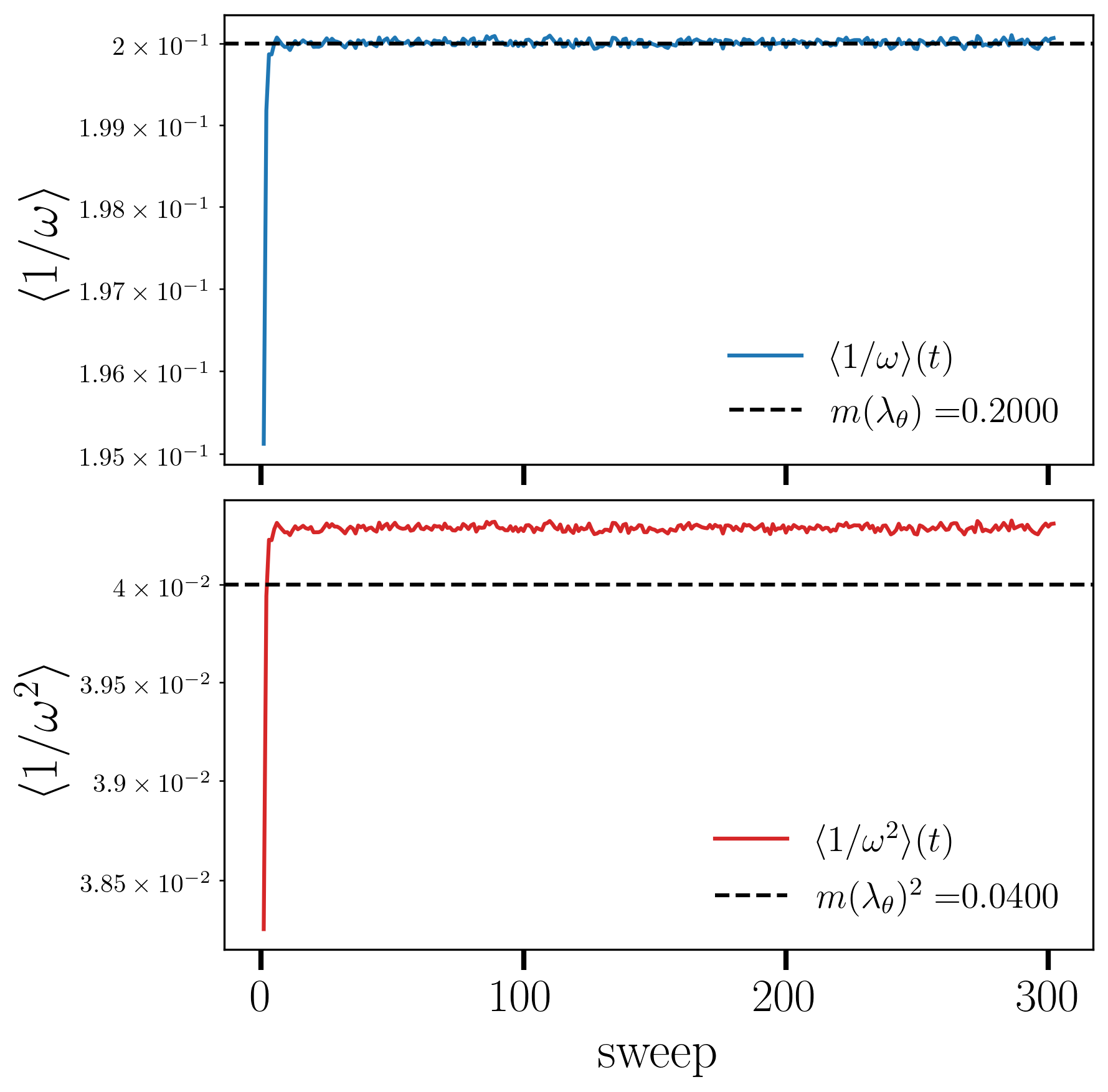} 
  \caption{Evolution of the first two inverse moments of $\omega$ under the population dynamics, for the Poissonian case with $k_{\max}=20$, $c=4$, $\varrho_W(W)=\delta(W-1)$ and $\theta=5$. (\emph{Top}) $\mathbb{E}[1/\omega]$ converges to $m(\lambda_\theta)$ from Eq.~\eqref{app:eq:mlambda} (dashed line). (\emph{Bottom}) $\mathbb{E}[1/\omega^2]$ converges slightly above $m(\lambda_\theta)^2$ (dashed line); the residual gap is $\mathrm{Var}(1/\omega)$, the quantity neglected by the closure.}
  \label{fig:invomega}
\end{figure}

We finally use an equilibrated population to assess the accuracy of the closure introduced in Eq.~\eqref{app:eq:mlambda}. Since the first $\delta$-function on the right-hand side of Eq.~\eqref{eq:RDE} fixes $\omega$ in terms of the $\omega_\ell$ of the sampled neighbours alone, the marginal of $\uppi(\omega,h)$ in $\omega$ obeys a closed recursion, in which $\theta$ and $q$ enter only through the value of $\lambda_\theta$ at which it is evaluated. As $Q(\lambda_\theta)$ in Eq.~\eqref{app:eq:C_xcenteredAppendix} is a functional of that marginal alone, it is sufficient to evolve the $\omega$ population, and a single sweep in $\lambda_\theta$ covers all signal strengths. This also gives access to $\lambda_\theta<\lambda_{\theta=0}$, where we observe that the $h$ population no longer converges.
The comparison is shown in Fig.~\ref{fig:closure}. The population is initialised at $\omega_i=\lambda_\theta$ for all $i$, the $k=1$ fixed point of the recursion, and equilibrated for $80$ sweeps before any measurement is taken; the stationary state does not depend on this choice. The closed-form $\tilde Q(\lambda_\theta)$ of Eq.~\eqref{eq:Qlambda} agrees with $Q(\lambda_\theta)$ sampled from the population to better than $10^{-3}$ in relative terms over the whole range, and to $4\times 10^{-4}$ at $\lambda_{\theta=0}$; the relative error on $\theta_{\rm crit}$, inherited directly through $\theta_{\rm crit}=1/(\sigma_x^2\tilde Q(\lambda_{\theta=0}))$, is of the same order.

In Fig.~\ref{fig:invomega}, we show the evolution of the first two inverse moments of $\omega$ during the population dynamics algorithm in the recovery phase. At convergence, $\mathbb{E}[1/\omega]$ reaches the value $m(\lambda_\theta)$ computed explicitly from Eq.~\eqref{app:eq:mlambda} (top panel), thus verifying Eq.~\eqref{app:eq:ml}. The second inverse moment $\mathbb{E}[1/\omega^2]$ (bottom panel) instead settles slightly above $m(\lambda_\theta)^2$, the value it would take were the closure exact: the gap between the two is the variance of $1/\omega$ over $\uppi$, which is precisely the quantity discarded in passing from the first to the second line of Eq.~\eqref{app:eq:mlambda}. This is done for $k_{\mathrm{max}}=20$, $c=4$, $\rho_W(W)=\delta(W-1)$ and standard Gaussian signal strength $\theta=5$.
We therefore can conclude that while the approximation discards the variance, Fig.~\ref{fig:closure} shows that it is accurate enough, compared to population dynamics which are much more numerically demanding, to determine the relevant spectral properties.


\bibliographystyle{unsrt}
\bibliography{refs.bib}
\end{document}